\documentclass[conference]{IEEEtran}
\IEEEoverridecommandlockouts

\usepackage{times}
\usepackage[numbers]{natbib}
\usepackage{multicol}
\usepackage[bookmarks=true]{hyperref}

\usepackage[dvipsnames]{xcolor}
\usepackage{titletoc}
\usepackage{booktabs}
\usepackage{array}   
\usepackage{graphicx}
\usepackage{subcaption}
\usepackage{epsfig} 
\usepackage{algorithm}
\usepackage{algpseudocode}
\usepackage{tabularx}
\usepackage{multirow}
\usepackage{stfloats}
\usepackage{caption}
\usepackage[table]{xcolor}

\usepackage{float}
\usepackage{rotating}
\usepackage{pdflscape}
\usepackage{adjustbox}
\usepackage{tikz}
\usepackage{amsthm,amsmath,amssymb,mathtools}
\usepackage{graphicx}
\usepackage{caption}
\usepackage{cuted}
\usepackage{stfloats}

\newtheorem{assumption}{Assumption}
\newtheorem{definition}{Definition}
\newtheorem{proposition}{Proposition}
\newtheorem{lemma}{Lemma}

\newtheorem{corollary}{Corollary}

\usetikzlibrary{positioning}

\usepackage{eso-pic}

\begin{document}
\AddToShipoutPictureBG*{%
  \AtPageUpperLeft{%
    \hspace{0.5\paperwidth}%
    \raisebox{-1.1cm}{%
      \makebox[0pt][c]{%
        \parbox{\textwidth}{%
          \centering
          \normalsize
          \textit{This paper has been accepted for publication in Robotics: Science and Systems (RSS) 2026.}
        }%
      }%
    }%
  }%
}
\title{Learning Point Cloud Geometry as a Statistical Manifold: Theory and Practice}

\author{
Jinwoo Lee$^{*,1}$\quad Jiwoo Kim$^{*,1}$\quad Woojae Shin$^{1}$\quad Giseop Kim$^{2}$\quad Hyondong Oh$^{1,\dagger}$ \thanks{$^{*}$Equal contribution. $^{\dagger}$Corresponding author.} \\[2mm] {\small $^{1}$ Korea Advanced Institute of Science and Technology (KAIST), Daejeon, Republic of Korea}\\ {\small $^{2}$ Daegu Gyeongbuk Institute of Science and Technology (DGIST), Daegu, Republic of Korea}\\ {\small Emails: \texttt{\{jinwoolee, tars0523, oj7987, h.oh\}@kaist.ac.kr}, \texttt{gsk@dgist.ac.kr}} }

\maketitle

\begin{strip}
    \centering
    \vspace{-5.8em}

    % ---------- Row 1 ----------
    \begin{minipage}[t]{0.45\textwidth}
        \centering
        \includegraphics[width=\linewidth,keepaspectratio,
        trim=700 400 700 360,clip]{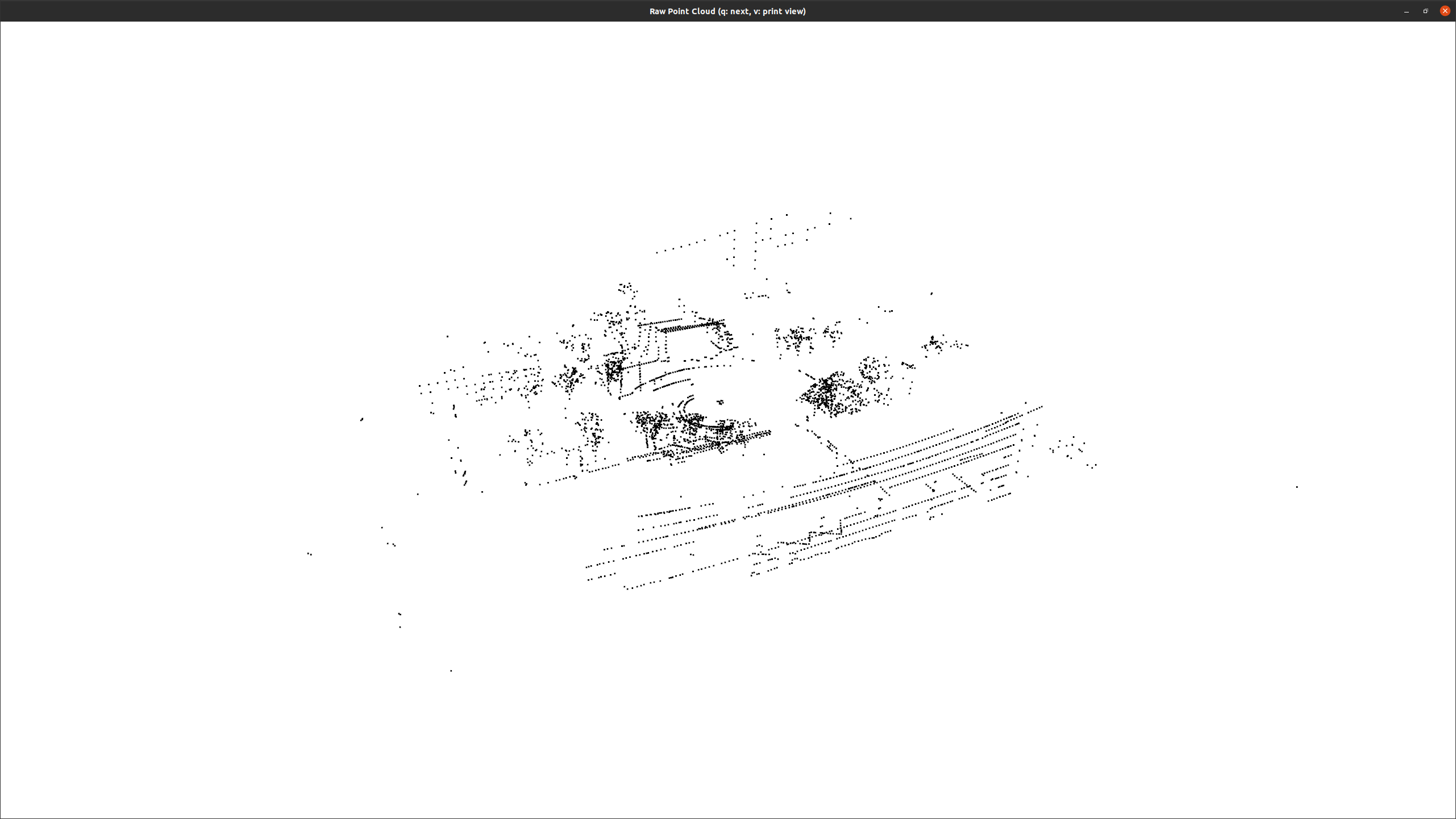}
        {Raw}
    \end{minipage}
    \hspace{0.001\textwidth}
    \begin{minipage}[t]{0.45\textwidth}
        \centering
        \includegraphics[width=\linewidth,keepaspectratio,
        trim=700 400 700 360,clip]{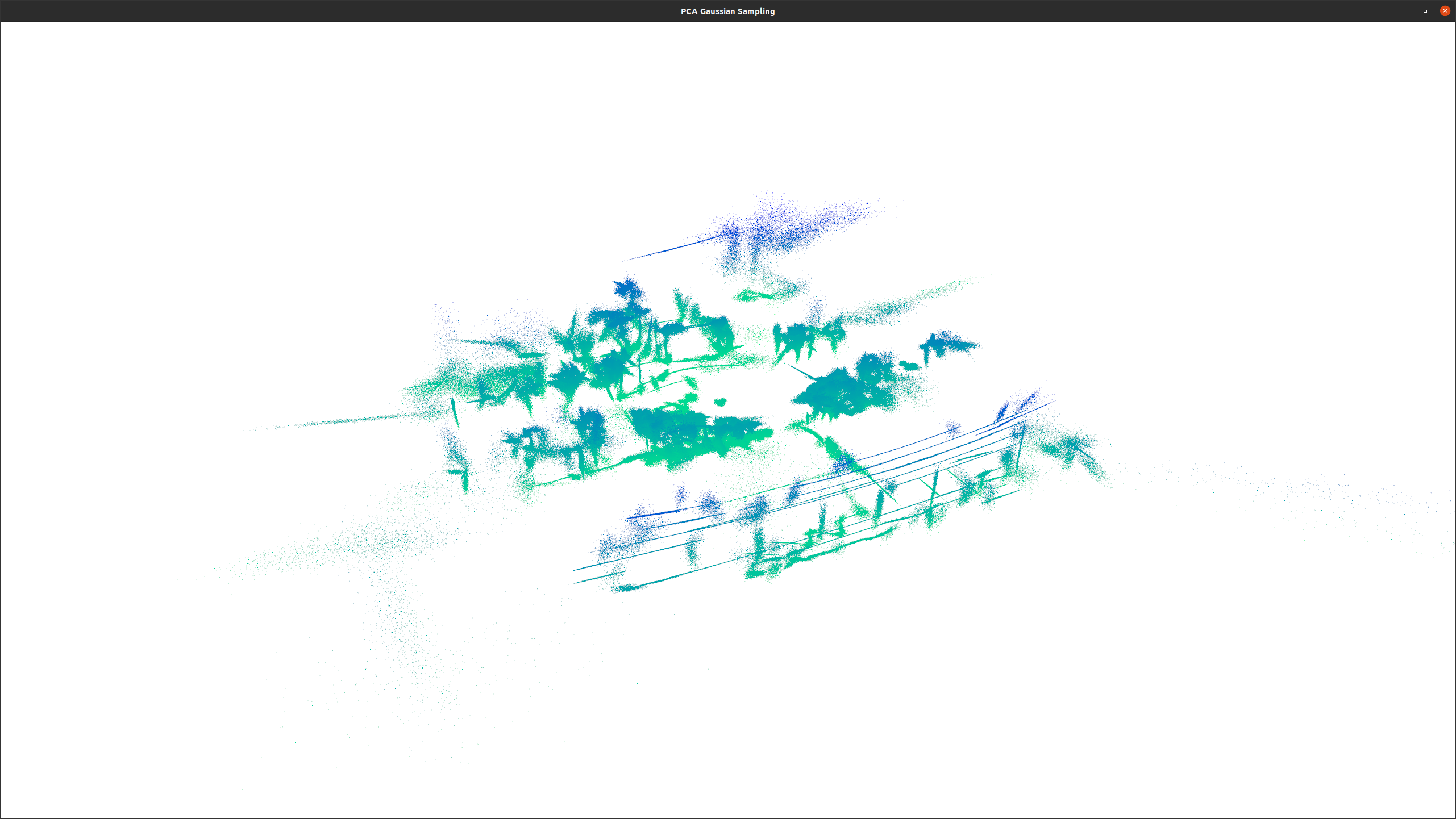}
        {PCA}
    \end{minipage}

    % ---------- Row 2 ----------
    \begin{minipage}[t]{0.45\textwidth}
        \centering
        \includegraphics[width=\linewidth,keepaspectratio,
        trim=700 400 700 360,clip]{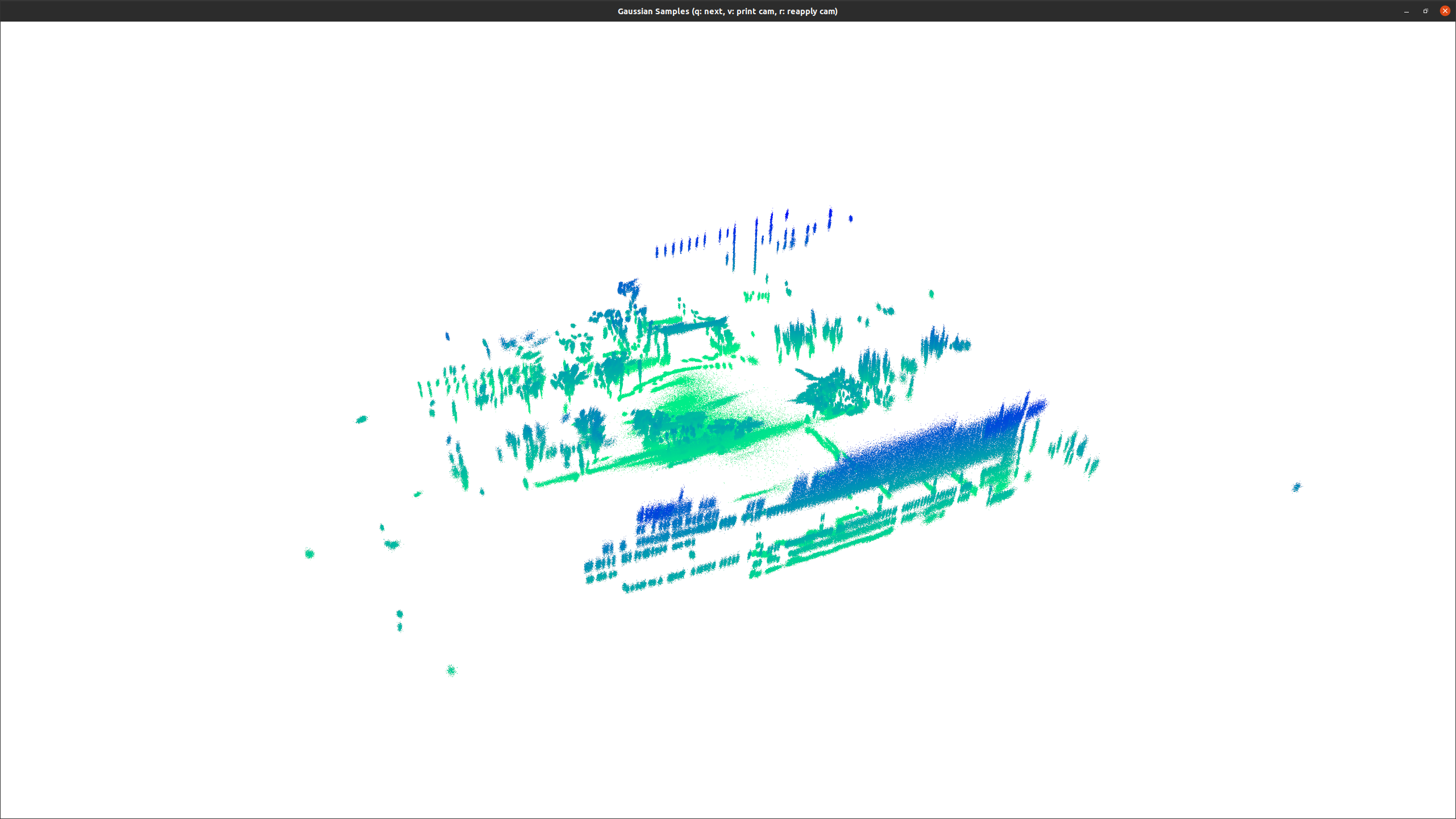}
        {POLI (Proposed)}
    \end{minipage}
    \hspace{0.001\textwidth}
    \begin{minipage}[t]{0.45\textwidth}
        \centering
        \includegraphics[width=\linewidth,keepaspectratio,
        trim=700 400 700 360,clip]{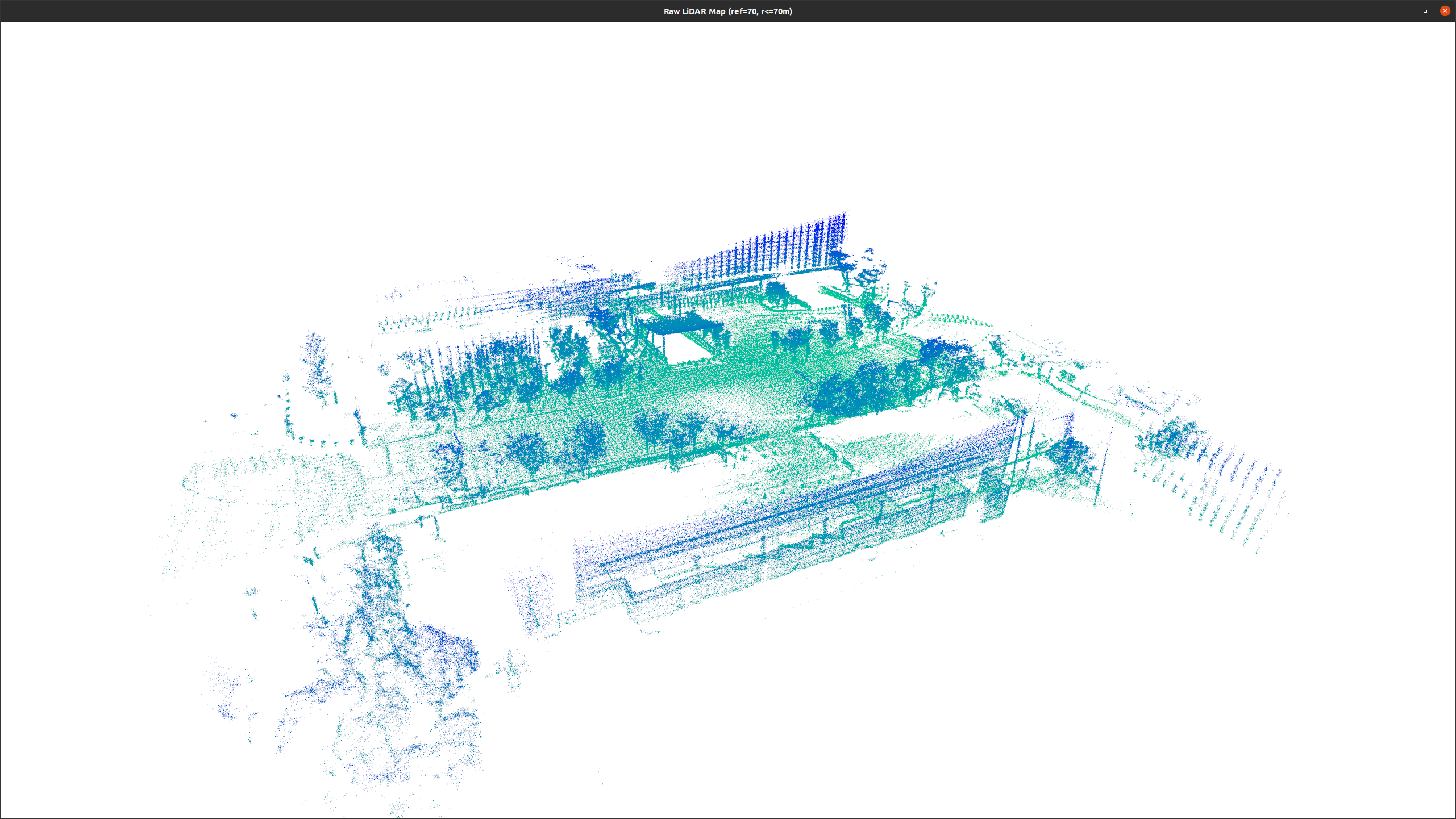}
        { Ground Truth}
    \end{minipage}

   \captionof{figure}{
    Point cloud geometries estimated by different methods on the HeLiPR dataset, visualized with 300 samples per distribution. PCA struggles to recover underlying geometry, whereas POLI preserves geometry aligned with the ground truth.
    }
    \label{fig:front}
    \vspace{-1.0em}
\end{strip}

%%%%%%%%%%%%%%%%%%%%%%%%%%
% Abstract
%%%%%%%%%%%%%%%%%%%%%%%%%%
\begin{abstract}
Point clouds are a fundamental representation for robotic perception tasks such as localization, mapping, and object pose estimation. However, LiDAR-acquired point clouds are inherently sparse and non-uniform, providing incomplete observations of the underlying geometry. Such sparsity and non-uniformity hinder reliable geometric reasoning, leading to degraded performance in downstream perception tasks. 
To mitigate these issues, prior work has attempted to compensate for the sparsity and non-uniformity of point clouds by estimating point cloud geometry. However, in the absence of an explicit model of point cloud geometry, existing approaches have predominantly relied on either hand-crafted statistics of local point distributions or end-to-end supervised deep learning, which often suffer from limited scalability or require large amounts of accurately labeled training data.
To address these challenges, we explicitly model and estimate point cloud geometry under a principled mathematical formulation. Theoretically, we represent the point cloud geometry as a statistical manifold induced by a family of Gaussian distributions that captures the local geometry of each point. Building on this formulation, we design a probabilistic model that predicts per-point local geometry in the form of a Gaussian distribution.
Practically, we introduce a deep neural network to instantiate the estimation of these Gaussian distributions, and term the resulting estimator as Point-to-Ellipsoid (POLI). 
By consistently estimating point-wise local geometry across diverse point clouds, POLI learns a mapping between point cloud observations and their underlying geometry.
Importantly, this mapping is learned in a self-supervised manner, removing the reliance on labeled data while maintaining strong geometric inductive biases. The resulting representation integrates seamlessly into existing robotic perception pipelines without requiring architectural modifications. Extensive experiments demonstrate that the proposed theory and practice enable accurate and robust estimation of point cloud geometry and consistently improve performance across a wide range of robotic perception tasks. The implementation of our proposed method is publicly available at \url{https://github.com/jinwoolee1230/POLI}.
\end{abstract}
%%%%%%%%%%%%%%%%%%%%%%%%%%
% Intro
%%%%%%%%%%%%%%%%%%%%%%%%%%
\IEEEpeerreviewmaketitle
\section{Introduction}

Point clouds are widely used in robotic perception tasks such as localization, mapping, and object pose estimation~\citep{besl1992icp,lim2024quatro++,lim2024kiss,rusu2008pfh, segal2009generalized,tombari2010shot,vizzo2023kiss,zhou2016fgr}. However, point clouds acquired by LiDAR are sparse and non-uniform, providing incomplete observations of the underlying geometry. This limits reliable geometric reasoning and degrades the performance of downstream perception tasks~\citep{kwon2022implicit,qiu2022pu,vizzo2022make,yang2024tulip}. 

To mitigate these limitations, existing robotic perception systems have fundamentally relied on estimating the underlying geometry from point clouds to enrich the available geometric information~\citep{rusu2008pfh,segal2009generalized}. Without an explicit model of point cloud geometry, most systems treat local point distributions as implicit representations of the underlying surface geometry and estimate point-wise local geometry using hand-crafted statistics over local neighborhoods~\citep{hoppe1992surface, lim2024kiss,rusu2008pfh,segal2009generalized,seo2025buffer,tombari2010shot,xu2021fastlio,xu2022fastlio2}. However, in practice, point distributions are strongly influenced by sensor resolution and viewpoint, reflecting observation artifacts rather than underlying geometry.

To overcome such heuristic limitations, end-to-end supervised deep learning approaches have been proposed~\citep{benshabat2020deepfit, benshabat2019nestinet, vizzo2022make, yang2024tulip}. While these methods have shown improved scalability, they delegate most of the geometric reasoning to neural networks, resulting in strong dependence on large-scale annotated data and limited incorporation of geometric inductive biases. Moreover, their representations have limitations in being integrated into robotic downstream tasks, and it is difficult to guarantee consistent geometric representations~\citep{yang2024tulip}. 

In both cases, the lack of a principled model limits the accuracy and scalability of point cloud geometry estimation. By contrast, if the relationship between point clouds and underlying geometry is formulated in a principled mathematical way, accurate and scalable geometry estimation becomes possible, with the potential to benefit a wide range of robotic perception systems. Motivated by this perspective, we propose a principled mathematical formulation for point cloud geometry estimation, together with its theoretical foundations and a practical implementation.

Specifically, we model point cloud geometry as a statistical manifold induced by a family of three-dimensional (3D) Gaussian distributions, where each distribution encodes the local geometry associated with a point. Based on this formulation, we construct a principled probabilistic framework for estimating point-wise covariances and introduce deep learning as a practical instantiation of this framework. This learning framework equips neural networks with strong geometric inductive biases while enabling self-supervised learning without requiring accurate labels. 
The resulting covariance estimator, termed Point-to-Ellipsoid (POLI), learns a mapping between point cloud observations and their underlying geometry by consistently estimating point-wise local geometry across diverse point clouds. 
Moreover, POLI can be seamlessly integrated into existing robotic perception pipelines~\citep{koide2024glim, lim2024kiss, xu2022fastlio2, yang2020teaser} without requiring structural modifications.
%%%%%%%%%%%%%%%%%%%%%%%%%%
% Relatedworks
%%%%%%%%%%%%%%%%%%%%%%%%%%
\section{Point Cloud Geometry Estimation}

Point cloud geometry plays a fundamental role in robotic perception pipelines, as it enables informative representations of point clouds. 
Existing methods for point cloud geometry estimation can be broadly categorized into two classes: hand-crafted neighborhood statistics and deep learning–based approaches. In this section, we review representative methods from both categories and discuss their limitations.

\subsection{Hand-Crafted Neighborhood Statistics}
Hand-crafted neighborhood statistics, methods that approximate the underlying surface geometry using local point distributions within a neighborhood, were originally introduced to improve robotic perception performance. Notably, iterative closest point (ICP)-based registration methods estimate local geometry to construct reliable error metrics. Point-to-plane ICP~\citep{rusinkiewicz2001efficient} estimates a local planar surface from neighboring points, enabling a plane-based residual. GICP~\citep{segal2009generalized} represents local geometry using a covariance matrix obtained from the local point distribution, enabling an adaptive residual metric. VGICP~\citep{koide2021vgicp} aggregates point-wise local statistics within voxel grids, enabling robust local geometry estimation.

In correspondence estimation, local geometry is estimated to obtain distinctive point features. PFH~\citep{rusu2008pfh} and FPFH~\citep{rusu2009fast} encode local geometry using angular relationships between point normals and inter-point directions within a local neighborhood. SHOT~\citep{tombari2010shot} encodes local geometry by accumulating normal orientation distributions within a local reference frame, and RoPS~\citep{guo2013rops} represents local geometry using statistics of local point distributions obtained from rotational projections.

While these statistical approaches have been effective across a wide range of applications, they estimate local geometry directly from observed point neighborhoods using statistical techniques such as principal component analysis (PCA). Consequently, their accuracy depends on the observation process, including sensor resolution and viewpoint. In addition, their performance is sensitive to neighborhood selection, limiting their scalability, as shown in Fig.~\ref{fig:front}.

\subsection{Learning-Based Approaches}

To improve the scalability of conventional methods, learning-based approaches have been proposed. Many existing works focus on point-wise surface normal estimation~\citep{lenssen2020deep, li2025learning, li2025high}, which provides a compact description of local geometry. However, normals alone fail to capture geometric properties such as local planar extent or curvature, limiting their utility in downstream tasks.

Another line of work aims to densify point clouds or reconstruct complete surface geometry using generative models~\citep{kwon2022implicit,nunes2024lidar_diffusion,qiu2022pu,vizzo2022make,yang2024tulip,zhou20213d}. While these methods can produce visually dense reconstructions, predicting geometry in unobserved regions often leads to inaccurate and inconsistent estimates. Moreover, they typically require dense scans for training, which are difficult to obtain in practice, and can also incur high computational cost. More discussion is provided in Supplementary Material~SF.

Despite their empirical success, many learning-based approaches rely on fully end-to-end training without explicitly modeling the underlying geometric structure. As a result, geometric reasoning is implicitly delegated to neural networks, leading to a strong dependence on large amounts of carefully annotated training data and fundamental limitations in accuracy and scalability. From this perspective, we introduce a new problem formulation that models point cloud geometry as a statistical manifold, together with a solid theoretical foundation and a practical instantiation via neural networks.
Crucially, our formulation enables self-supervised learning, requiring neither dense scans nor perfect transformation labels, thereby substantially expanding the range of datasets that can be leveraged for geometry estimation.
%%%%%%%%%%%%%%%%%%%%%%%%%%
% Problem
%%%%%%%%%%%%%%%%%%%%%%%%%%
\section{Theory: Modeling Point Cloud Geometry}
\label{sec:probstate}
In this section, we present a theoretical formulation that models point cloud geometry as a statistical manifold and derives a probabilistic framework for its estimation.

\subsection{Point Cloud Geometry as a Statistical Manifold}
A statistical manifold is a smooth manifold formed by a family of probability distributions parameterized by a continuous parameter space~\citep{amari2000methods}.
Since each element on a statistical manifold represents a probability distribution rather than a single point measurement, modeling local surface geometry with individual distributions provides an expressive representation of the underlying surface while reducing the reliance on dense observations. Building on this idea, we model point cloud geometry as a statistical manifold, denoted by $\mathcal{M}_\mathbf{g}$, induced by Gaussian distributions that encode local surface geometry.
Beyond modeling, we formulate the corresponding estimation problem and provide a practical instantiation by adopting Gaussian distributions, which are analytically tractable and well-suited for compact geometric representation.

\begin{definition}[Statistical Manifold of World Geometry]
The statistical manifold $\mathcal{M}_{\mathbf{g}}$ represents the underlying world surface manifold through a probabilistic approximation in a global reference frame. The manifold $\mathcal{M}_\mathbf{g}$ is defined as a family of 3D Gaussian distributions, each parameterized by a mean $\mathbf{\bar{x}}_{\mathbf{g}_i} \in \mathbb{R}^3$ and a covariance matrix $\mathbf{C}_{\mathbf{g}_i} \in \mathbb{S}_{+}^{3}$.
Each Gaussian distribution defines the probability density that an observed point $\mathbf{x}_i$ is measured with respect to a local tangent plane at $\mathbf{\bar{x}}_{\mathbf{g}_i}$.
\end{definition}

\begin{figure}[t]
    \centering
    \includegraphics[width=0.9\linewidth]{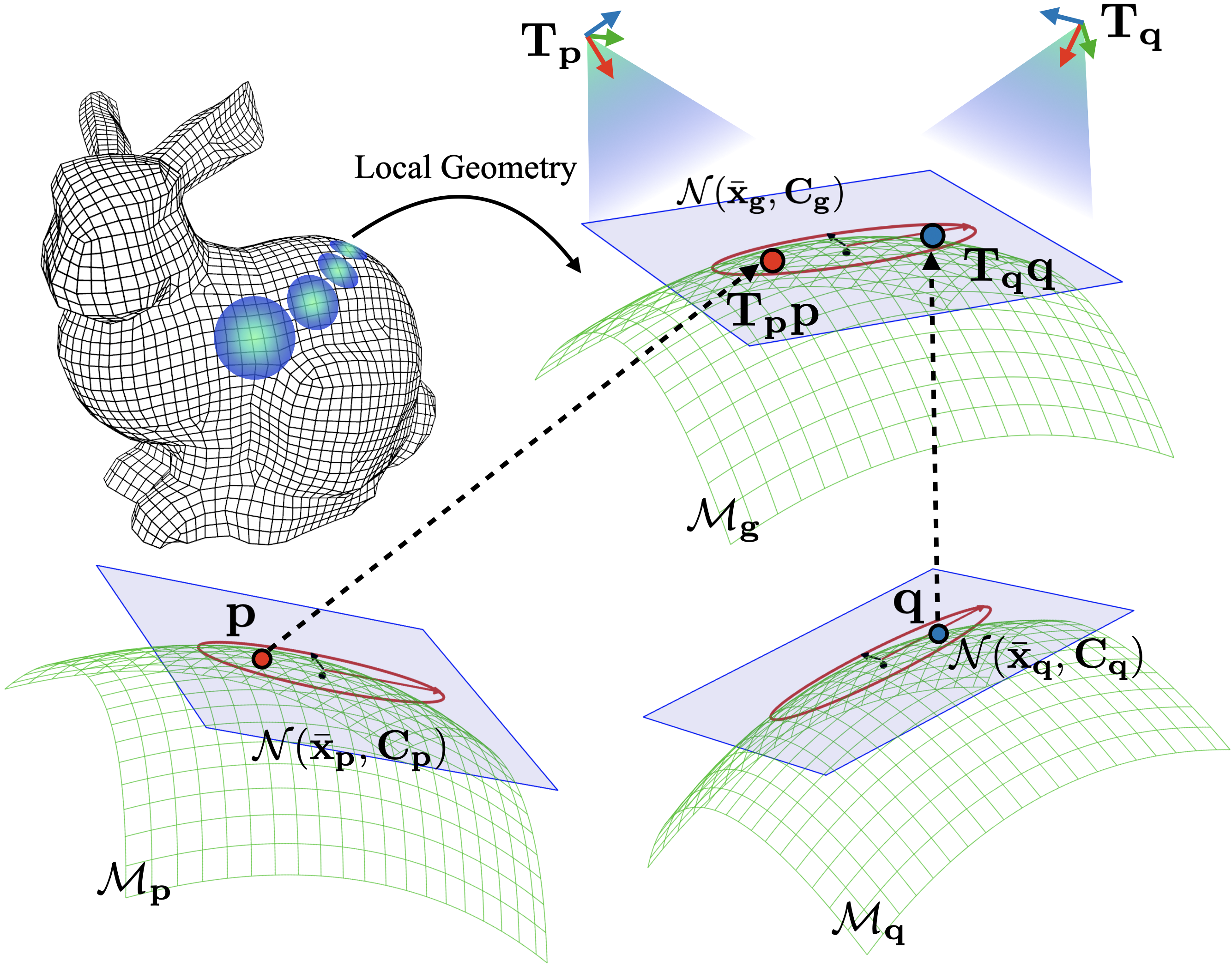}
    \caption{
    A visualization illustrating how underlying geometry can be represented as a statistical manifold.
    It explains that two points observed in different coordinate frames can be regarded as samples drawn from the same underlying distribution.
    }
    \label{fig:pdf_first_page}
    \vspace{-0.7cm}
\end{figure}

When $\mathbf{x}_i$ is acquired from a high-precision LiDAR sensor, the resulting Gaussian distribution approximates the local planar geometry as shown in Fig.~\ref{fig:pdf_first_page}. In this formulation, the covariance matrix $\mathbf{C}_{\mathbf{g}_i}$ provides a quantitative representation of the geometry~\citep{pauly2003multi} in the neighborhood of $\mathbf{\bar{x}}_{\mathbf{g}_i}$.

\begin{proposition}[Statistical Manifold Representation from Different Sensor Frames]
Given the sensor pose $\mathbf{T_p} = (\mathbf{R_p}, \mathbf{t_p}) \in SE(3)$ relative to the global reference frame, the parameters $\mathbf{\bar{x}}_{\mathbf{p}_i}$ and $\mathbf{C}_{\mathbf{p}_i}$ of the local statistical manifold $\mathcal{M}_\mathbf{p}$ are defined by transforming the global model $\mathcal{M}_\mathbf{g}$ into the sensor frame via $\mathbf{T_p}^{-1}$:

\begin{equation*}
    \mathbf{\bar{x}}_{\mathbf{p}_i} \triangleq \mathbf{R}_{\mathbf{p}}^\top (\mathbf{\bar{x}}_{\mathbf{g}_i} - \mathbf{t_p}), \qquad
    \mathbf{C}_{\mathbf{p}_i} \triangleq \mathbf{R}_{\mathbf{p}}^\top\mathbf{C}_{\mathbf{g}_i}\mathbf{R}_{\mathbf{p}}.
\end{equation*}
Analogously, for a second observation frame with pose $\mathbf{T_q} = (\mathbf{R_q}, \mathbf{t_q}) \in SE(3)$, the parameters $\mathbf{\bar{x}}_{\mathbf{q}_i}$ and $\mathbf{C}_{\mathbf{q}_i}$ of the statistical manifold $\mathcal{M}_\mathbf{q}$ are defined via $\mathbf{T_q}^{-1}$:
\begin{equation*}
    \mathbf{\bar{x}}_{\mathbf{q}_i} \triangleq \mathbf{R_q}^\top (\mathbf{\bar{x}}_{\mathbf{g}_i} - \mathbf{t_q}), \qquad
    \mathbf{C}_{\mathbf{q}_i} \triangleq \mathbf{R_q}^\top \mathbf{C}_{\mathbf{g}_i}\mathbf{R_q}.
\end{equation*}
\end{proposition}

% Let $\mathcal{P}\triangleq(\mathbf{p}_1,\dots,\mathbf{p}_N)$ and $\mathcal{Q}\triangleq(\mathbf{q}_1,\dots,\mathbf{q}_N)$ denote two scans acquired from frames $\mathbf{T}_p$ and $\mathbf{T}_q$, respectively, with known correspondences $(\mathbf{p}_i,\mathbf{q}_i)$. We denote the associated per-point covariances by $\mathbf{C}_p \triangleq (\mathbf{C}_{\mathbf{p}_1},\dots,\mathbf{C}_{\mathbf{p}_N})$ and $\mathbf{C_q} \triangleq (\mathbf{C}_{\mathbf{q}_1},\dots,\mathbf{C}_{\mathbf{q}_N})$. Although sampled independently, $\mathbf{p}$ and $\mathbf{q}$ originate from manifolds that are related by an isometric rigid-body transformation of the same underlying world surface, and therefore share identical local geometry (i.e., $\mathbf{C}_{\mathbf{g}_i}$). Under this local geometric model, the displacement $\mathbf{d}_i$ for the $i$-th pair is defined by the relative transformation $\mathbf{T}\triangleq \mathbf{T}_q^{-1}\mathbf{T}_p$ as:
Let $\mathcal{P}\triangleq(\mathbf{p}_1,\dots,\mathbf{p}_N)$ and $\mathcal{Q}\triangleq(\mathbf{q}_1,\dots,\mathbf{q}_N)$ denote two scans acquired from frames $\mathbf{T}_\mathbf{p}$ and $\mathbf{T}_\mathbf{q}$, respectively, with known correspondences $(\mathbf{p}_i,\mathbf{q}_i)$. Each point is independently sampled as a noisy observation from its local Gaussian distribution, i.e., $\mathbf{p}_i \sim \mathcal{N}(\mathbf{\bar{x}}_{\mathbf{p}_i}, \mathbf{C}_{\mathbf{p}_i})$ and $\mathbf{q}_i \sim \mathcal{N}(\mathbf{\bar{x}}_{\mathbf{q}_i}, \mathbf{C}_{\mathbf{q}_i})$, with per-point covariances $\mathbf{C}_\mathbf{p} \triangleq (\mathbf{C}_{\mathbf{p}_1},\dots,\mathbf{C}_{\mathbf{p}_N})$ and $\mathbf{C_q} \triangleq (\mathbf{C}_{\mathbf{q}_1},\dots,\mathbf{C}_{\mathbf{q}_N})$. Note that, $\mathcal{P}$ and $\mathcal{Q}$ originate from manifolds related by an isometric rigid-body transformation of the same underlying world surface, they share identical local geometry (i.e., $\mathbf{C}_{\mathbf{g}_i}$). Under this local geometric model, the displacement $\mathbf{d}_i$ for the $i$-th pair is defined by the relative transformation $\mathbf{T}\triangleq \mathbf{T}_\mathbf{q}^{-1}\mathbf{T}_\mathbf{p}$ as:
\begin{equation}
\label{eq:residual}
\mathbf{d}_i \triangleq \mathbf{q}_i - \mathbf{T}\mathbf{p}_i,
\qquad
\mathbf{d}_i \sim \mathcal{N}\!\left(\mathbf{0},\,2\mathbf{C}_{\mathbf{q}_i}\right).
\end{equation}
Derivation of Eq.~\eqref{eq:residual} is provided in Supplementary Material~SA.

\begin{lemma}[Re-parameterization]
\label{lem:reparam}
Given the relative transformation $\mathbf{T}$, the conditional likelihood of a target point $\mathbf{q}_i$ given its source point $\mathbf{p}_i$ depends only on the residual in Eq.~\eqref{eq:residual}:
\begin{align}
p(\mathbf{q}_i \mid \mathbf{p}_i,\mathbf{T},\mathbf{C}_{\mathbf{q}_i})
&\equiv
p(\mathbf{d}_i \mid \mathbf{T},\mathbf{C}_{\mathbf{q}_i}).
\label{eq:reparam}
\end{align}
This re-parameterization is justified by an invertible change of variables with unit Jacobian. Under the local geometric model in Eq.~\eqref{eq:residual}, we obtain:
\begin{equation*}
\label{eq:qi_conditional}
\begin{aligned}
p(\mathbf{q}_i \!\mid\! \mathbf{p}_i,\mathbf{T},\mathbf{C}_{\mathbf{q}_i})\!
&=\!
\frac{1}{\sqrt{(2\pi)^3\,\lvert 2\mathbf{C}_{\mathbf{q}_i}\rvert}}
\exp\!\left(
-\tfrac{1}{2}\,\mathbf{d}_i^\top (2\mathbf{C}_{\mathbf{q}_i})^{-1}\mathbf{d}_i
\right).
\end{aligned}
\end{equation*}
\end{lemma}
Derivation of Eq.~\eqref{eq:reparam} is provided in Supplementary Material~SA.

\begin{proposition}[Factorized Correspondence Likelihood]
\label{prop:correspondence}
The conditional likelihood of the target set $\mathcal{Q}$ given the fixed source set $\mathcal{P}$ factorizes as:
\begin{equation}
\begin{aligned}
p(\mathcal{Q} \mid &\mathcal{P},\mathbf{T},\mathbf{C_q}) 
=
\prod_{i=1}^{N}p(\mathbf{q}_i \mid \mathbf{p}_i,\mathbf{T},\mathbf{C}_{\mathbf{q}_i}) \\
=
\prod_{i=1}^{N}&\frac{1}{\sqrt{(2\pi)^3\,|2\mathbf{C}_{\mathbf{q}_i}|}}
\exp\!\left(
-\frac{1}{2}\,\mathbf{d}_i^\top(2\mathbf{C}_{\mathbf{q}_i})^{-1}\mathbf{d}_i
\right),
\end{aligned}
\label{eq:correspondence}
\end{equation}
where the last equality follows from Lemma~\ref{lem:reparam}.
\end{proposition}

\subsection{Probabilistic Modeling for Covariance Estimation}\
Under a probabilistic model that relates two scans acquired from the same underlying scene, their relative pose, and the local geometry, we derive a maximum likelihood estimation (MLE) framework. Within this framework, we formulate an optimization problem to estimate the underlying local geometry in terms of point-wise covariances.\\

\begin{figure}[t]
    \centering
    \includegraphics[width=\linewidth, trim= 0 400 0 400, clip]{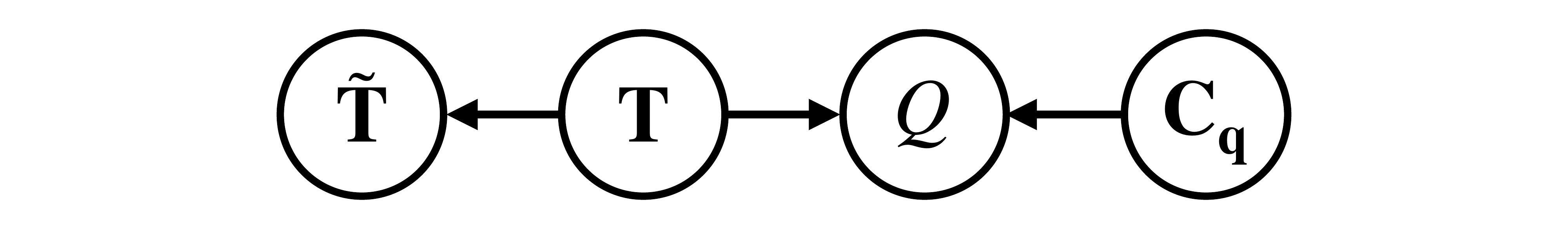}
    \caption{Probabilistic model as a directed acyclic graph, where edges denote direct conditional dependencies.}
    \label{fig:main}
\end{figure}

\begin{assumption}[Probabilistic Modeling]
\label{asum:prob_model}
We aim to estimate the covariances $\mathbf{C}_\mathbf{q}$ from observations $\mathcal{D} = (\mathcal{Q}, \tilde{\mathbf{T}})$, conditioned on the source set $\mathcal{P}$.
Here, $\tilde{\mathbf{T}}$ denotes a relative pose measurement between the sensor frames at which the scans are acquired, representing an observation of the underlying transformation $\mathbf{T}$ (e.g., a relative pose measured by a GNSS–INS system). We write the marginal likelihood as:
\begin{align}
\!p(\mathcal{D}\!\mid\! \mathcal{P},\mathbf{C_q})\!
&=
\int p(\mathcal{D},\mathbf{T}\mid \mathcal{P},\mathbf{C_q})\, d\mathbf{T}
\notag\\
& \textcolor{gray}{\text{(marginalization)}}
\notag\\
\!\!=
\int
p(\mathcal{Q}\mid &\mathbf{\tilde T},\mathcal{P},\mathbf{T},\mathbf{C_q})\;
p(\mathbf{\tilde T}\mid \mathbf{T},\mathcal{P},\mathbf{C_q})\;
p(\mathbf{T}\mid\mathcal{P},\mathbf{C_q})\;
d\mathbf{T}
\notag\\
& \textcolor{gray}{\text{(chain rule)}}
\notag\\
=\int
p(\mathcal{Q}\mid &\mathcal{P},\mathbf{T},\mathbf{C_q})\;
p(\mathbf{\tilde T}\mid \mathbf{T})\;
p(\mathbf{T})\;
d\mathbf{T}
\notag\\
&\textcolor{gray}{(\text{conditional independence})}
\notag\\
\propto \int p(\mathcal{Q}\mid& \mathcal{P},\mathbf{T},\mathbf{C_q}) p(\mathbf{\tilde T}\mid \mathbf{T})
\, d\mathbf{T}.\label{eq:prob_model_marginal}
\\
&\textcolor{gray}{\text{(uniform prior for } p(\mathbf{T})\text{)}}
\notag
\end{align}
We model the relationships among the variables as follows.
The latent pose $\mathbf{T}$ is assumed to be independent of $(\mathcal{P}, \mathbf{C}_\mathbf{q})$, and we adopt a non-informative prior
$p(\mathbf{T}) \propto 1$.
The observed pose $\tilde{\mathbf{T}}$ depends solely on the latent pose $\mathbf{T}$, and is conditionally independent of $\mathcal{Q}$
given $(\mathcal{P}, \mathbf{T}, \mathbf{C}_\mathbf{q})$. Intuitively, once $\mathcal{P}$ is given, the latent pose $\mathbf{T}$---representing the relative transformation between $\mathcal{P}$ and $\mathcal{Q}$---serves as the primary factor driving both $\tilde{\mathbf{T}}$ and $\mathcal{Q}$. Specifically, $\tilde{\mathbf{T}}$ is generated directly from a realization of $\mathbf{T}$, while $\mathcal{Q}$ is determined
by the combination of $\mathbf{T}$ and the intrinsic factor $\mathbf{C}_\mathbf{q}$. These dependency relationships are illustrated in Fig.~\ref{fig:main}. We assume that any potential dependencies beyond those explicitly modeled are negligible, allowing us to focus on the key dependencies in our formulation. Further discussion is provided in Supplementary Material~SE.
\end{assumption}

\begin{definition}[Energy Function]
\label{def:energy}
To simplify the expression of $p(\mathcal{D}\mid\mathcal{P}, \mathbf{C}_\mathbf{q})$ and facilitate the derivation of its MLE, we derive the energy function of the integrand in Eq.~\ref{eq:prob_model_marginal}, i.e., the negative log-probability of the corresponding distribution. For the pose likelihood (i.e., $p(\mathbf{\tilde T}\mid \mathbf{T})$), we assume a Gaussian noise model:
\begin{align}
p(\mathbf{\tilde T} \mid \mathbf{T})
&=
\frac{1}{(2\pi)^{3}\,|\Gamma|^{1/2}}
\exp\left(
-\frac{1}{2}\,\xi(\mathbf{T})^\top \Gamma^{-1}\,\xi(\mathbf{T})
\right),
\label{eq:pose_likelihood}
\end{align}
where $\xi(\mathbf{T})=\mathrm{Log}\!\left(\mathbf{T}^{-1}\mathbf{\tilde T}\right)\in\mathbb{R}^6$ and $\Gamma$ denotes the covariance of the measurement noise. Combining Eqs.~(\ref{eq:correspondence}) and (\ref{eq:pose_likelihood}), we define the energy:
\begin{align}
\Phi(\mathbf{T};\mathbf{C_q})
&\triangleq
-\log\!\Big(
p(\mathcal{Q}\mid \mathcal{P},\mathbf{T},\mathbf{C_q})\;
p(\mathbf{\tilde T}\mid \mathbf{T})
\Big)\notag
\\
&=
\frac{1}{2}\sum_{i=1}^{N}
\Big[
\log|2\mathbf{C_q}_{i}|
+
\mathbf{d}_{i}^\top
(2\mathbf{C_q}_{i})^{-1}
\mathbf{d}_{i}
\Big]\notag
\\
&+
\frac{1}{2}\log|\Gamma|
+
\frac{1}{2}\,\xi(\mathbf{T})^\top
\Gamma^{-1}
\xi(\mathbf{T})
+
\text{const.}
\label{eq:original_energy_simplified}
\end{align}
Then, the marginal likelihood from Assumption~\ref{asum:prob_model} can be written as:
\begin{equation}
\begin{aligned}
p(\mathcal{D}\mid \mathcal{P},\mathbf{C_q})
&\propto
\int
p(\mathcal{Q}\mid \mathcal{P},\mathbf{T},\mathbf{C_q})\;
p(\mathbf{\tilde T}\mid \mathbf{T})\;
d\mathbf{T}
\\
&\propto
\int \exp\!\big(-\Phi(\mathbf{T};\mathbf{C_q})\big)\, d\mathbf{T}.
\end{aligned}
\label{eq:marginal_energy}
\end{equation}
\end{definition}

\begin{proposition}[Laplace Approximation]
\label{prop:laplace}
The integral in Eq.~\eqref{eq:marginal_energy} is generally intractable due to the nonlinearity of $\Phi(\mathbf{T};\mathbf{C}_\mathbf{q})$ over $SE(3)$.
We therefore adopt a Laplace approximation, which approximates the integral by fitting a Gaussian to the integrand around its mode (most probable pose $\hat{\mathbf{T}}$). This yields a closed-form, differentiable surrogate objective with stable gradients, avoiding the computational overhead of sampling-based alternatives~\citep{mackay2003information}. Since the integrand is proportional to $\exp\!\big(-\Phi(\mathbf{T};\mathbf{C}_\mathbf{q})\big)$, the dominant contribution comes from the neighborhood of the minimizer of $\Phi$.
Accordingly, we first compute the mode pose
\begin{equation}
\hat{\mathbf{T}} \triangleq \arg\min_{\mathbf{T}} \Phi(\mathbf{T};\mathbf{C}_{\mathbf{q}}),
\label{eq:optimal_transform}
\end{equation}
and apply a second-order Taylor expansion of $\Phi$ around $\hat{\mathbf{T}}$
\begin{equation}
p(\mathcal{D}\mid \mathcal{P},\mathbf{C}_{\mathbf{q}})
\approx
K_0\,(2\pi)^3\,\big|H(\hat{\mathbf{T}})\big|^{-1/2}\,
\exp\!\big(-\Phi(\hat{\mathbf{T}};\mathbf{C}_{\mathbf{q}})\big),
\label{eq:laplace}
\end{equation}
where $K_0$ is constant and $H(\hat{\mathbf{T}})\in\mathbb{R}^{6\times 6}$ denotes the Hessian of $\Phi$ with respect to a $SE(3)$ perturbation, evaluated at $\hat{\mathbf{T}}$:
\begin{equation}
H(\hat{\mathbf{T}})
\triangleq
\left.\nabla_{\mathbf{T}}^{2}\Phi(\mathbf{T};\mathbf{C}_{\mathbf{q}})\right|_{\mathbf{T}=\hat{\mathbf{T}}}.
\label{eq:hessian}
\end{equation}
\end{proposition}

\begin{corollary}[Geometry Estimation]
\label{cor:mle}
Consequently, the MLE for the covariances is obtained by maximizing the approximate marginal likelihood, equivalently
\begin{align}
\hat{\mathbf{C}}_q
&=
\arg\max_{\mathbf{C_q}}\;
p(\mathcal{D}\mid \mathcal{P},\mathbf{C_q})
\\
&\approx
\arg\min_{\mathbf{C_q}}
\frac{1}{2}\log|H(\hat{\mathbf{T}})|
+
\Phi(\hat{\mathbf{T}};\mathbf{C_q}).
\label{eq:map_approximation}
\end{align}
\end{corollary}

With this probabilistic formulation, we estimate point cloud geometry in the form of point-wise covariances $\mathbf{C}_{\mathbf{q}}$, which are elements of the statistical manifold representing the underlying geometry.
This probabilistic framework only requires reasonably accurate relative pose measurements $\tilde{\mathbf{T}}$ and consistent correspondences $(\mathbf{p}_i,\mathbf{q}_i)$, enabling self-supervised learning without relying on dense point clouds or explicit ground-truth geometry in the following section.
%%%%%%%%%%%%%%%%%%%%%%%%%%
% Practive
%%%%%%%%%%%%%%%%%%%%%%%%%%
\section{Practice: Learning Point Cloud Geometry}
\label{method}
In this section, we describe how to solve the optimization problem in Corollary~\ref{cor:mle}. The covariance set to be optimized, $\mathbf{C}_{\mathbf{q}} \in \mathbb{R}^{6 \times N}$, which comprises six parameters per covariance across $N$ correspondences, requires a high-dimensional, nonconvex optimization problem that is impractical to solve using classical optimization methods. Therefore, we reformulate the problem as learning a mapping from the input point cloud $\mathcal{Q}$ to the set of covariance matrix $\mathbf{C}_{\mathbf{q}}$. Specifically, we train a neural network to predict $\mathbf{C}_{\mathbf{q}}$ such that the objective in Eq.~\eqref{eq:map_approximation} is minimized. The remainder of this section is organized as follows: We first present a neural network architecture of a covariance estimator that predicts point-wise covariances, then introduce a certifiably correct and differentiable optimization layer for the mode pose estimation and gradient calculation, and finally summarize the overall training and inference procedure. 
Through this formulation, we enable the network to infer the underlying geometry of the point cloud input as a set of point-wise covariances that constitute the statistical manifold.

\subsection{Covariance Estimator}
\begin{figure}[t]
    \centering
    \includegraphics[width=0.98\linewidth, trim= 20 30 20 0, clip]{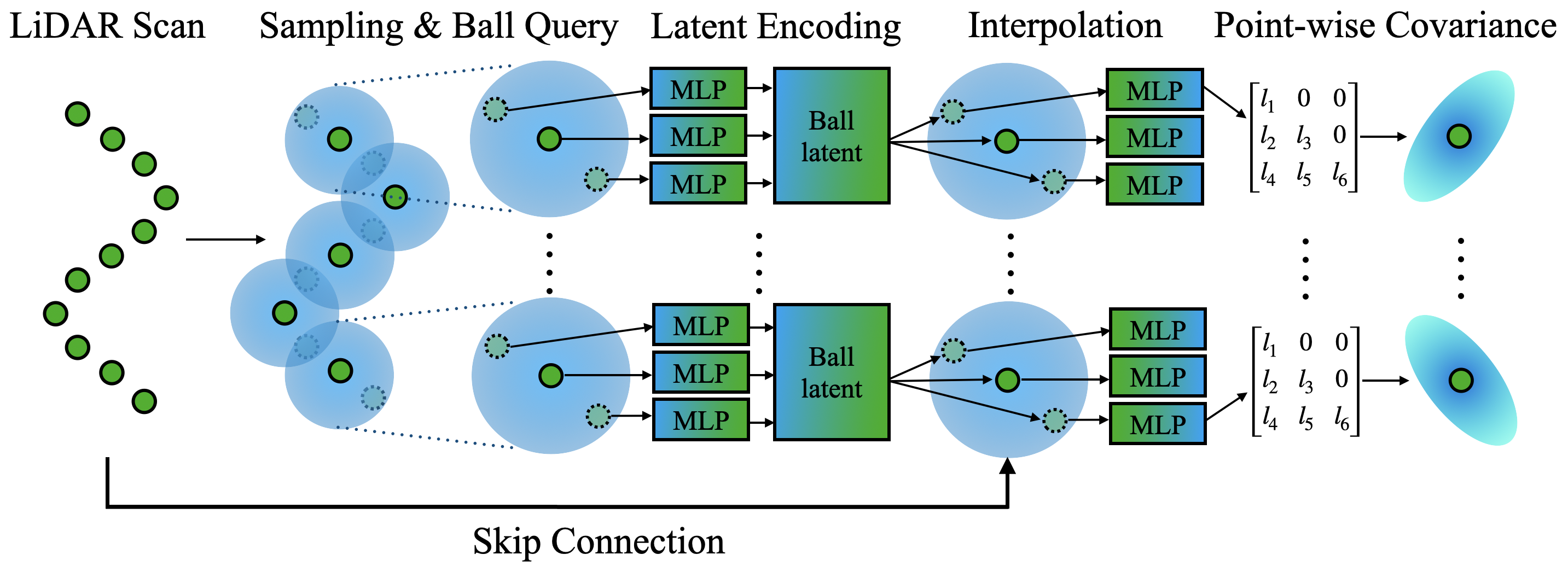}
    \caption{Network architecture of covariance estimator.}
    \label{fig:network_architecture}
    \vspace{-0.5cm}
\end{figure}

A neural network that learns to predict point-wise covariance must be permutation invariant, extract local features in the underlying geometry, and handle variable-sized and non-uniform point sets. Therefore, we design a neural network inspired by PointNet++~\citep{qi2017pointnet++} to predict a per-point covariance matrix. Given a point cloud, a subset of points is selected via farthest point sampling, and neighborhoods are queried around them using a ball query. Each neighborhood is normalized to its centroid frame and encoded by applying a shared multi layer perceptron (MLP) to each point individually, followed by max pooling to obtain a fixed-length representation of the region. The resulting local features are back-propagated to the original point cloud using distance-weighted interpolation and skip connections, producing point-wise feature representations.
For each point \(\mathbf{q}_i\), the network predicts six parameters that define a lower-triangular matrix \(\mathbf{L}_i\). 
The covariance is then constructed as \(\mathbf{C_q}_{i}=\mathbf{L}_i{\mathbf{L}_i\!}^\top\), which guarantees \(\mathbf{C_q}_{i}\succeq 0\). Network architecture is illustrated in Fig.~\ref{fig:network_architecture}

\subsection{Certifiably Correct Differentiable Optimization Layer}
Given network-predicted covariances $\mathbf{C_q}$, we should estimate the mode pose in Eq.~\eqref{eq:optimal_transform}:
\begin{align}
\hat{\mathbf{T}}
&= \arg\min_{\mathbf{T}}\sum_{i=1}^N \mathbf{d}^\top_{i} {\bigl(2\mathbf{C_q}_{i}\bigr)\!}^{-1}\mathbf{d}_{i}+
\xi(\mathbf{T})\!^\top \Gamma^{-1}\xi(\mathbf{T}).
\label{eq:map_pose_general}
\end{align} 
However, to the best of our knowledge, solving Eq.~\eqref{eq:map_pose_general} is challenging due to the nonlinearity introduced by $SE(3)$ and the anisotropy of the weighting factor~\citep{holmes2024semidefinite, holmes2025sdprlayers}. Moreover, even when using iterative solvers, convergence to the true optimum is not guaranteed, so gradients obtained by implicit function theorem (IFT)~\citep{blondel2022efficient} may be biased toward suboptimal solutions~\citep{holmes2025sdprlayers}. 
Accordingly, we relax Eq.~\eqref{eq:map_pose_general} to:
\begin{align}
\hat{\mathbf{T}}
\;\approx\;
& \arg\min_{\mathbf{T}}\sum_{i=1}^N \mathbf{d}^\top_{i} \bigl(2\mathbf{C_q}_{i}\bigr)^{-1}\mathbf{d}_{i}.
\label{eq:reduced_estimator}
\end{align}
This relaxation is well suited to our training setup, where the correspondence set $\mathcal{C}\triangleq(\mathcal{P},\mathcal{Q})$ is constructed under a reference pose $\tilde{\mathbf{T}}$ by selecting nearest-neighbor pairs between $\mathcal{P}$ and $\mathcal{Q}$ and rejecting pairs whose distances exceed a predefined threshold.
As a result, information about $\tilde{\mathbf{T}}$ is implicitly embedded in $\mathcal{C}$ and continues to guide the optimization even after the relaxation. Although this relaxation does not exactly solve the original problem, it enables us to leverage SDPRLayer~\citep{holmes2025sdprlayers}. 

SDPRLayer~\citep{holmes2025sdprlayers} provides certifiably globally optimal solutions, supports differentiation through the optimizer (allowing gradients to be computed at the optimum), and is computationally efficient. By making the inner pose estimation step both certifiably optimal and differentiable, we can train (i.e., backpropagate) the entire network end-to-end on large-scale datasets in a self-supervised manner. Details on the certifiable global optimization and differentiation are provided in Supplementary Material~SB, while the relaxation leading to Eq.~\eqref{eq:reduced_estimator} is described in Supplementary Material~SC.

\subsection{Overall Training and Inference Procedure}
\begin{figure}[t]
\centering

\begin{minipage}{\linewidth}
{\small
\begin{algorithm}[H]
\caption{Training Stage of POLI (\textbf{PO}int-to-el\textbf{LI}psoid)}
\label{alg:training}
\begin{algorithmic}[1]
\Require Neural Network $\texttt{NN}(\cdot;\mathbf{\Theta})$, $\mathcal{C}$, $\tilde{\mathbf T}$, Learning Rate $\eta$, point cloud $\mathcal{P}$ and point cloud $\mathcal{Q}$

\Ensure Updated parameters $\mathbf{\Theta}$
\State {\textcolor{Blue}{/* Predict Lower-Triangular Matrices */}}
\State $\mathbf{L}\!\leftarrow\!\texttt{NN}(\mathcal{Q};\mathbf{\Theta})$
\State {\textcolor{Blue}{/* Construct Covariances */}}
\State $\mathbf{C_q}\!\leftarrow\!\mathbf{L}\mathbf{L}^\top$
\State {\textcolor{Blue}{/* Certifiably Correct Differentiable Optimization */}}
\State $\hat{\mathbf{T}},\frac{\mathbf{d}\mathbf{\hat{T}}}{\mathbf{d}\mathbf{C_q}}\leftarrow\texttt{SDPRLayer}(\mathbf{C_q},\mathcal{C}, \mathcal{P}, \mathcal{Q})$
\State {\textcolor{Blue}{/* Loss Calculation */}}
\State $\displaystyle
\mathcal{L}(\mathbf{\Theta})
= \Phi(\hat{\mathbf T};\mathbf C_q)
+\tfrac{1}{2}\log|H(\hat{\mathbf T})|$
\State {\textcolor{Blue}{/* Neural Network Update */}}
\State $
\mathbf{\Theta} \leftarrow \mathbf{\Theta} - \eta\nabla_{\mathbf{\Theta}} \mathcal{L}
$
\end{algorithmic}
\end{algorithm}
}
\end{minipage}

\begin{minipage}{\linewidth}
{\small
\begin{algorithm}[H]
\caption{Inference Stage of POLI}
\label{alg:inference}
\begin{algorithmic}[1]
\Require Trained Neural Network $\texttt{NN}(\cdot;\mathbf{\Theta}^{*})$, point cloud $\mathcal{Q}$

\Ensure Estimated covariance $\mathbf{C_q}$

\State {\textcolor{Blue}{/* Predict Lower-Triangular Matrices */}}
\State $\mathbf{L}\!\leftarrow\!\texttt{NN}(\mathcal{Q};\mathbf{\Theta}^{*})$

\State {\textcolor{Blue}{/* Construct Covariances */}}
\State $\mathbf{C_q}\!\leftarrow\!\mathbf{L}\mathbf{L}^{\top}$

\State \Return $\mathbf{C_q}$
\end{algorithmic}
\end{algorithm}
}
\end{minipage}

\vspace{0.5em}

\includegraphics[width=\linewidth, trim=150 70 200 100, clip]{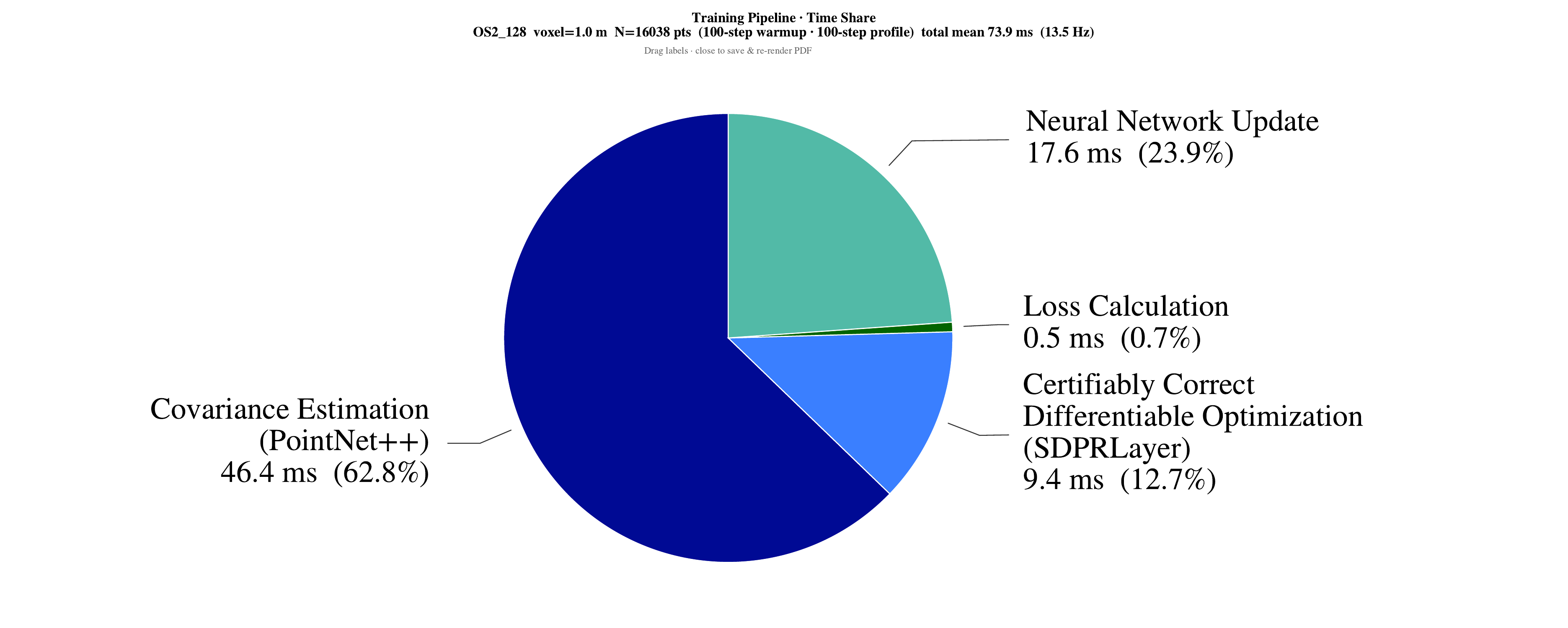}
\captionof{figure}{Runtime composition of POLI training pipeline.}
\label{fig:train_time}
\vspace{-0.5cm}
\end{figure}

Algorithms~\ref{alg:training} and~\ref{alg:inference} summarize the
training and inference procedures of POLI, respectively. During training,
the neural network predicts a point-wise lower-triangular matrix
$\mathbf{L}$ from the target point cloud $\mathcal{Q}$, and the covariance
set is constructed as $\mathbf{C_q}=\mathbf{L}\mathbf{L}^{\top}$. Given
$\mathbf{C_q}$ and the correspondence set $\mathcal{C}$,
SDPRLayer~\citep{holmes2025sdprlayers} estimates the relative
transformation $\hat{\mathbf{T}}$ between $\mathcal{P}$ and $\mathcal{Q}$
in a differentiable manner. The network is then optimized by
back-propagating the proposed loss through this differentiable
optimization layer. The total training time is reported in Fig.~\ref{fig:train_time}. For an input containing approximately 16,000 points, each training step requires 73.9 ms.

During inference, the trained network predicts lower-triangular matrices $\mathbf{L}$ for $\mathcal{Q}$, from which the point-wise covariance sets are obtained as $\mathbf{C_q}=\mathbf{L}\mathbf{L}^{\top}$. The resulting covariances serve as local geometry for downstream tasks.

\section{Experiments}
POLI learns to predict the underlying geometry of point clouds by estimating point-wise covariance, making it a versatile module for a broad range of robotic perception tasks. In this section, we evaluate the effectiveness of POLI when integrated into robotic perception tasks: object pose estimation, LiDAR odometry, and scan augmentation. By leveraging accurately estimated point-wise covariance, we integrate it into existing pipelines through multiple representations.

For object pose estimation, we incorporate the point-wise normals predicted by POLI into an FPFH-based local descriptor~\citep{rusu2008pfh} and evaluate its effectiveness by measuring the registration success rate. For LiDAR scan matching, we integrate the covariance predicted by POLI into the GICP~\citep{segal2009generalized} framework and evaluate its LiDAR odometry performance. We also assess POLI’s scan augmentation capability by measuring downstream performance improvements when POLI-based augmentation is applied to robotic perception pipelines. Detailed experimental setups and results are provided in the Supplementary Material SD.

\begin{figure}
    \centering
    \includegraphics[width=\linewidth, trim= 5 0 7 0, clip]{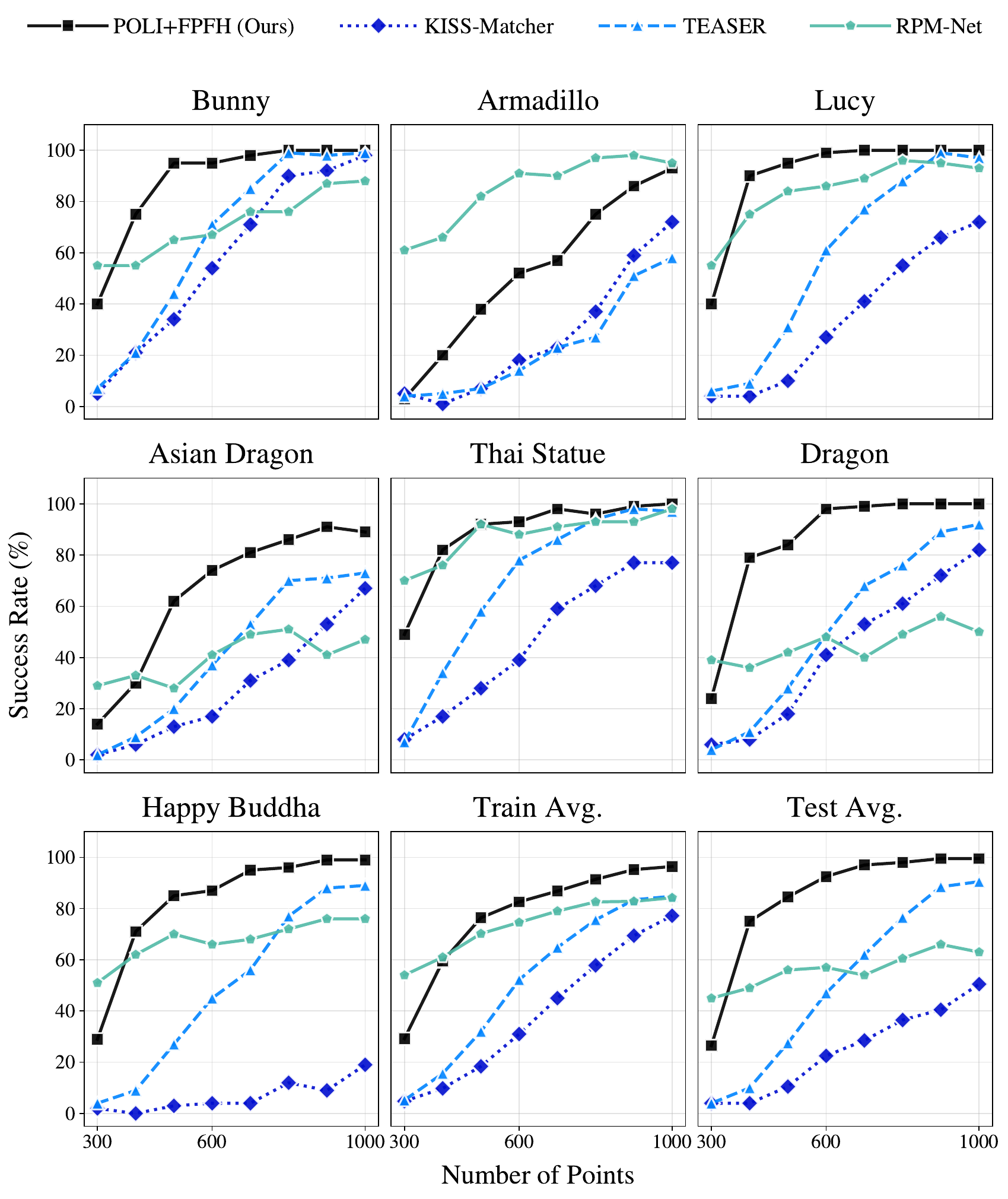}
    \caption{Performance comparison on object pose estimation across different sparsity levels.}
    \label{fig:obj_pose_est}
    \vspace{-0.4cm}
\end{figure}

\subsection{Object Pose Estimation}

\textbf{Setup.}
We evaluated POLI on the Stanford 3D Scanning Repository~\citep{curless1996volumetric}.
The training split consists of \emph{Bunny, Armadillo, Asian Dragon, Lucy,} and \emph{Thai Statue}, while the test split consists of \emph{Dragon} and \emph{Happy Buddha}.
For training, we generated scan pairs by independently sampling $N\in\{500,1{,}000\}$ points from the training objects and applying random axis--angle rotations up to $60^\circ$.
For testing, we generated point clouds from the test objects with $N\in\{300,400,500,600,700,800,900,1{,}000\}$ points, using 100 pairs per object. For correspondence estimation, we followed the Open3D FPFH pipeline~\citep{Zhou2018Open3D}, using POLI normals (the eigenvector of $\mathbf{C}_\mathbf{q}$ along its smallest eigenvalue) and selecting inliers via ROBIN~\citep{shi2021robin}. As baselines, we evaluated KISS-Matcher~\citep{lim2024kiss}, TEASER~\citep{yang2020teaser}, and RPM-Net~\citep{9157132} (Note that RPM-Net is trained on a large dataset and involves iterative registration).

\textbf{Result.} A registration is considered successful if the rotation error is $\le 10^\circ$~\citep{lim2024quatro++}. As shown in Fig.~\ref{fig:obj_pose_est}, POLI+FPFH achieves higher registration success rates across most settings and sparsity levels, and generalizes to the unseen \emph{Dragon} and \emph{Happy Buddha}, indicating benefits of accurate geometry estimation.

\begin{figure*}[t]
    % ---------- Images ----------
    \includegraphics[
        height=4.35cm,
        keepaspectratio,
        trim=40 50 0 25,
        clip
    ]{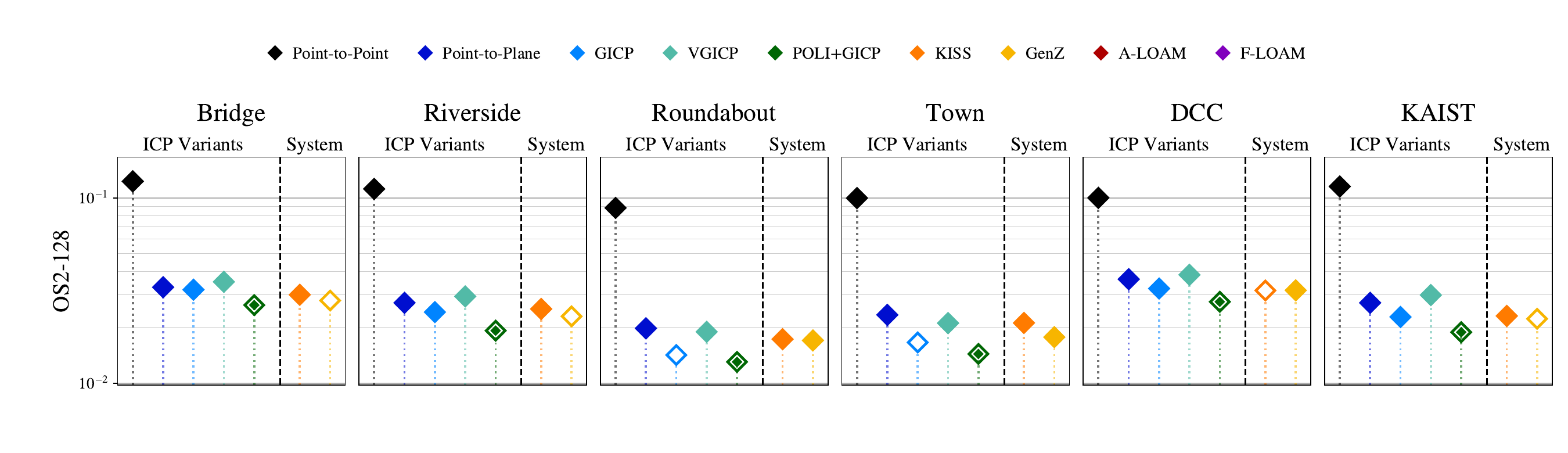}
    \includegraphics[
        height=3.0425cm,
        keepaspectratio,
        trim=40 50 0 115,
        clip
    ]{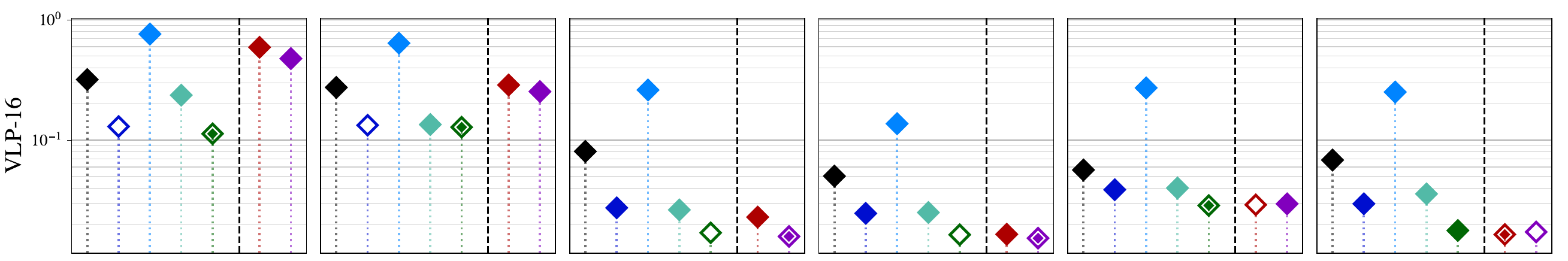}
    \caption[Evaluation of LiDAR odometry performance (VLP-16 / OS2-128)]{Comparison of LiDAR odometry performance across different algorithms based on RPE (m). The best results are marked with diamonds enclosed by rings, and the second-best results are marked with hollow diamonds.}
    \label{fig:odom_comparison_short}
    \vspace{-0.3cm}
\end{figure*}

\begin{figure}[t]
    \centering
    \includegraphics[width=\linewidth]{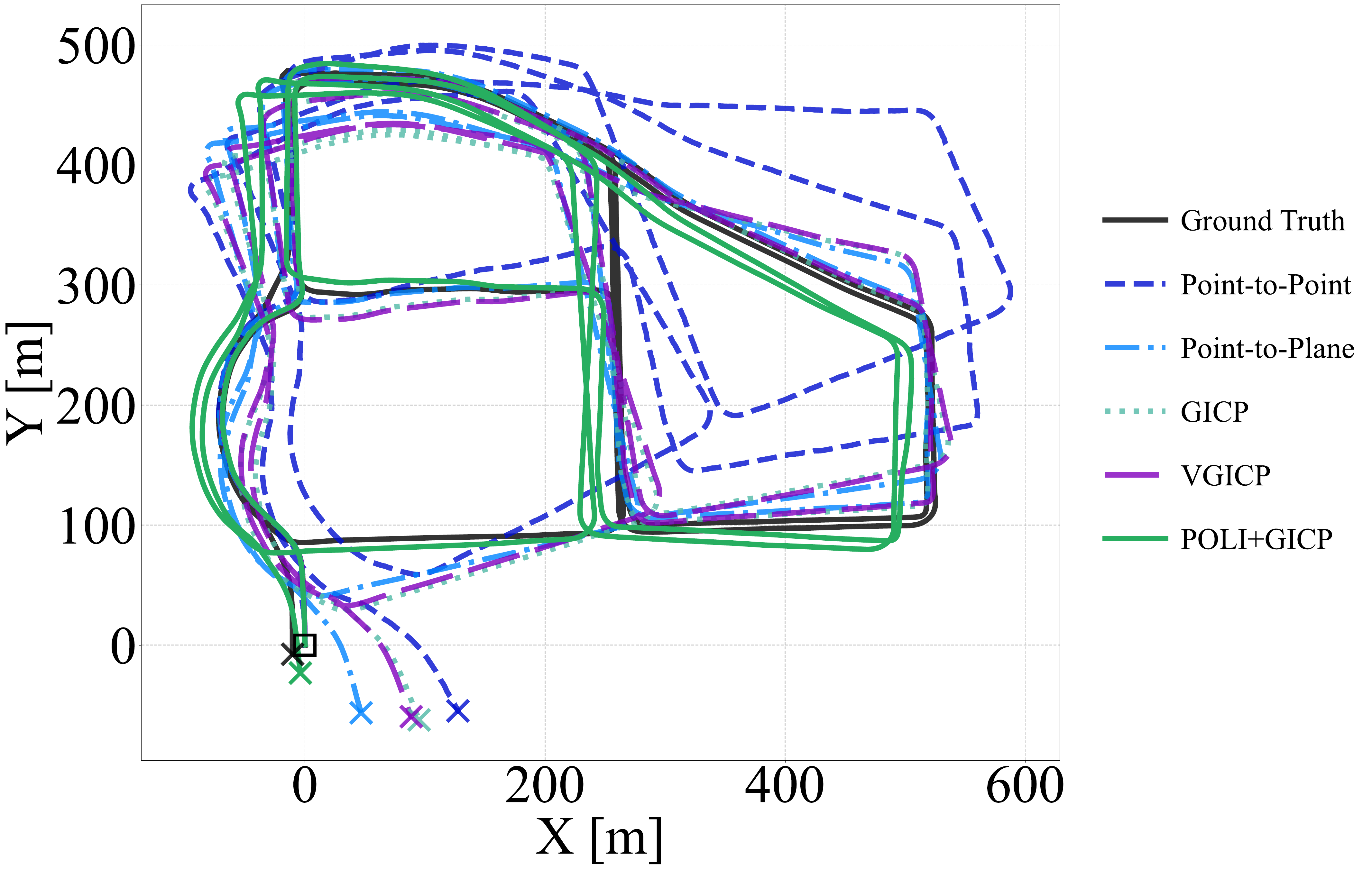}
    \caption{Comparison of LiDAR odometry trajectories for different ICP variants on the DCC04 sequence.}
    \label{fig:dcc_traj}
    \vspace{-0.4cm}
\end{figure}

\subsection{LiDAR Odometry}
\label{sec: lidar odometry}

\indent \textbf{Setup.}
We conducted our LiDAR scan matching experiments on the HeLiPR dataset~\citep{jung2024helipr}. 
The \textit{Roundabout}, \textit{Town}, \textit{Riverside}, and \textit{Bridge} sequences were used for training, while the \textit{KAIST} and \textit{DCC} sequences were reserved for testing. 
Notably, the sequences in HeLiPR~\citep{jung2024helipr} exhibit no spatial overlap. Experiments were performed using two LiDAR sensors: the Velodyne VLP-16 and the Ouster OS2-128. 

Voxel grid downsampling was applied with voxel sizes of 0.2\,m for the VLP-16 and 1.0\,m for the OS2-128. 
As baselines, we evaluated point-to-point ICP~\citep{besl1992icp}, point-to-plane ICP~\citep{rusinkiewicz2001efficient}, GICP~\citep{segal2009generalized}, and VGICP~\citep{koide2021vgicp} under the same experimental settings. 
In addition, we compared our method with LiDAR odometry systems. For the VLP-16, A-LOAM~\citep{zhang2014loam} and F-LOAM~\citep{wang2021floam} were used, while for the OS2-128, KISS-ICP~\citep{vizzo2023kiss} and GenZ-ICP~\citep{lee2024genz} were evaluated. 
To assess the original performance of these LiDAR odometry systems, we used their default configurations.

\textbf{Result.}
Performance was evaluated using relative pose error (RPE), as shown in Fig.~\ref{fig:odom_comparison_short}.
The proposed method consistently outperformed classical ICP-based approaches across all sequences. 
Furthermore, it achieved performance comparable to LiDAR odometry systems. 
Despite being evaluated on unseen environments (\textit{DCC} and \textit{KAIST}), POLI+GICP demonstrated consistent performance, showing its generalizability as shown in Fig. ~\ref{fig:dcc_traj}.
These results indicate that estimating local geometry with POLI and using it for scan matching is highly beneficial.

\begin{figure*}[!t]
    \centering
    \setlength{\tabcolsep}{2pt}
    \renewcommand{\arraystretch}{1.0}

    \renewcommand\thesubfigure{\alph{subfigure}}
    \captionsetup[sub]{font=footnotesize, justification=centering, singlelinecheck=true, skip=1pt,
                       labelformat=parens, labelsep=space}

    % ---------------- Row 1: 2 images ----------------
    \begin{tabularx}{\textwidth}{@{}XXX@{}}
        \subcaptionbox{Original point cloud}{%
            \centering
            \includegraphics[width=\linewidth, trim= 600 400 600 550, clip]{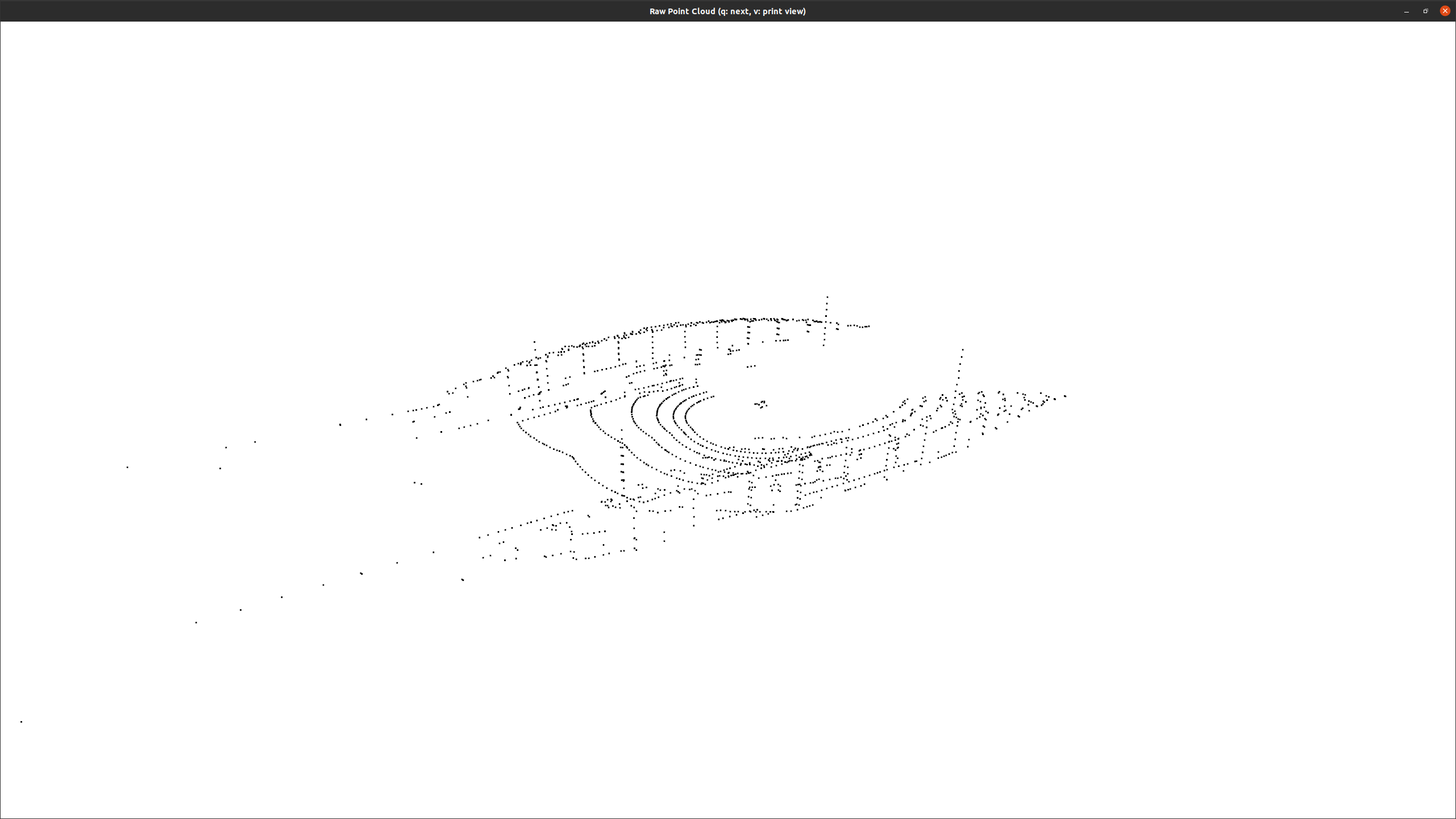}
        }
        &
        \subcaptionbox{Estimated covariances from POLI\label{fig:foo}}{%
            \centering
            \includegraphics[width=\linewidth, trim= 600 400 600 550, clip]{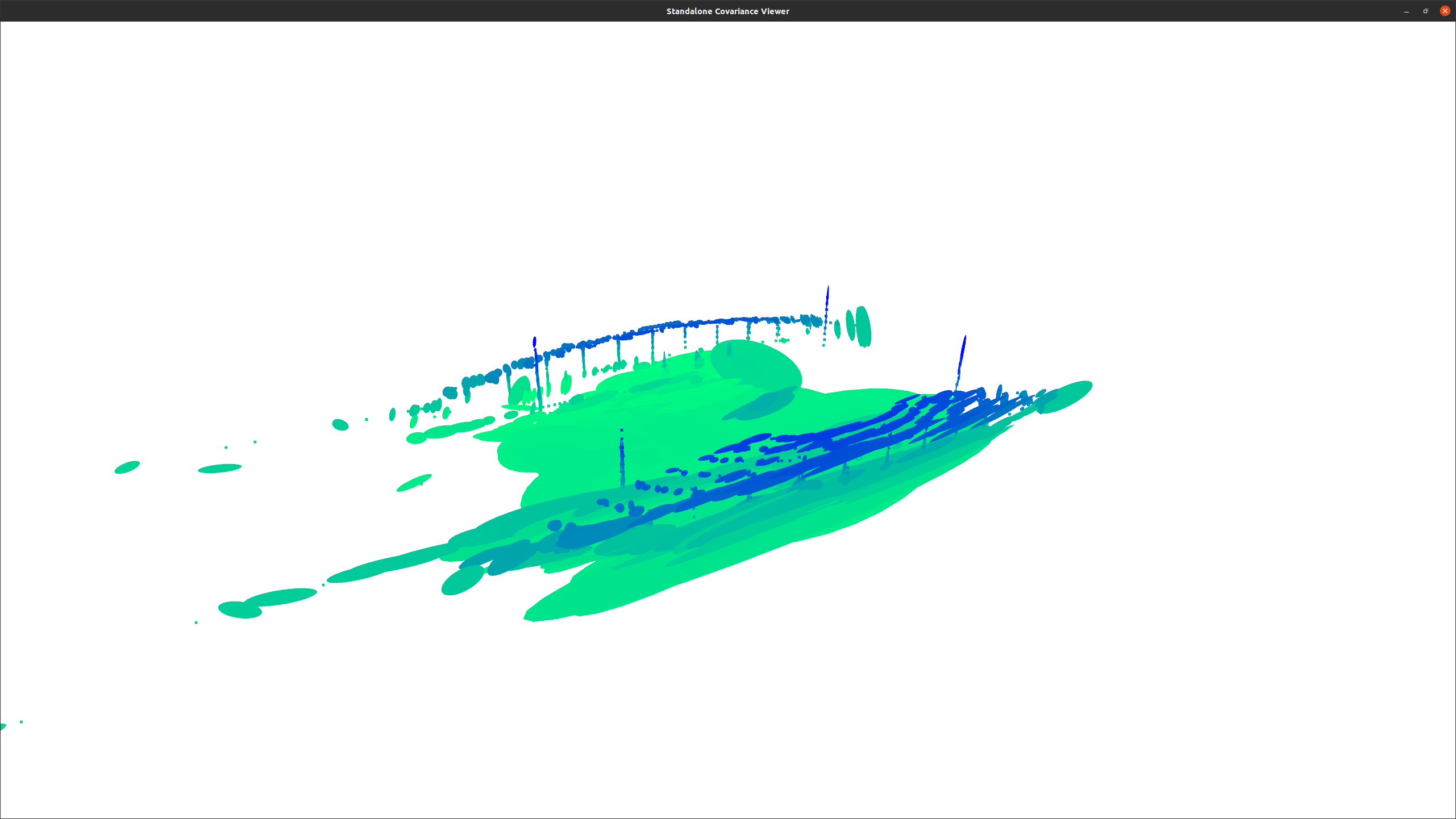}
        }
        &
        \subcaptionbox{Augmented scan from POLI\label{fig:aug2a}}{%
            \centering
            \includegraphics[width=\linewidth, trim= 600 400 600 550, clip]{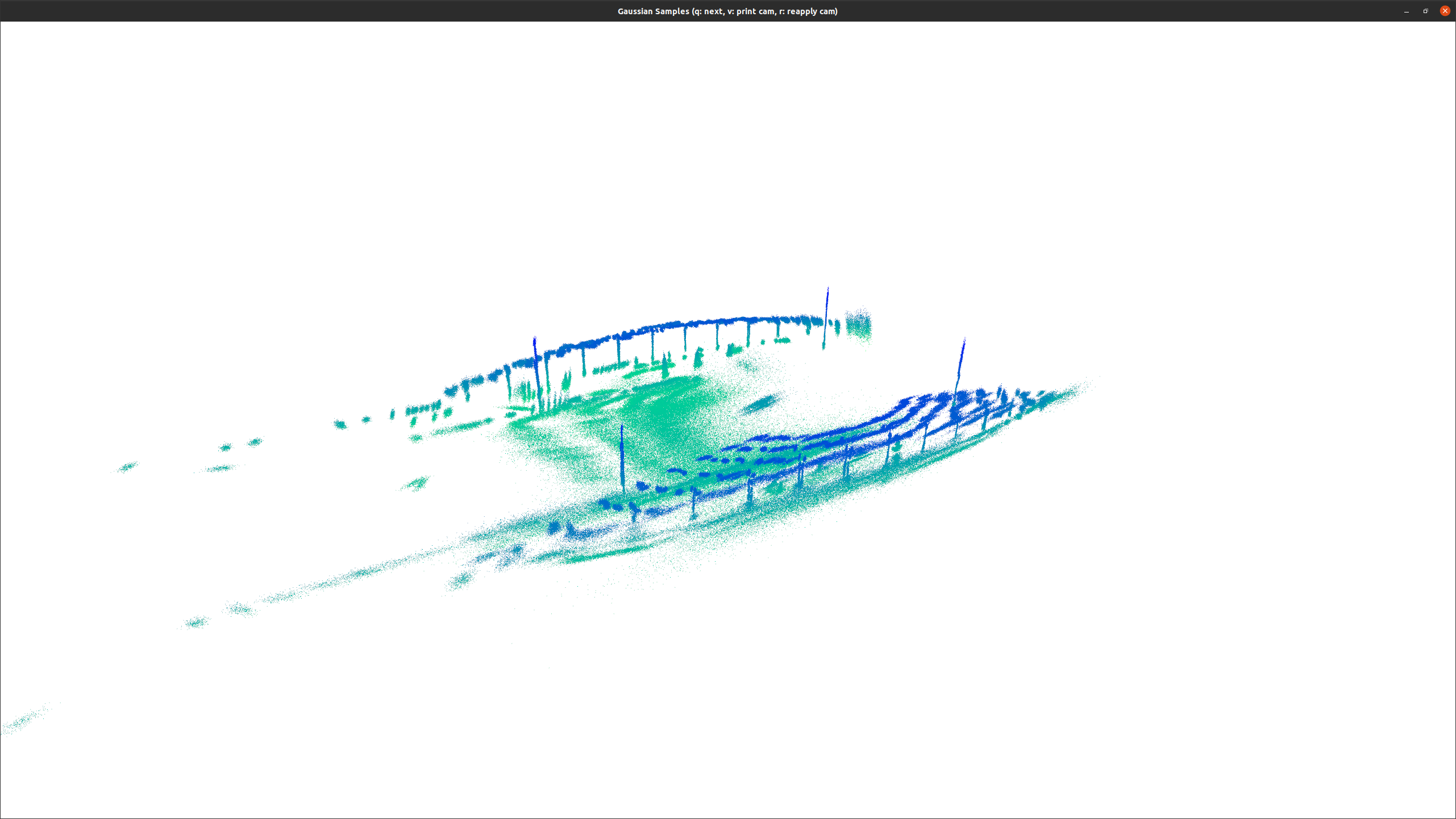}
        }
        \\
    \end{tabularx}

\caption{
Scan augmentation process using POLI: raw scan, estimated covariances, and augmented scan generated by sampling.
}

    \label{fig:XXX}
\end{figure*}

% ===================== Table 1 (Odometry) =====================
\begin{table*}[t!]
\centering
\scriptsize
\setlength{\tabcolsep}{4pt}
\renewcommand{\arraystretch}{1.05}
\caption{LiDAR odometry results with augmentation methods, evaluated using RPE (m) at $\Delta = 100$~\citep{grupp2017evo}. The best results are shown in bold, and the second-best results are underlined.}
\label{tab:helipr_tulip_odom}

\begin{tabular*}{\textwidth}{
@{\extracolsep{\fill}}
l|cccccc|ccccccccccc
}
\toprule

& \multicolumn{6}{c|}{\textbf{HeLiPR~\citep{jung2024helipr}}}
& \multicolumn{11}{c}{\textbf{KITTI~\citep{geiger2012we}}} \\
\midrule

\textbf{Method}
& \textit{Bridge} & \textit{Riverside} & \textit{Roundabout} & \textit{Town} & \textit{DCC} & \textit{KAIST}
& 00 & 01 & 02 & 03 & 04 & 05 & 06 & 07 & 08 & 09 & 10 \\
\midrule

KISS-ICP~\citep{vizzo2023kiss}
& 14.94 & 8.95 & 5.13 & 5.37 & 9.74 & 3.83
& \underline{2.50} & \underline{14.82} & \underline{2.98} & \textbf{1.77} & 2.49 & \underline{2.19} & 3.89 & \underline{1.50} & \underline{3.44} & 2.71 & \underline{2.03} \\

+PCA~\citep{Zhou2018Open3D}
& 14.37 & \underline{7.82} & 5.00 & 5.13 & 11.21 & 4.28
& 9.54 & 14.37 & 21.95 & 4.74 & 2.18 & 4.64 & 16.44 & 4.90 & 5.06 & 5.76 & 4.20 \\

+TULIP~\citep{yang2024tulip}
& \underline{12.80} & 15.48 & \underline{3.52} & \underline{3.27} & \underline{6.60} & \underline{2.59}
& 2.76 & \textbf{6.78} & 16.79 & 2.36 & \textbf{1.33} & 2.57 & \textbf{3.52} & 1.83 & 3.75 & \underline{2.55} & 2.76 \\

\textbf{+POLI}
& \textbf{3.89} & \textbf{1.92} & \textbf{1.16} & \textbf{1.03} & \textbf{1.77} & \textbf{0.95}
& \textbf{2.32} & 16.24 & \textbf{2.95} & \underline{1.78} & \underline{2.01} & \textbf{1.97} & \underline{3.58} & \textbf{1.30} & \textbf{3.42} & \textbf{2.39} & \textbf{1.66} \\

\bottomrule

\end{tabular*}
\vspace{-0.4cm}

\end{table*}

% % ===================== Table 2 (Global registration success rate) =====================
% \begin{table}[t!]
% \centering
% \scriptsize
% \setlength{\tabcolsep}{3pt}
% \renewcommand{\arraystretch}{1.05}
% \caption{ Success rate (\%) of global registration using augmentation methods. The best results are shown in bold, and the second-best results are underlined.}
% \label{tab:performance_comparison}

% % ---------------- HeLiPR ----------------
% \begin{tabular*}{\columnwidth}{@{\extracolsep{\fill}}lcccc@{}}
% \toprule
% \multicolumn{5}{c}{\textbf{HeLiPR~\citep{jung2024helipr}}} \\
% \midrule
% Method
% & Raw
% & + PCA~\citep{Zhou2018Open3D}
% & + TULIP~\citep{yang2024tulip}
% & \textbf{+ POLI} \\
% \midrule
% KISS-Matcher~\citep{lim2024kiss}
% & 65.33\%
% & \underline{66.67}\%
% & 46.67\%
% & \textbf{75.67}\% \\
% TEASER~\citep{yang2020teaser}
% & \underline{74.33}\%
% & 72.33\%
% & 61.00\%
% & \textbf{81.67}\% \\
% \bottomrule
% \end{tabular*}
% % \addlinespace[1mm]
% % ---------------- KITTI ----------------
% \begin{tabular*}{\columnwidth}{@{\extracolsep{\fill}}lcccc@{}}
% \multicolumn{5}{c}{\textbf{KITTI~\citep{geiger2012we}}} \\

% \midrule
% Method
% & Raw
% & + PCA~\citep{Zhou2018Open3D}
% & + TULIP~\citep{yang2024tulip}
% & \textbf{+ POLI} \\
% \midrule
% KISS-Matcher~\citep{lim2024kiss}
% & 54.55\%
% & 55.64\%
% & \textbf{65.09}\%
% & \underline{60.73}\% \\
% TEASER~\citep{yang2020teaser}
% & 64.36\%
% & 63.09\%
% & \underline{71.82}\%
% & \textbf{79.09}\% \\
% \bottomrule
% \end{tabular*}
% \end{table}

\begin{table}[t!]
\centering
\scriptsize
\setlength{\tabcolsep}{3pt}
\renewcommand{\arraystretch}{1.05}
\caption{Success rate (\%) of global registration using augmentation methods. The best results are shown in bold, and the second-best results are underlined.}
\label{tab:performance_comparison}

% ---------------- HeLiPR ----------------
\begin{tabular*}{\columnwidth}{@{\extracolsep{\fill}}lcccc@{}}
\toprule
\multicolumn{5}{c}{\textbf{HeLiPR~\citep{jung2024helipr}}} \\
\midrule
Method
& Raw
& + PCA~\citep{Zhou2018Open3D}
& + TULIP~\citep{yang2024tulip}
& \textbf{+ POLI} \\
\midrule
KISS-Matcher~\citep{lim2024kiss}
& 65.33\%
& \underline{66.67}\%
& 46.67\%
& \textbf{75.67}\% \\
TEASER~\citep{yang2020teaser}
& \underline{74.33}\%
& 72.33\%
& 61.00\%
& \textbf{81.67}\% \\
\bottomrule
\end{tabular*}

\vspace{1mm}

% ---------------- KITTI ----------------
\begin{tabular*}{\columnwidth}{@{\extracolsep{\fill}}lcccc@{}}
\toprule
\multicolumn{5}{c}{\textbf{KITTI~\citep{geiger2012we}}} \\
\midrule
Method
& Raw
& + PCA~\citep{Zhou2018Open3D}
& + TULIP~\citep{yang2024tulip}
& \textbf{+ POLI} \\
\midrule
KISS-Matcher~\citep{lim2024kiss}
& 54.55\%
& 55.64\%
& \textbf{65.09}\%
& \underline{60.73}\% \\
TEASER~\citep{yang2020teaser}
& 64.36\%
& 63.09\%
& \underline{71.82}\%
& \textbf{79.09}\% \\
\bottomrule
\end{tabular*}
\end{table}

% ===================== Table 3 (Different resolutions) =====================
\begin{table}[t!]
\centering
\scriptsize
\setlength{\tabcolsep}{2pt}
\renewcommand{\arraystretch}{1.05}
\caption{ Global registration success rate~(\%) across different resolutions on \citep{geiger2012we}. The best results are shown in bold, and the second-best results are underlined.}
\label{tab:success_all_beams_compact}

\begin{tabular*}{\columnwidth}{@{\extracolsep{\fill}}lcccccc@{}}
\toprule
& \multicolumn{3}{c}{ KISS-Matcher~\citep{lim2024kiss}} & \multicolumn{3}{c}{TEASER~\citep{yang2020teaser}} \\
\cmidrule(lr){2-4} \cmidrule(lr){5-7}
Resolution & Raw & + \textbf{POLI} & Gain (\%) & Base & + \textbf{POLI} & Gain (\%) \\
\midrule
16 beams & 54.55 & \underline{60.73} & $+$11.3 & 64.36 & \textbf{79.09} & $+$22.9 \\
32 beams & \textbf{87.27} & \underline{86.18} & $-$1.3  & \underline{71.82} & \textbf{92.10} & $+$28.2 \\
64 beams & \underline{94.91} & \textbf{96.91} & $+$2.1  & \textbf{97.82} & \underline{97.64} & $-$0.2 \\
\bottomrule
\end{tabular*}
\end{table}

% ===================== Table 4 (RPE/APE in DCC) =====================
\begin{table}[t!]
\centering
\scriptsize
\setlength{\tabcolsep}{2pt}
\renewcommand{\arraystretch}{1.05}
\caption{RPE and absolute pose error (APE) (m)~\citep{grupp2017evo} in \textit{DCC}~\citep{jung2024helipr}. The best results are shown in bold.}
\label{tab:dcc_compact_no_gain}

\begin{tabular*}{\columnwidth}{@{\extracolsep{\fill}}lcccccc@{}}
\toprule
& \multicolumn{3}{c}{FAST-LIO~\citep{xu2022fastlio2}} & \multicolumn{3}{c}{GLIM~\citep{koide2024glim}} \\
\cmidrule(lr){2-4} \cmidrule(lr){5-7}
Metric ([m], [°]) & Raw & + \textbf{POLI} & Gain (\%) & Raw & + \textbf{POLI} & Gain (\%) \\
\midrule
RPE ($\Delta$=100) Trans. & 4.734  & \textbf{3.200} & $-$32.4 & 7.006 & \textbf{6.864} & $-$2.0  \\
RPE ($\Delta$=100) Rot.   & 5.296  & \textbf{3.180} & $-$40.0 & 7.649 & \textbf{7.390} & $-$3.4  \\
APE Trans. & 101.05 & \textbf{44.79} & $-$55.7 & 56.94 & \textbf{35.61} & $-$37.5 \\
APE Rot.& 36.61  & \textbf{12.80} & $-$65.1 & 17.23 & \textbf{10.06} & $-$41.6 \\
\bottomrule
\end{tabular*}
\vspace{-0.4cm}
\end{table}

\subsection{Scan Augmentation}
A key advantage of \textsc{POLI} is that it models point-cloud geometry probabilistically via estimated covariances, enabling principled scan augmentation through sampling and mathematically grounded densification (Fig.~\ref{fig:XXX}(a,b,c) and Fig.~S.4). 
This allows us to densify LiDAR scans and plug it into existing robotic perception pipelines with minimal integration effort. 
We validate this via cross-dataset experiments on KITTI~\citep{geiger2012we} and HeLiPR~\citep{jung2024helipr}, focusing on LiDAR odometry and global registration. For both LiDAR odometry and scan registration, we use the VLP-16 scan dataset from HeLiPR and the 16-beam setting of KITTI used in~\citep{yang2024tulip}. For global registration, we randomly sample scan pairs within a 12 m range. For POLI, we employ the network trained in Section~\ref{sec: lidar odometry} on the HeLiPR~\citep{jung2024helipr} VLP-16 dataset.

The results are summarized in Tables~\ref{tab:helipr_tulip_odom}, and \ref{tab:performance_comparison}, and visualized in Figs.~S.2 and 3. Here, “+ Method” (e.g., + \textsc{POLI}) denotes applying the downstream pipeline to point clouds densified using the corresponding augmentation method. 
Compared against PCA~\citep{Zhou2018Open3D} and TULIP~\citep{yang2024tulip} (state-of-the-art LiDAR upsampling algorithm trained in KITTI~\citep{geiger2012we}), \textsc{POLI} consistently performs best on both KITTI~\citep{geiger2012we} and HeLiPR~\citep{jung2024helipr}. While TULIP~\citep{yang2024tulip} suffers from performance degradation under unseen data, \textsc{POLI} remains robust to variations in sensor type, field of view, and mounting configuration (e.g., the heavily occluded frontal region in HeLiPR~\citep{jung2024helipr}), highlighting the benefit of its strong geometric inductive bias induced by the combination of principled modeling and deep learning. 

In addition, we evaluate global registration performance on the KITTI~\citep{geiger2012we} dataset under varying LiDAR resolutions by downsampling the original scans to 16-, 32-, and 64-channel configurations. Importantly, no resolution-specific models are trained; instead, a single model is used across all LiDAR resolutions. As shown in Table~\ref{tab:success_all_beams_compact}, POLI improves performance in most cases, demonstrating strong robustness to resolution changes. In particular, the performance gains are more pronounced in low-resolution settings, indicating that POLI effectively mitigates information loss caused by sparse sensing. Note that POLI inference takes $10$ ms per $10,000$ points scan on an NVIDIA RTX\,4060\,Ti GPU. Furthermore, we assess the plug-and-play capability of POLI by integrating it into several existing LiDAR localization and mapping systems~\citep{koide2024glim, xu2022fastlio2}. Without any system-specific tuning or retraining, POLI yields substantial performance improvements across all evaluated pipelines, as illustrated in Table~\ref{tab:dcc_compact_no_gain}. These results suggest that POLI generalizes well across diverse robotic perception systems.

\section{Limitations}
Although the proposed method achieves consistent improvements across multiple benchmarks, several limitations remain. First, our method faces challenges under out-of-distribution (OOD) scenarios. For example, when trained on HeLiPR~\cite{jung2024helipr} and tested on KITTI~\cite{geiger2012we}, POLI still improves downstream performance, but with smaller gains than in the in-domain setting. As shown in Table~\ref{tab:helipr_tulip_odom}, POLI substantially outperforms other augmentation baselines on HeLiPR~\cite{jung2024helipr}, whereas its advantage becomes less pronounced on KITTI~\cite{geiger2012we}. Second, our method relies on a local low-order geometric representation. A 3D Gaussian covariance captures only second-order moments—i.e., the first-fundamental-form aspects of the surface such as local anisotropy and tangential extent—and is therefore agnostic to the second fundamental form that encodes local surface curvature~\cite{do2016differential}. As a result, while our covariances faithfully describe dominant structures such as local planarity~\cite{pauly2003multi}, they do not natively encode higher-order differential invariants (e.g., curvatures). Third, finding the global optimum of Eq.~(\ref{eq:map_pose_general}) in a differentiable manner remains challenging. A promising future direction is to develop a tighter relaxation of Eq.~(\ref{eq:map_pose_general}), for example, by exploiting Chordal Relaxation formulations.

\section{Conclusion}

This paper revisits the role of point cloud geometry in robotic perception and highlights that a broad range of downstream tasks are fundamentally grounded in accurate estimation of underlying geometric structure. From this perspective, we introduced a unified theoretical and practical framework that models surface geometry as a statistical manifold and employs deep learning for principled estimation of point cloud geometry. Unlike prior learning-based methods and heuristic statistical approaches, our formulation is grounded in explicit mathematical principles that govern both geometric modeling and learning, thereby enabling self-supervised learning with strong geometric inductive biases.

As a practical implementation, we proposed POLI, a learning-based point cloud geometry estimator that directly predicts point-wise geometry in the form of Gaussian covariances. 
By enforcing geometric constraints during training in a principled manner, POLI learns a mapping between point cloud and their underlying geometry, producing consistent and informative geometric representations.
Extensive experiments demonstrate that the proposed theory and practice successfully estimate the point cloud geometry, and consistently improve performance across a wide range of robotic perception tasks, where the learned geometry can be seamlessly integrated as 3D covariances, surface normals, or augmented point clouds. 
Overall, this work bridges geometric modeling and learning-based inference, offering a unified perspective on point cloud geometry representation and establishing statistical manifold-based covariance learning for geometric modeling as a promising paradigm for robust and scalable 3D robotic perception.

\section*{Acknowledgment}

This work was supported by the National Research Foundation of Korea (NRF) grant funded by the Korea government (MSIT) (No. 2023R1A2C2003130), and by the Unmanned Vehicles Core Technology Research and Development Program through the National Research Foundation of Korea (NRF) and the Unmanned Vehicle Advanced Research Center (UVARC), funded by the Ministry of Science and ICT, Republic of Korea (No. 2020M3C1C1A01082375).
% \clearpage
% \newpage

\bibliographystyle{plainnat}
\bibliography{references}

@inproceedings{segal2009generalized,
  title={\href{https://www.robots.ox.ac.uk/~avsegal/resources/papers/Generalized_ICP.pdf}{Generalized-{ICP}}},
  author={Segal, Avi and Haehnel, Dirk and Thrun, Sebastian},
  booktitle={Proceedings of Robotics: Science and Systems},
  year={2009},
URL = {https://www.robots.ox.ac.uk/~avsegal/resources/papers/Generalized_ICP.pdf}
}

@article{vizzo2023kiss,
  title={\href{https://ieeexplore.ieee.org/abstract/document/10015694/}{KISS-{ICP}: In Defense of Point-to-Point {ICP} -- Simple, Accurate, and Robust Registration If Done the Right Way}},
  author={Vizzo, Ignacio and Guadagnino, Tiziano and Mersch, Benedikt and Wiesmann, Louis and Behley, Jens and Stachniss, Cyrill},
  journal={IEEE Robotics and Automation Letters},
  volume={8},
  number={2},
  pages={1029--1036},
  year={2023},
URL={https://ieeexplore.ieee.org/abstract/document/10015694/}
}

@article{lee2024genz,
  title={\href{https://ieeexplore.ieee.org/abstract/document/10753079/}{GenZ-ICP: Generalizable and Degeneracy-Robust {LiDAR} Odometry Using an Adaptive Weighting}},
  author={Lee, Daehan and Lim, Hyungtae and Han, Soohee},
  journal={IEEE Robotics and Automation Letters},
  volume={9},
  number={1},
  pages={456--463},
  year={2024},
URL={https://ieeexplore.ieee.org/abstract/document/10753079}
}

@inproceedings{wang2021floam,
  title={\href{https://ieeexplore.ieee.org/abstract/document/9636655}{F-LOAM: Fast Lidar Odometry and Mapping}},
  author={Wang, Han and Wang, Chen and Chen, Chun-Lin and Xie, Lihua},
  booktitle={Proceedings of the IEEE/RSJ International Conference on Intelligent Robots and Systems},
  year={2021},
URL={https://ieeexplore.ieee.org/abstract/document/9636655}
}

@inproceedings{rusinkiewicz2001efficient,
  title={\href{https://ieeexplore.ieee.org/abstract/document/924423}{Efficient Variants of the {ICP} Algorithm}},
  author={Rusinkiewicz, Szymon and Levoy, Marc},
  booktitle={Proceedings of the International Conference on {3D} Digital Imaging and Modeling},
  year={2001},
URL={https://ieeexplore.ieee.org/abstract/document/924423}
}

@inproceedings{qi2017pointnet++,
  title={\href{https://proceedings.neurips.cc/paper/2017/hash/d8bf84be3800d12f74d8b05e9b89836f-Abstract.html}{PointNet++: Deep Hierarchical Feature Learning on Point Sets in a Metric Space}},
  author={Qi, Charles Ruizhongtai and Yi, Li and Su, Hao and Guibas, Leonidas J.},
  booktitle={Advances in Neural Information Processing Systems},
  year={2017},
URL={https://proceedings.neurips.cc/paper/2017/hash/d8bf84be3800d12f74d8b05e9b89836f-Abstract.html}
}

@article{jung2024helipr,
  title={\href{https://journals.sagepub.com/doi/full/10.1177/02783649241242136}{HeLiPR: Heterogeneous {LiDAR} Dataset for Inter-{LiDAR} Place Recognition Under Spatiotemporal Variations}},
  author={Jung, Minwoo and Yang, Wooseong and Lee, Dongjae and Gil, Hyeonjae and Kim, Giseop and Kim, Ayoung},
  journal={The International Journal of Robotics Research},
  volume={43},
  number={3},
  pages={1867--1883},
  year={2024},
URL={https://journals.sagepub.com/doi/full/10.1177/02783649241242136}
}

@misc{grupp2017evo,
  title={\href{https://github.com/MichaelGrupp/evo}{EVO: Python Package for the Evaluation of Odometry and {SLAM}}},
  author={Grupp, Michael},
  year={2017},
    URL={https://github.com/MichaelGrupp/evo}
}

@book{amari2000methods,
  title={Methods of information geometry},
  author={Amari, Shun-ichi and Nagaoka, Hiroshi},
  volume={191},
  year={2000},
  publisher={American Mathematical Soc.},
URL={https://books.google.co.kr/books?hl=ko&lr=&id=vc2FWSo7wLUC&oi=fnd&pg=PR7&dq=Methods+of+information+geometry&ots=4JsyKx44L_&sig=ZMJpkAT36TZSpRPPkYM-gxyXLLM&redir_esc=y#v=onepage&q=Methods%20of%20information%20geometry&f=false}
}

@inproceedings{lim2024kiss,
  title={\href{https://ieeexplore.ieee.org/abstract/document/11127458}{KISS-Matcher: Fast and Robust Point Cloud Registration Revisited}},
  author={Lim, Hyungtae and Kim, Daebeom and Shin, Gunhee and Shi, Jingnan and Vizzo, Ignacio and Myung, Hyun and Park, Jaesik and Carlone, Luca},
  booktitle={Proceedings of the IEEE International Conference on Robotics and Automation},
  year={2025},
URL={https://ieeexplore.ieee.org/abstract/document/11127458}
}

@article{yang2020teaser,
  title={\href{https://ieeexplore.ieee.org/abstract/document/9286491/}{Teaser: Fast and certifiable point cloud registration}},
  author={Yang, Heng and Shi, Jingnan and Carlone, Luca},
  journal={IEEE Transactions on Robotics},
  volume={37},
  number={2},
  pages={314--333},
  year={2020},
  publisher={IEEE},
URL={https://ieeexplore.ieee.org/abstract/document/9286491/}
}

@INPROCEEDINGS{9157132,
  author={Yew, Zi Jian and Lee, Gim Hee},
  booktitle={2020 IEEE/CVF Conference on Computer Vision and Pattern Recognition (CVPR)}, 
  title={\href{https://openaccess.thecvf.com/content_CVPR_2020/html/Yew_RPM-Net_Robust_Point_Matching_Using_Learned_Features_CVPR_2020_paper.html}{RPM-Net: Robust Point Matching Using Learned Features}}, 
  year={2020},
  volume={},
  number={},
  pages={11821-11830},
  doi={10.1109/CVPR42600.2020.01184},
URL={https://openaccess.thecvf.com/content_CVPR_2020/html/Yew_RPM-Net_Robust_Point_Matching_Using_Learned_Features_CVPR_2020_paper.html}}

@article{holmes2025sdprlayers,
  title={\href{https://ieeexplore.ieee.org/abstract/document/11029144/}{{SDPRL}ayers: Certifiable Backpropagation Through Polynomial Optimization Problems in Robotics}},
  author={Holmes, Connor and D{\"u}mbgen, Frederike and Barfoot, Timothy D},
  journal={IEEE Transactions on Robotics},
  year={2025},
  publisher={IEEE},
URL={https://ieeexplore.ieee.org/abstract/document/11029144/}
}

@inproceedings{geiger2012we,
  title={\href{https://ieeexplore.ieee.org/abstract/document/6248074}{Are we ready for autonomous driving? the kitti vision benchmark suite}},
  author={Geiger, Andreas and Lenz, Philip and Urtasun, Raquel},
  booktitle={2012 IEEE conference on computer vision and pattern recognition},
  pages={3354--3361},
  year={2012},
  organization={IEEE},
URL={https://ieeexplore.ieee.org/abstract/document/6248074}
}

@inproceedings{curless1996volumetric,
  title={\href{https://dl.acm.org/doi/abs/10.1145/237170.237269}{A volumetric method for building complex models from range images}},
  author={Curless, Brian and Levoy, Marc},
  booktitle={Proceedings of the 23rd annual conference on Computer graphics and interactive techniques},
  pages={303--312},
  year={1996},
URL={https://dl.acm.org/doi/abs/10.1145/237170.237269}
}

@article{Zhou2018Open3D,
  author  = {Qian-Yi Zhou and Jaesik Park and Vladlen Koltun},
  title   = {\href{https://arxiv.org/abs/1801.09847}{{Open3D}: A Modern Library for {3D} Data Processing}},
  journal = {arXiv preprint arXiv:1801.09847},
  year    = {2018},
  url     = {https://arxiv.org/abs/1801.09847},
URL={https://arxiv.org/abs/1801.09847}
}

@inproceedings{shi2021robin,
  title={\href{https://ieeexplore.ieee.org/abstract/document/9562007}{ROBIN: a graph-theoretic approach to reject outliers in robust estimation using invariants}},
  author={Shi, Jingnan and Yang, Heng and Carlone, Luca},
  booktitle={2021 IEEE International Conference on Robotics and Automation (ICRA)},
  pages={13820--13827},
  year={2021},
  organization={IEEE},
URL={https://ieeexplore.ieee.org/abstract/document/9562007}
}

@inproceedings{hoppe1992surface,
  title={\href{https://dl.acm.org/doi/abs/10.1145/133994.134011}{Surface reconstruction from unorganized points}},
  author={Hoppe, Hugues and DeRose, Tony and Duchamp, Tom and McDonald, John and Stuetzle, Werner},
  booktitle={Proceedings of the 19th annual conference on computer graphics and interactive techniques},
  pages={71--78},
  year={1992},
URL={https://dl.acm.org/doi/abs/10.1145/133994.134011}
}

@inproceedings{rusu2009fast,
  title={\href{https://ieeexplore.ieee.org/abstract/document/5152473}{Fast Point Feature Histograms ({FPFH}) for {3D} registration}},
  author={Rusu, Radu Bogdan and Blodow, Nico and Beetz, Michael},
  booktitle={2009 IEEE International Conference on Robotics and Automation},
  pages={3212--3217},
  year={2009},
  organization={IEEE}
}

@inproceedings{pauly2003multi,
  title={\href{https://onlinelibrary.wiley.com/doi/full/10.1111/1467-8659.00675}{Multi-scale feature extraction on point-sampled surfaces}},
  author={Pauly, Mark and Keiser, Richard and Gross, Markus},
  booktitle={Computer Graphics Forum},
  volume={22},
  number={3},
  pages={281--289},
  year={2003},
  organization={Wiley Online Library},
URL={https://onlinelibrary.wiley.com/doi/full/10.1111/1467-8659.00675}
}

@article{besl1992icp,
  author  = {Paul J. Besl and Neil D. McKay},
  title   = {\href{https://www.spiedigitallibrary.org/conference-proceedings-of-spie/1611/1/Method-for-registration-of-3-D-shapes/10.1117/12.57955.short}{A Method for Registration of {3D} Shapes}},
  journal = {IEEE Transactions on Pattern Analysis and Machine Intelligence},
  volume  = {14},
  number  = {2},
  pages   = {239--256},
  year    = {1992},
  doi     = {10.1109/34.121791},
URL={https://www.spiedigitallibrary.org/conference-proceedings-of-spie/1611/1/Method-for-registration-of-3-D-shapes/10.1117/12.57955.short}
}

@inproceedings{koide2021vgicp,
  author    = {Kenji Koide and Masashi Yokozuka and Shuji Oishi and Atsuhiko Banno},
  title     = {\href{https://ieeexplore.ieee.org/abstract/document/9560835}{Voxelized {GICP} for Fast and Accurate {3D} Point Cloud Registration}},
  booktitle = {Proceedings of the IEEE International Conference on Robotics and Automation (ICRA)},
  pages     = {11054--11059},
  year      = {2021},
URL={https://ieeexplore.ieee.org/abstract/document/9560835}
}

@INPROCEEDINGS{zhang2014loam, 
    AUTHOR    = {Ji Zhang AND Sanjiv Singh}, 
    TITLE     = {\href{https://www.ri.cmu.edu/pub_files/2014/7/Ji_LidarMapping_RSS2014_v8.pdf}{LOAM: Lidar Odometry and Mapping in Real-time}}, 
    BOOKTITLE = {Proceedings of Robotics: Science and Systems}, 
    YEAR      = {2014}, 
    ADDRESS   = {Berkeley, USA}, 
    MONTH     = {July},
    DOI       = {10.15607/RSS.2014.X.007} ,
URL={https://www.ri.cmu.edu/pub_files/2014/7/Ji_LidarMapping_RSS2014_v8.pdf}
}

@article{xu2021fastlio,
  author  = {Wei Xu and Yixi Cai and Dawei He and Jiarong Lin and Fu Zhang},
  title   = {\href{https://ieeexplore.ieee.org/abstract/document/9372856/}{{FAST-LIO}: A Fast, Robust {LiDAR}-Inertial Odometry Package by Tightly-Coupled {Iterated Kalman Filter}}},
  journal = {IEEE Robotics and Automation Letters},
  year    = {2021},
  doi     = {10.1109/LRA.2021.3064227},
URL={https://ieeexplore.ieee.org/abstract/document/9372856/}
}

@article{xu2022fastlio2,
  author  = {Wei Xu and Yixi Cai and Dawei He and Jiarong Lin and Fu Zhang},
  title   = {\href{https://ieeexplore.ieee.org/abstract/document/9697912/}{{FAST-LIO2}: Fast Direct {LiDAR}-Inertial Odometry}},
  journal = {IEEE Transactions on Robotics},
  year    = {2022},
  doi     = {10.1109/TRO.2022.3141876},
URL={https://ieeexplore.ieee.org/abstract/document/9697912/}
}

@inproceedings{rusu2008pfh,
  author    = {Radu Bogdan Rusu and Zoltan C. Marton and Nico Blodow and Michael Beetz},
  title     = {\href{https://ebooks.iospress.nl/doi/10.3233/978-1-58603-887-8-119}{Persistent Point Feature Histograms for {3D} Point Clouds}},
  booktitle = {Proceedings of the 10th International Conference on Intelligent Autonomous Systems (IAS)},
  year      = {2008},
URL={https://ebooks.iospress.nl/doi/10.3233/978-1-58603-887-8-119}
}

@inproceedings{tombari2010shot,
  author    = {Federico Tombari and Samuele Salti and Luigi Di Stefano},
  title     = {\href{https://link.springer.com/chapter/10.1007/978-3-642-15558-1_26}{Unique Signatures of Histograms for Local Surface Description}},
  booktitle = {Proceedings of the European Conference on Computer Vision (ECCV)},
  pages     = {356--369},
  year      = {2010},
URL={https://link.springer.com/chapter/10.1007/978-3-642-15558-1_26}
}

@article{guo2013rops,
  author  = {Guo, Yulan and Sohel, Ferdous and Bennamoun, Mohammed and Lu, Min and Wan, Jianwei},
  title   = {\href{https://link.springer.com/article/10.1007/s11263-013-0627-y}{Rotational Projection Statistics for {3D} Local Surface Description and Object Recognition}},
  journal = {International Journal of Computer Vision},
  volume  = {105},
  number  = {1},
  pages   = {63--86},
  year    = {2013},
  doi     = {10.1007/s11263-013-0627-y},
URL={https://link.springer.com/article/10.1007/s11263-013-0627-y}
}

@inproceedings{zhou2016fgr,
  author    = {Qian{-}Yi Zhou and Jaesik Park and Vladlen Koltun},
  title     = {\href{https://link.springer.com/chapter/10.1007/978-3-319-46475-6_47}{Fast Global Registration}},
  booktitle = {Proceedings of the European Conference on Computer Vision (ECCV)},
  pages     = {766--782},
  year      = {2016},
URL={https://link.springer.com/chapter/10.1007/978-3-319-46475-6_47}
}

@article{seo2025buffer,
  title={\href{https://arxiv.org/abs/2503.07940}{BUFFER-X: Towards Zero-Shot Point Cloud Registration in Diverse Scenes}},
  author={Seo, Minkyun and Lim, Hyungtae and Lee, Kanghee and Carlone, Luca and Park, Jaesik},
  journal={arXiv preprint arXiv:2503.07940},
  year={2025},
URL={https://arxiv.org/abs/2503.07940}
}

@article{lim2024quatro++,
  title={\href{https://journals.sagepub.com/doi/full/10.1177/02783649231207654}{Quatro++: Robust global registration exploiting ground segmentation for loop closing in LiDAR SLAM}},
  author={Lim, Hyungtae and Kim, Beomsoo and Kim, Daebeom and Mason Lee, Eungchang and Myung, Hyun},
  journal={The International Journal of Robotics Research},
  volume={43},
  number={5},
  pages={685--715},
  year={2024},
  publisher={SAGE Publications Sage UK: London, England},
URL={https://journals.sagepub.com/doi/full/10.1177/02783649231207654}
}

@InProceedings{nunes2024lidar_diffusion,
  author    = {Nunes, Lucas and Marcuzzi, Rodrigo and Mersch, Benedikt and Behley, Jens and Stachniss, Cyrill},
  title     = {\href{https://openaccess.thecvf.com/content/CVPR2024/html/Nunes_Scaling_Diffusion_Models_to_Real-World_3D_LiDAR_Scene_Completion_CVPR_2024_paper.html}{Scaling Diffusion Models to Real-World {3D} {LiDAR} Scene Completion}},
  booktitle = {Proceedings of the IEEE/CVF Conference on Computer Vision and Pattern Recognition (CVPR)},
  year      = {2024},
  pages     = {14770--14780},
URL={https://openaccess.thecvf.com/content/CVPR2024/html/Nunes_Scaling_Diffusion_Models_to_Real-World_3D_LiDAR_Scene_Completion_CVPR_2024_paper.html}
}

@article{vizzo2022make,
  title={\href{https://ieeexplore.ieee.org/abstract/document/9812507}{Make it dense: Self-supervised geometric scan completion of sparse 3D lidar scans in large outdoor environments}},
  author={Vizzo, Ignacio and Mersch, Benedikt and Marcuzzi, Rodrigo and Wiesmann, Louis and Behley, Jens and Stachniss, Cyrill},
  journal={IEEE Robotics and Automation Letters},
  volume={7},
  number={3},
  pages={8534--8541},
  year={2022},
  publisher={IEEE},
URL={https://ieeexplore.ieee.org/abstract/document/9812507}
}

@inproceedings{zhou20213d,
  title={\href{https://openaccess.thecvf.com/content/ICCV2021/html/Zhou_3D_Shape_Generation_and_Completion_Through_Point-Voxel_Diffusion_ICCV_2021_paper.html?ref=https://githubhelp.com}{{3D} shape generation and completion through point-voxel diffusion}},
  author={Zhou, Linqi and Du, Yilun and Wu, Jiajun},
  booktitle={Proceedings of the IEEE/CVF International Conference on Computer Vision},
  pages={5826--5835},
  year={2021},
URL={https://openaccess.thecvf.com/content/ICCV2021/html/Zhou_3D_Shape_Generation_and_Completion_Through_Point-Voxel_Diffusion_ICCV_2021_paper.html?ref=https://githubhelp.com}
}

@inproceedings{lenssen2020deep,
  title={\href{https://openaccess.thecvf.com/content_CVPR_2020/html/Lenssen_Deep_Iterative_Surface_Normal_Estimation_CVPR_2020_paper.html}{Deep iterative surface normal estimation}},
  author={Lenssen, Jan Eric and Osendorfer, Christian and Masci, Jonathan},
  booktitle={Proceedings of the IEEE/CVF Conference on Computer Vision and Pattern Recognition},
  pages={11247--11256},
  year={2020},
URL={https://openaccess.thecvf.com/content_CVPR_2020/html/Lenssen_Deep_Iterative_Surface_Normal_Estimation_CVPR_2020_paper.html}
}

@inproceedings{li2025high,
  title={\href{https://openaccess.thecvf.com/content/CVPR2025/html/Li_High-quality_Point_Cloud_Oriented_Normal_Estimation_via_Hybrid_Angular_and_CVPR_2025_paper.html}{High-quality Point Cloud Oriented Normal Estimation via Hybrid Angular and Euclidean Distance Encoding}},
  author={Li, Yuanqi and Huang, Jingcheng and Wang, Hongshen and Lv, Peiyuan and Liu, Yansong and Zheng, Jiuming and Guo, Jie and Guo, Yanwen},
  booktitle={Proceedings of the Computer Vision and Pattern Recognition Conference},
  pages={1287--1296},
  year={2025},
URL={https://openaccess.thecvf.com/content/CVPR2025/html/Li_High-quality_Point_Cloud_Oriented_Normal_Estimation_via_Hybrid_Angular_and_CVPR_2025_paper.html}
}

@inproceedings{li2025learning,
  title={\href{https://openaccess.thecvf.com/content/ICCV2025/html/Li_Learning_Normals_of_Noisy_Points_by_Local_Gradient-Aware_Surface_Filtering_ICCV_2025_paper.html}{Learning Normals of Noisy Points by Local Gradient-Aware Surface Filtering}},
  author={Li, Qing and Feng, Huifang and Gong, Xun and Liu, Yu-Shen},
  booktitle={Proceedings of the IEEE/CVF International Conference on Computer Vision},
  pages={28828--28838},
  year={2025},
URL={https://openaccess.thecvf.com/content/ICCV2025/html/Li_Learning_Normals_of_Noisy_Points_by_Local_Gradient-Aware_Surface_Filtering_ICCV_2025_paper.html}
}

@article{koide2024glim,
  title={\href{https://www.sciencedirect.com/science/article/pii/S0921889024001349}{{GLIM}: {3D} range-inertial localization and mapping with {GPU}-accelerated scan matching factors}},
  author={Koide, Kenji and Yokozuka, Masashi and Oishi, Shuji and Banno, Atsuhiko},
  journal={Robotics and Autonomous Systems},
  volume={179},
  pages={104750},
  year={2024},
  publisher={Elsevier},
URL={https://www.sciencedirect.com/science/article/pii/S0921889024001349}
}

@inproceedings{benshabat2019nestinet,
  title     = {\href{https://openaccess.thecvf.com/content_CVPR_2019/html/Ben-Shabat_Nesti-Net_Normal_Estimation_for_Unstructured_3D_Point_Clouds_Using_Convolutional_CVPR_2019_paper.html}{Nesti-{Net}: Normal Estimation for Unstructured {3D} Point Clouds Using Convolutional Neural Networks}},
  author    = {Ben-Shabat, Yizhak and Lindenbaum, Michael and Fischer, Anath},
  booktitle = {Proceedings of the IEEE/CVF Conference on Computer Vision and Pattern Recognition (CVPR)},
  year      = {2019},
  doi       = {10.1109/CVPR.2019.01035},
URL={https://openaccess.thecvf.com/content_CVPR_2019/html/Ben-Shabat_Nesti-Net_Normal_Estimation_for_Unstructured_3D_Point_Clouds_Using_Convolutional_CVPR_2019_paper.html}
}

@inproceedings{benshabat2020deepfit,
  title     = {\href{https://link.springer.com/chapter/10.1007/978-3-030-58452-8_2}{DeepFit: {3D} Surface Fitting via Neural Network Weighted Least Squares}},
  author    = {Ben-Shabat, Yizhak and Gould, Stephen},
  booktitle = {European Conference on Computer Vision (ECCV)},
  year      = {2020},
  doi       = {10.1007/978-3-030-58452-8_2},
URL={https://link.springer.com/chapter/10.1007/978-3-030-58452-8_2}
}

@article{holmes2024semidefinite,
  title={\href{https://ieeexplore.ieee.org/abstract/document/10706005}{On semidefinite relaxations for matrix-weighted state-estimation problems in robotics}},
  author={Holmes, Connor and D{\"u}mbgen, Frederike and Barfoot, Timothy},
  journal={IEEE Transactions on Robotics},
  volume={40},
  pages={4805--4824},
  year={2024},
  publisher={IEEE},
URL={https://ieeexplore.ieee.org/abstract/document/10706005}
}

@inproceedings{yang2024tulip,
  title={\href{https://openaccess.thecvf.com/content/CVPR2024/html/Yang_TULIP_Transformer_for_Upsampling_of_LiDAR_Point_Clouds_CVPR_2024_paper.html}{Tulip: Transformer for upsampling of lidar point clouds}},
  author={Yang, Bin and Pfreundschuh, Patrick and Siegwart, Roland and Hutter, Marco and Moghadam, Peyman and Patil, Vaishakh},
  booktitle={Proceedings of the IEEE/CVF Conference on Computer Vision and Pattern Recognition},
  pages={15354--15364},
  year={2024},
URL={https://openaccess.thecvf.com/content/CVPR2024/html/Yang_TULIP_Transformer_for_Upsampling_of_LiDAR_Point_Clouds_CVPR_2024_paper.html}
}

@article{blondel2022efficient,
  title={\href{https://proceedings.neurips.cc/paper_files/paper/2022/hash/228b9279ecf9bbafe582406850c57115-Abstract-Conference.html}{Efficient and modular implicit differentiation}},
  author={Blondel, Mathieu and Berthet, Quentin and Cuturi, Marco and Frostig, Roy and Hoyer, Stephan and Llinares-L{\'o}pez, Felipe and Pedregosa, Fabian and Vert, Jean-Philippe},
  journal={Advances in neural information processing systems},
  volume={35},
  pages={5230--5242},
  year={2022},
URL={https://proceedings.neurips.cc/paper_files/paper/2022/hash/228b9279ecf9bbafe582406850c57115-Abstract-Conference.html}
}

@inproceedings{qiu2022pu,
  title={\href{https://openaccess.thecvf.com/content/ACCV2022/html/Qiu_PU-Transformer_Point_Cloud_Upsampling_Transformer_ACCV_2022_paper.html}{Pu-transformer: Point cloud upsampling transformer}},
  author={Qiu, Shi and Anwar, Saeed and Barnes, Nick},
  booktitle={Proceedings of the Asian conference on computer vision},
  pages={2475--2493},
  year={2022},
URL={https://openaccess.thecvf.com/content/ACCV2022/html/Qiu_PU-Transformer_Point_Cloud_Upsampling_Transformer_ACCV_2022_paper.html}
}

@inproceedings{kwon2022implicit,
  title={\href{https://ieeexplore.ieee.org/abstract/document/9811992}{Implicit LiDAR network: LiDAR super-resolution via interpolation weight prediction}},
  author={Kwon, Youngsun and Sung, Minhyuk and Yoon, Sung--Eui},
  booktitle={2022 international conference on robotics and automation (ICRA)},
  pages={8424--8430},
  year={2022},
  organization={IEEE},
URL={https://ieeexplore.ieee.org/abstract/document/9811992}
}

@book{mackay2003information,
  title={Information theory, inference and learning algorithms},
  author={MacKay, David JC},
  year={2003},
  publisher={Cambridge university press},
URL={https://books.google.co.kr/books?hl=ko&lr=&id=AKuMj4PN_EMC&oi=fnd&pg=PR11&dq=Information+theory,+inference+and+learning+algorithms&ots=EOnrla3wBb&sig=6ziiQReq1O5rpGa9MGPyKNwGdeI&redir_esc=y#v=onepage&q=Information%20theory%2C%20inference%20and%20learning%20algorithms&f=false}
}

@inproceedings{briales2017convex,
  title={Convex global 3d registration with lagrangian duality},
  author={Briales, Jesus and Gonzalez-Jimenez, Javier},
  booktitle={Proceedings of the IEEE conference on computer vision and pattern recognition},
  pages={4960--4969},
  year={2017}
}

@book{do2016differential,
  title={Differential geometry of curves and surfaces: revised and updated second edition},
  author={Do Carmo, Manfredo P},
  year={2016},
  publisher={Courier Dover Publications}
}

\setcounter{section}{0}
\renewcommand{\thesection}{S\Alph{section}.}
\renewcommand{\thesubsection}{\thesection\arabic{subsection}.}

% spacing after the number (SA.1 Title)
\makeatletter
\renewcommand{\@seccntformat}[1]{\csname the#1\endcsname~}
\makeatother
\setcounter{footnote}{0}
\renewcommand{\thefootnote}{S\arabic{footnote}}
\setcounter{equation}{0}
\renewcommand{\theequation}{S\arabic{equation}}
\clearpage
\onecolumn

\newpage

\clearpage
\newpage
\begin{titlepage}
  \fontfamily{ptm}\selectfont
  \centering
  \thispagestyle{empty}

  % ============= PAGE 1: Title Block =============

\vspace*{\fill}

% --- Subtitle banner ---
{\large\scshape%
  S\,u\,p\,p\,l\,e\,m\,e\,n\,t\,a\,r\,y\ \ M\,a\,t\,e\,r\,i\,a\,l \par}
\vspace{8mm}

% --- Main Title ---
{\huge Learning Point Cloud Geometry as a Statistical Manifold: \\[2mm]
 Theory and Practice \par}

\vspace*{\fill}

% --- Authors ---
{\large
  Jinwoo Lee\textsuperscript{*,1}\quad
  Jiwoo Kim\textsuperscript{*,1}\quad
  Woojae Shin\textsuperscript{1}\quad
  Giseop Kim\textsuperscript{2}\quad
  Hyondong Oh\textsuperscript{\dag,1}
\par}
\vspace{4mm}

{\normalsize
  \textsuperscript{1}\,Korea Advanced Institute of Science and Technology (KAIST), Daejeon, Republic of Korea\\[1mm]
  \textsuperscript{2}\,Daegu Gyeongbuk Institute of Science and Technology (DGIST), Daegu, Republic of Korea\\[2mm]
  {\small Emails: \texttt{\{jinwoolee, tars0523, oj7987, h.oh\}@kaist.ac.kr}, \texttt{gsk@dgist.ac.kr}}
\par}

\vspace{6mm}

{\normalsize
  \textsuperscript{*}\,Equal contribution.\quad
  \textsuperscript{\dag}\,Corresponding author.\par}

\vspace*{\fill}

\newpage

  % ============= PAGE 2: TOC + List of Figures =============

  \vspace*{5mm}

  \begin{minipage}{0.94\linewidth}

    {\LARGE Table of Contents \par}
    \vspace{2mm}
    {\rule{\linewidth}{0.4pt}}
    \vspace{5mm}

    \renewcommand{\labelenumi}{S\Alph{enumi}.}
    \setlength{\parskip}{0pt}

    \begin{enumerate}
      \setlength{\itemsep}{9pt}

      \item Derivation of Residual Distribution and Likelihood
        \begin{enumerate}
          \renewcommand{\labelenumii}{S\Alph{enumi}.\arabic{enumii}.}
          \setlength{\itemsep}{4pt}
          \item Residual Distribution of $\mathbf{d}_i$
          \item Likelihood Re-Parameterization via $\mathbf{d}_i$
        \end{enumerate}

      \item Certifiably Optimal and Differentiable Weighted Least Square Problem
        \begin{enumerate}
          \renewcommand{\labelenumii}{S\Alph{enumi}.\arabic{enumii}.}
          \setlength{\itemsep}{4pt}
          \item Certifiably Optimal Solution for Weight Least Square Problem
          \item Implicit Function Theorem Under Abadie Constraint Qualification
        \end{enumerate}

      \item Relaxation on Mode Pose Estimation

      \item Experimental Details
        \begin{enumerate}
          \renewcommand{\labelenumii}{S\Alph{enumi}.\arabic{enumii}.}
          \setlength{\itemsep}{4pt}
          \item Object Pose Estimation
          \item LiDAR Odometry
          \item Scan Augmentation: LiDAR Odometry
          \item Scan Augmentation: Global Registration
          \item Computation Time Analysis
        \end{enumerate}

      \item Detailed Discussion on Probabilistic Modeling

      \item Advantages of Ellipsoidal Representation
        \begin{enumerate}
          \renewcommand{\labelenumii}{S\Alph{enumi}.\arabic{enumii}.}
          \setlength{\itemsep}{4pt}
          \item Advantages of Gaussian Ellipsoidal Representation
          \item Scene Completion vs.\ Local Geometric Abstraction
          \item Towards Sensor-Agnostic Geometric Perception
        \end{enumerate}
    \end{enumerate}

    \vspace{10mm}

    {\LARGE List of Figures \par}
    \vspace{2mm}
    {\rule{\linewidth}{0.4pt}}
    \vspace{5mm}

    \renewcommand{\arraystretch}{1.6}
    \begin{tabular}{@{} l@{\hspace{1.5em}} p{0.82\linewidth} @{}}
      Fig.\ SC1. & Comparison of Optimization Trajectories Between Relaxed and Exact Methods \\
      Fig.\ SD1. & Evaluation of LiDAR Odometry Performance \\
      Fig.\ SE1. & Runtime Comparison \\
      Fig.\ SE1. & Probabilistic Modeling Comparison \\
      Fig.\ SE2. & Object Pose Estimation Comparison Between T-Q-C and T-T-Q-C \\
      Fig.\ SE3. & Scene Augmentation Between T-Q-C and T-T-Q-C \\
      Fig.\ S1.  & Ellipsoidal Geometric Structure Estimation Using POLI \\
      Fig.\ S2.  & Localization and Mapping With POLI-Augmented Scan \\
      Fig.\ S3.  & Global Registration With POLI-Augmented Scan \\
      Fig.\ S4.  & Data Augmentation Using POLI \\
    \end{tabular}

  \end{minipage}

  \vfill
\end{titlepage}

\clearpage
\newpage
\section{Derivation of Residual Distribution and Likelihood Re-parameterization}
\vspace{+3mm}
\label{app:res2corr}
\subsection{Residual distribution of $\mathbf d_i$}
\vspace{+3mm}
\label{app:di_condition}

Following Section~\ref{sec:probstate}, we have:
\begin{align*}
\mathbf{p}_i 
&\sim \mathcal{N}\!\Big(
\mathbf{R}_p^\top(\bar{\mathbf{x}}_{\mathbf{g}_i}-\mathbf{t}_p),\ 
\mathbf{R}_p^\top\mathbf{C}_{\mathbf{g}_i}\mathbf{R}_p
\Big),
\\
\mathbf{q}_i 
&\sim \mathcal{N}\!\Big(
\mathbf{R}_q^\top(\bar{\mathbf{x}}_{\mathbf{g}_i}-\mathbf{t}_q),\ 
\mathbf{R}_q^\top\mathbf{C}_{\mathbf{g}_i}\mathbf{R}_q
\Big).
\end{align*}
Define the residual $\mathbf d_i$ as:
\begin{align*}
\mathbf{d}_i 
&\triangleq \mathbf{q}_i-\mathbf{T}\mathbf{p}_i .
\end{align*}
\noindent
Using the Gaussian perturbation forms:
\begin{align*}
\mathbf p_i &= \mathbf{R}_p^\top(\bar{\mathbf{x}}_{\mathbf{g}_i}-\mathbf{t}_p)+\boldsymbol\epsilon_{\mathbf{p}_i},
\\
\mathbf q_i &= \mathbf{R}_q^\top(\bar{\mathbf{x}}_{\mathbf{g}_i}-\mathbf{t}_q)+\boldsymbol\epsilon_{\mathbf{q}_i},
\end{align*}
we obtain
\begin{align*}
\mathbf{d}_i
&=
\Big(\mathbf{R}_q^\top(\bar{\mathbf{x}}_{\mathbf{g}_i}-\mathbf{t}_q)
+\boldsymbol\epsilon_{\mathbf{q}_i}\Big)
-\mathbf{T}\Big(\mathbf{R}_p^\top(\bar{\mathbf{x}}_{\mathbf{g}_i}-\mathbf{t}_p)
+\boldsymbol\epsilon_{\mathbf{p}_i}\Big)
=
\underbrace{
\mathbf{R}_q^\top(\bar{\mathbf{x}}_{\mathbf{g}_i}-\mathbf{t}_q)
-\mathbf{T}\mathbf{R}_p^\top(\bar{\mathbf{x}}_{\mathbf{g}_i}-\mathbf{t}_p)
}_{=\mathbf{0}\ \ (\mathbf{T}=\mathbf{T}_q^{-1}\mathbf{T}_p)}
+\boldsymbol\epsilon_{\mathbf{q}_i}-\mathbf{R}\boldsymbol\epsilon_{\mathbf{p}_i},
\end{align*}
where $\mathbf{R}\triangleq \mathbf{T}[:3,:3]=\mathbf{R}_q^\top\mathbf{R}_p$.
Since $\boldsymbol\epsilon_{\mathbf{p}_i}$ and $\boldsymbol\epsilon_{\mathbf{q}_i}$ are independent
zero-mean Gaussian perturbations with covariances
$\mathbf{R}_p^\top\mathbf{C}_{\mathbf{g}_i}\mathbf{R}_p$ and
$\mathbf{R}_q^\top\mathbf{C}_{\mathbf{g}_i}\mathbf{R}_q$, respectively,
we have:
\begin{align*}
\mathbf d_i
&\sim \mathcal{N}\!\Big(
\mathbf 0,\ 
\mathbf{R}_q^\top\mathbf{C}_{\mathbf{g}_i}\mathbf{R}_q
+\mathbf{R}\big(\mathbf{R}_p^\top\mathbf{C}_{\mathbf{g}_i}\mathbf{R}_p\big)\mathbf{R}^\top
\Big).
\end{align*}
Moreover, using $\mathbf R=\mathbf R_q^\top \mathbf R_p$, we obtain:
\begin{align*}
\mathbf{R}\big(\mathbf{R}_p^\top\mathbf{C}_{\mathbf{g}_i}\mathbf{R}_p\big)\mathbf{R}^\top
&=
\mathbf R_q^\top \mathbf R_p
\big(\mathbf{R}_p^\top\mathbf{C}_{\mathbf{g}_i}\mathbf{R}_p\big)
\mathbf R_p^\top \mathbf R_q
=
\mathbf R_q^\top \mathbf C_{\mathbf g_i}\mathbf R_q.
\end{align*}
Therefore, the residual distribution becomes:
\begin{equation}
\mathbf d_i 
\sim \mathcal{N}\!\Big(
\mathbf 0,\ 
2\,\mathbf R_q^\top \mathbf C_{\mathbf g_i}\mathbf R_q
\Big)
=
\mathcal{N}\!\big(\mathbf 0,\ 2\mathbf C_{\mathbf q_i}\big).
\label{eq:di_gaussian}
\end{equation}

\vspace{+3mm}
\subsection{Likelihood re-parameterization via $\mathbf d_i$}
\vspace{+3mm}
\label{app:reparam}

Fix $\mathbf p_i$ and $\mathbf T=(\mathbf R,\mathbf t)$.
Define the residual:
\begin{align*}
\mathbf d_i 
\triangleq \mathbf q_i - \mathbf T\mathbf p_i
=
\mathbf q_i - (\mathbf R\mathbf p_i+\mathbf t).
\end{align*}
This defines a bijective change of variables in $\mathbf q_i$,
with an inverse map:
\begin{align*}
\mathbf q_i 
=
\mathbf d_i + \mathbf T\mathbf p_i
=
\mathbf d_i + (\mathbf R\mathbf p_i+\mathbf t).
\end{align*}
The Jacobian of $\mathbf d_i$ with respect to $\mathbf q_i$ is:
\begin{align*}
\frac{\partial \mathbf d_i}{\partial \mathbf q_i} = \mathbf I_3,
\qquad
\left|\det\!\left(\frac{\partial \mathbf d_i}{\partial \mathbf q_i}\right)\right| = 1.
\end{align*}
By the change-of-variables formula for probability densities, we have the re-parameterization:
\begin{align*}
p(\mathbf q_i\mid \mathbf p_i,\mathbf T,\mathbf C_{\mathbf q_i})
&=
p(\mathbf d_i\mid \mathbf p_i,\mathbf T,\mathbf C_{\mathbf q_i})
\left|\det\!\left(\frac{\partial \mathbf d_i}{\partial \mathbf q_i}\right)\right|
\nonumber\\
&=
p(\mathbf d_i\mid \mathbf p_i,\mathbf T,\mathbf C_{\mathbf q_i}).
\end{align*}
Following the Eq.~\eqref{eq:di_gaussian}, the above identity implies the re-parameterization:
\begin{align*}
p(\mathbf q_i\mid \mathbf p_i,\mathbf T,\mathbf C_{\mathbf q_i})
\equiv
p(\mathbf d_i\mid \mathbf T,\mathbf C_{\mathbf q_i}).
\end{align*}

\clearpage
\newpage
\section{Certifiably Optimal and Differentiable Weighted Least Square Problem}
\vspace{+3mm}
\label{app:optimality}
In this section, we provide a detailed description of the SDPRLayer~\citep{holmes2025sdprlayers}, specifically addressing how to obtain the globally optimal solution for the relaxed problem in Eq.~\eqref{eq:reduced_estimator} and how to derive the corresponding gradients. 

\vspace{+3mm}
\subsection{Certifiably Optimal Solution for Weight Least Square Problem} 
\vspace{+3mm}
The optimization problem from Eq.~\eqref{eq:reduced_estimator} must (\emph{i}) guarantee a \emph{globally} optimal solution and (\emph{ii}) allow computing the gradient with respect to the weight matrices $\mathbf{W}_{i}\triangleq(2\mathbf{C}_{\mathbf{q}_i})^{-1}$ at that optimum:
\begin{align*}
\hat{\mathbf{T}}
\;=\;
& \arg\min_{\mathbf{T}}\sum_{i=1}^N \mathbf{d}^\top_{i} \mathbf{W}_{i}\mathbf{d}_{i}.
\end{align*}
By encoding the rotation constraints as polynomial equalities, we can obtain a homogeneous quadratically constrained quadratic program (QCQP):
\begin{align*}
\tag{QCQP}\label{prob:primal_qcqp}
\begin{aligned}
\min_{\mathbf{x}\in\mathbb{R}^{13}}
\quad &
\mathbf{x}^\top \mathbf{Q}\,\mathbf{x}
\\[2pt]
\text{s.t.}\quad &
\mathbf{x}^\top \mathbf{A}_\ell \mathbf{x} \;=\; 0,
\qquad \ell=1,\dots,m,
\\
&
\mathbf{x}^\top \mathbf{A}_0 \mathbf{x} \;=\; 1.
\end{aligned}
\end{align*}
This formulation is composed of the following elements (for a detailed derivation, we refer the reader to~\citep{briales2017convex}):
\[
\mathbf{x}
=
\begin{bmatrix}
h \\
\mathrm{vec}(\mathbf{R}) \\
\mathbf{t}
\end{bmatrix}
\in\mathbb{R}^{13},
\qquad
h\in\mathbb{R},\;\mathbf{R}\in{SO(3)},\;\mathbf{t}\in\mathbb{R}^{3}.
\]
% \begin{align*}
% \mathbf{Q}_{i}
% &=
% \begin{bmatrix}
% \; \mathbf{q}_{i}^\top\mathbf{W}_{i}\mathbf{q}_{i} & -\bigl(\mathbf{p}_i\otimes \mathbf{W}_{i}\mathbf{q}_{i}\bigr)^\top & \bigl(\mathbf{W}_{i}\mathbf{q}_{i}\bigr)^\top \\
% -\bigl(\mathbf{p}_i\otimes \mathbf{W}_{i}\mathbf{q}_{i}\bigr) & \; (\mathbf{p}_i\mathbf{p}_i^\top)\otimes \mathbf{W}_{i} & -\,\mathbf{p}_i\otimes \mathbf{W}_{i} \\
% \bigl(\mathbf{W}_{i}\mathbf{q}_{i}\bigr) & -\bigl(\mathbf{p}_i\otimes \mathbf{W}_{i}\bigr)^\top & \; \mathbf{W}_{i}
% \end{bmatrix}.
% \label{eq:Qij_block}
% \end{align*}
% \begin{align*}
% \mathbf{Q}\;=\;\sum_{i=1} ^N\mathbf{Q}_{i}\;\in\;\mathbb{S}^{13}.
% \end{align*}
\begin{align*}
\mathbf{Q}_{i}
&=
\begin{bmatrix}
\mathbf{q}_{i}^\top\mathbf{W}_{i}\mathbf{q}_{i}
&
-\bigl(\mathbf{p}_i\otimes \mathbf{W}_{i}\mathbf{q}_{i}\bigr)^\top
&
\bigl(\mathbf{W}_{i}\mathbf{q}_{i}\bigr)^\top
\\
-\bigl(\mathbf{p}_i\otimes \mathbf{W}_{i}\mathbf{q}_{i}\bigr)
&
(\mathbf{p}_i\mathbf{p}_i^\top)\otimes \mathbf{W}_{i}
&
-\,\mathbf{p}_i\otimes \mathbf{W}_{i}
\\
\bigl(\mathbf{W}_{i}\mathbf{q}_{i}\bigr)
&
-\bigl(\mathbf{p}_i\otimes \mathbf{W}_{i}\bigr)^\top
&
\mathbf{W}_{i}
\end{bmatrix}
\end{align*}

\begin{align*}
\mathbf{Q}
=
\sum_{i=1}^{N}\mathbf{Q}_{i}
\in
\mathbb{S}^{13}.
\end{align*}

% \begin{align}
% \mathbf{r}_i^\top \mathbf{r}_j - \delta_{ij}\,h^2 = 0, \label{eq:orth}\\ 
% i,j\in\{1,2,3\},\textcolor{gray}{\quad\text{(orthonormality)}}\notag  \\
% \sum_{r=1}^3 R_{rj}^2 - \sum_{c=1}^3 R_{ic}^2 = 0,\label{eq:rowcol}\\
% i,j\in\{1,2,3\}, 
% \textcolor{gray}{\quad\text{(row/column length balancing)}}\notag  \\
% \mathbf{r}_j \times \mathbf{r}_k - h\,\mathbf{r}_i = \mathbf{0}, \label{eq:handed}\\
% (i,j,k)\text{ cyclic},\textcolor{gray}{\quad\text{(right-handedness)}}\notag  \\
% \mathbf{x}^\top \mathbf{A}_0 \mathbf{x} \;=\; 1,
% \qquad
% \mathbf{A}_0 \;=\; \mathbf{e}_h \mathbf{e}_h^\top\;\;\;(\text{i.e., } h^2=1).\label{eq:norm}
% \end{align}

% \begin{align}
% \mathbf{r}_i^\top \mathbf{r}_j - \delta_{ij}\,h^2 &= 0
% \label{eq:orth}\\
% \multicolumn{2}{l}{
% \displaystyle i,j\in\{1,2,3\},
% \textcolor{gray}{\quad\text{(orthonormality)}}}
% \notag\\[0.3em]
% %
% \sum_{r=1}^3 R_{rj}^2 - \sum_{c=1}^3 R_{ic}^2 &= 0
% \label{eq:rowcol}\\
% \multicolumn{2}{l}{
% \displaystyle i,j\in\{1,2,3\},
% \textcolor{gray}{\quad\text{(row/column length balancing)}}}
% \notag\\[0.3em]
% %
% \mathbf{r}_j \times \mathbf{r}_k - h\,\mathbf{r}_i &= \mathbf{0}
% \label{eq:handed}\\
% \multicolumn{2}{l}{
% \displaystyle (i,j,k)\text{ cyclic},
% \textcolor{gray}{\quad\text{(right-handedness)}}}
% \notag\\[0.3em]
% %
% \mathbf{x}^\top \mathbf{A}_0 \mathbf{x} &= 1,
% \qquad
% \mathbf{A}_0 = \mathbf{e}_h \mathbf{e}_h^\top
% \;\;\;(\text{i.e., } h^2=1)
% \label{eq:norm}
% \end{align}
\begin{align}
\mathbf{r}_i^\top \mathbf{r}_j - \delta_{ij}\,h^2 &= 0
\label{eq:orth}\\
&\qquad i,j\in\{1,2,3\},
\textcolor{gray}{\text{(orthonormality)}}
\notag\\[0.3em]
\sum_{r=1}^3 R_{rj}^2 - \sum_{c=1}^3 R_{ic}^2 &= 0
\label{eq:rowcol}\\
&\qquad i,j\in\{1,2,3\},
\textcolor{gray}{\text{(row/column length balancing)}}
\notag\\[0.3em]
\mathbf{r}_j \times \mathbf{r}_k - h\,\mathbf{r}_i &= \mathbf{0}
\label{eq:handed}\\
&\qquad (i,j,k)\text{ cyclic},
\textcolor{gray}{\text{(right-handedness)}}
\notag\\[0.3em]
\mathbf{x}^\top \mathbf{A}_0 \mathbf{x} &= 1,
\qquad
\mathbf{A}_0 = \mathbf{e}_h \mathbf{e}_h^\top
\;\;\;(\text{i.e., } h^2=1)
\label{eq:norm}
\end{align}

Here $\mathbf{r}_i, \mathbf{r}_j$ denote the columns of the rotation matrix $\mathbf{R}$, and $R_{rj}, R_{ic}$ are its scalar entries.
Each scalar relation in Eqs.~\eqref{eq:orth}--~\eqref{eq:handed} is of the form $\mathbf{x}^\top \mathbf{A}_\ell \mathbf{x}=0$. This results in a total of $m=24$ rotation constraints: 6 from orthonormality Eq.~\eqref{eq:orth}, 9 from row/column balancing Eq.~\eqref{eq:rowcol}, and 9 from right-handedness Eq.~\eqref{eq:handed} (which consists of 3 vector equations). Each $\mathbf{A}_\ell$ is a suitable matrix in $\mathbb{S}^{13}$ (the set of $13\times 13$ symmetric matrices).
The final constraint Eq.~\eqref{eq:norm} is the single homogeneous normalizer, using $\mathbf{A}_0 = \mathbf{e}_h \mathbf{e}_h^\top$, where $\mathbf{e}_h = [1, 0, \dots, 0]^\top \in \mathbb{R}^{13}$ is the basis vector for $h$, to fix the global scale ($h^2=1$). Let us introduce the lifted variable $\mathbf{X}=\mathbf{x}\mathbf{x}^\top\in\mathbb{S}^{13}$. Dropping the nonconvex rank constraint $\operatorname{rank}(\mathbf{X})=1$ gives Shor’s semidefinite relaxation, which is a semidefinite program (SDP):
\begin{align}
\tag{SDP}
\label{eq:shor_relax}
\begin{aligned}
\min_{\mathbf{X}\in\mathbb{S}^{13}} \quad & \langle \mathbf{Q},\mathbf{X}\rangle \\
\text{s.t.} \quad & \langle \mathbf{A}_\ell,\mathbf{X}\rangle = 0, \quad \ell=1,\dots,m, \\
& \langle \mathbf{A}_0,\mathbf{X}\rangle = 1, \\
& \mathbf{X}\succeq \mathbf{0}.
\end{aligned}
\end{align}
The relaxation is \emph{tight} if the optimal solution $\mathbf{X}^\star$ to \ref{eq:shor_relax} is rank-one. 
In this instance, the globally optimal homogeneous vector $\hat{\mathbf{x}}$ can be exactly recovered via the decomposition $\mathbf{X}^\star=\hat{\mathbf{x}}\hat{\mathbf{x}}^\top$.  Conversely, if $\operatorname{rank}(\mathbf{X}^\star) > 1$, the relaxation is not tight. In such cases, redundant constraints are typically incorporated to strengthen the relaxation and encourage a rank-one solution. Finally, a candidate pose is extracted from the SDP solution via a rounding procedure to ensure feasibility for \ref{prob:primal_qcqp}; this candidate is accepted if the gap between the rounded primal objective and the SDP value is sufficiently small.\\

\subsection{Implicit Function Theorem Under Abadie Constraint Qualification}
In addition to obtaining a globally optimal solution, we must compute gradients for back-propagation through the layer. Consider the \ref{prob:primal_qcqp} with equality constraints
\[
g_\ell(\mathbf{x}) \triangleq \mathbf{x}^\top \mathbf{A}_\ell \mathbf{x}=0,\quad \ell=1,\dots,m,
\qquad
g_0(\mathbf{x}) \triangleq \mathbf{x}^\top \mathbf{A}_0 \mathbf{x}-1=0.
\]
Define the Lagrangian
\[
L(\mathbf{x},\boldsymbol{\lambda};\theta)
\;=\;
\mathbf{x}^\top \mathbf{Q}(\theta)\,\mathbf{x}
\;+\;
\sum_{\ell=0}^{m}\lambda_\ell\, g_\ell(\mathbf{x}),
\qquad
\boldsymbol{\lambda}\in\mathbb{R}^{m+1}.
\]
At a globally optimal primal--dual pair $(\hat{\mathbf{x}},\hat{\boldsymbol{\lambda}})$, the KKT conditions are
\begin{align}
\nabla_{\mathbf{x}} L(\hat{\mathbf{x}},\hat{\boldsymbol{\lambda}};\theta) &= \mathbf{0}, \label{eq:kkt_grad_L}\\
\mathbf{g}(\hat{\mathbf{x}}) &= \mathbf{0}, \label{eq:kkt_constraints}
\end{align}
where $\mathbf{g}(\mathbf{x}) \triangleq [g_0(\mathbf{x}),g_1(\mathbf{x}),\dots,g_m(\mathbf{x})]^\top$.
Since $\mathbf{Q}(\theta)$ and $\mathbf{A}_\ell$ are symmetric, the stationarity condition expands to
\[
\nabla_{\mathbf{x}} L(\mathbf{x},\boldsymbol{\lambda};\theta)
=
2\!\left(\mathbf{Q}(\theta)+\sum_{\ell=0}^{m}\lambda_\ell \mathbf{A}_\ell\right)\mathbf{x}.
\]
Let $\mathbf{z}=[\mathbf{x}^\top,\boldsymbol{\lambda}^\top]^\top$. The KKT conditions
Eqs.~\eqref{eq:kkt_grad_L}--~\eqref{eq:kkt_constraints} can be written compactly as a root-finding system
$\mathcal{K}(\mathbf{z},\theta)=\mathbf{0}$ with
\[
\mathcal{K}(\mathbf{z},\theta)
\;\coloneqq\;
\begin{bmatrix}
\nabla_{\mathbf{x}} L(\mathbf{x},\boldsymbol{\lambda};\theta)\\[2pt]
\mathbf{g}(\mathbf{x})
\end{bmatrix}
=
\begin{bmatrix}
2\!\left(\mathbf{Q}(\theta)+\sum_{\ell=0}^{m}\lambda_\ell \mathbf{A}_\ell\right)\mathbf{x}\\[2pt]
\mathbf{g}(\mathbf{x})
\end{bmatrix}.
\]

Under Abadie’s constraint qualification (ACQ), appropriate second-order sufficient conditions, and smooth dependence of the constraint manifold on $\theta$, the KKT system admits a locally unique differentiable \emph{selection} of solutions in a neighborhood of $(\hat{\mathbf{z}},\theta)$\footnotemark. Applying the implicit function theorem (IFT) to $\mathcal{K}(\mathbf{z},\theta)=\mathbf{0}$ yields
\[
\mathbf{M}\,\mathrm{d}\mathbf{z} = -\,\mathbf{N}\,\mathrm{d}\theta,
\qquad
\mathbf{M} \;=\; \nabla_{\mathbf{z}}\mathcal{K}(\hat{\mathbf{z}},\theta),
\qquad
\mathbf{N} \;=\; \nabla_{\theta}\mathcal{K}(\hat{\mathbf{z}},\theta),
\]
where $\mathbf{M}$ is the Jacobian with respect to the primal--dual variables and $\mathbf{N}$ is the Jacobian with respect to parameters.
Because redundant constraints may be present, we form a full-row-rank sub-Jacobian $\mathbf{G}_r$ by removing linearly dependent rows from the constraint Jacobian. Writing the resulting row-reduced differential KKT system as
\[
\mathbf{M}_r
\begin{bmatrix}\mathrm{d}\mathbf{x}\\ \mathrm{d}\boldsymbol{\lambda}_r\end{bmatrix}
\;=\; -\,\mathbf{N}\,\mathrm{d}\theta,
\qquad
\mathbf{M}_r \;=\; 2\begin{bmatrix}\bar{\mathbf{H}} & \mathbf{G}_r^\top \\ \mathbf{G}_r & \mathbf{0}\end{bmatrix},
\]
we use $\bar{\mathbf{H}}=\mathbf{Q}(\theta)+\sum_{\ell=0}^{m}\hat{\lambda}_\ell \mathbf{A}_\ell$ (the Lagrangian Hessian, also referred to as the certificate matrix) and $\mathbf{G}_r$ (a full-row-rank selection of the constraint Jacobian). The Jacobian of the optimal primal variable with respect to parameters is then
\begin{align}
\frac{\partial \hat{\mathbf{x}}}{\partial \theta}
\;=\;
-\mathbf{P}\,\mathbf{M}_r^\dagger\,\mathbf{N},
\qquad
\mathbf{P}=\begin{bmatrix}\mathbf{I} & \mathbf{0}\end{bmatrix},
\qquad
\mathbf{M}_r^\dagger=\mathbf{M}_r^\top\bigl(\mathbf{M}_r\mathbf{M}_r^\top\bigr)^{-1}.
\label{eq:implicit_grad}
\end{align}

These relations provide the Jacobian of the recovered globally optimal pose parameters (and hence $\hat{\mathbf{T}}$) with respect to the cost parameters $\theta$, enabling end-to-end differentiation through the SDPR layer. Accordingly, we collect all weight matrices into the cost-parameter vector
\(
\theta \triangleq \mathrm{vec}(\mathbf{W}_1,\dots,\mathbf{W}_N)
\)
(equivalently, $\theta\triangleq\{\mathbf{W}_i\}_{i=1}^N$). For any downstream training loss $\mathcal{L}(\hat{\mathbf{T}})$, gradients with respect to an individual weight entry $[\mathbf{W}_i]_{ab}$ follow from the chain rule:
\begin{align*}
\frac{\partial \mathcal{L}}{\partial [\mathbf{W}_i]_{ab}}
=
\frac{\partial \mathcal{L}}{\partial \hat{\mathbf{T}}}\,
\frac{\partial \hat{\mathbf{T}}}{\partial \hat{\mathbf{x}}}\,
\frac{\partial \hat{\mathbf{x}}}{\partial \theta}\,
\frac{\partial \theta}{\partial [\mathbf{W}_i]_{ab}},
\qquad i=1,\dots,N,\;\; a,b\in\{1,2,3\}.
\end{align*}

\footnotetext{The redundant constraints commonly used to tighten the SDP relaxation may violate the linear independence constraint qualification (LICQ). In this setting, ACQ serves as an appropriate regularity condition that guarantees the existence of a differentiable local solution selection, thereby enabling implicit differentiation of the KKT system after row-reducing the constraint Jacobian.}

\setcounter{figure}{0}          
\renewcommand{\thefigure}{SC\arabic{figure}}
\clearpage
\newpage
\section{Relaxation on Mode Pose Estimation}
\vspace{+3mm}
\label{app:approx}

\subsection{Relaxation on Mode Pose Estimation}
In this section, we provide a justification for the relaxation applied to the mode pose estimation in the main paper, i.e., the transition from Eq.~\eqref{eq:map_pose_general} to Eq.~\eqref{eq:reduced_estimator}.
\label{app:relax_mode_pose}

In the inner optimization stage, our goal is to estimate the mode pose by minimizing the correspondence residual together with an explicit pose regularization term. In the main text, the exact inner objective (Eq.~(13)) is written as
\begin{equation*}
\hat{\mathbf T}
=
\arg\min_{\mathbf T}
\sum_{i=1}^{N} 
\mathbf d_i^\top \big(2\mathbf C_{\mathbf q_i}\big)^{-1}\mathbf d_i
\;+\;
\xi(\mathbf T)^\top \Gamma^{-1}\xi(\mathbf T),
\end{equation*}
To simplify the presentation of the relaxation, we introduce abstract variables \(x\) and \(z\), where \(f(x; z)\) denotes the correspondence residual term and \(g(z)\) denotes the pose regularization term. Here, \(x\) corresponds to the covariance parameters and \(z\) represents the transformation. Under this notation, optimization of the exact objective using an alternating training scheme can be expressed as:
\begin{equation}
\begin{aligned}
z_t &= \arg\min_{z}\; f(x_t;z) + g(z), \\
x_{t+1} &= \arg\min_{x}\; f(x;z_t) + g(z_t).
\end{aligned}
\label{eq:exact_update}
\end{equation}
\noindent
Our relaxed inner stage corresponds to eliminating the explicit pose regularization from the \(z\)-update, while retaining it in the \(x\)-update:
\begin{equation}
\begin{aligned}
z_t^{\text{relaxed}} &= \arg\min_{z}\; f(x_t;z), \\
x_{t+1}^{\text{relaxed}} &= \arg\min_{x}\; f(x;z_t^{\text{relaxed}}) + g\!\big(z_t^{\text{relaxed}}\big).
\end{aligned}
\label{eq:relaxed_update}
\end{equation}

\noindent
The key observation is that, when the pose regularization term \( g(z) \) does not substantially alter the mode pose favored by the correspondence residual, replacing Eq.~\eqref{eq:exact_update} with Eq.~\eqref{eq:relaxed_update} yields a similar pose estimate. As a result, the subsequent \( x \)-updates are guided by nearly identical poses, and both the exact and relaxed formulations converge to comparable fixed points. This behavior is illustrated in Fig.~\ref{fig:overall_result}.

In Scenario A (Fig. SC~\ref{fig:top_left}), the optima favored by \( f \) and \( g \) are inconsistent, leading to a significant discrepancy between \( z^\star_f \) (the minimizer of \( f \)) and \( z^\star_{f+g} \) (the minimizer of \( f+g \)); consequently, the relaxed optimization can deviate from the exact solution. In contrast, in Scenario B (Fig. SC~\ref{fig:top_right}), the minimizers of \( f \) and \( f+g \) are well aligned, both favoring \( z \approx 0 \). In this case, the relaxed optimization trajectory closely follows the exact one and converges to the same global optimum.

Empirically, in the neural network training experiment, optimizing the inner problem under the relaxed formulation leads to a monotonic decrease in both the correspondence residual and the pose regularization term throughout training (Fig. SC~\ref{fig:bottom_left}--\ref{fig:bottom_right}). This observation indicates that the pose estimates produced by the relaxed solver remain consistent with those obtained from the exact formulation. We attribute this behavior primarily to the correspondence construction strategy used in our pipeline. Specifically, correspondences are established under a reference pose \( \tilde{\mathbf T} \) via nearest-neighbor matching with distance thresholding. Consequently, correspondences that would only become valid under poses far from \( \tilde{\mathbf T} \) are suppressed a priori, which implicitly constrains the feasible pose space even in the absence of an explicit pose regularization term.

\begin{figure*}[h] 
\captionsetup[subfigure]{labelformat=empty}
    \centering
    \begin{subfigure}[b]{0.49\textwidth}
        \centering
        \includegraphics[width=\textwidth, trim= 0 0 100 0, clip]{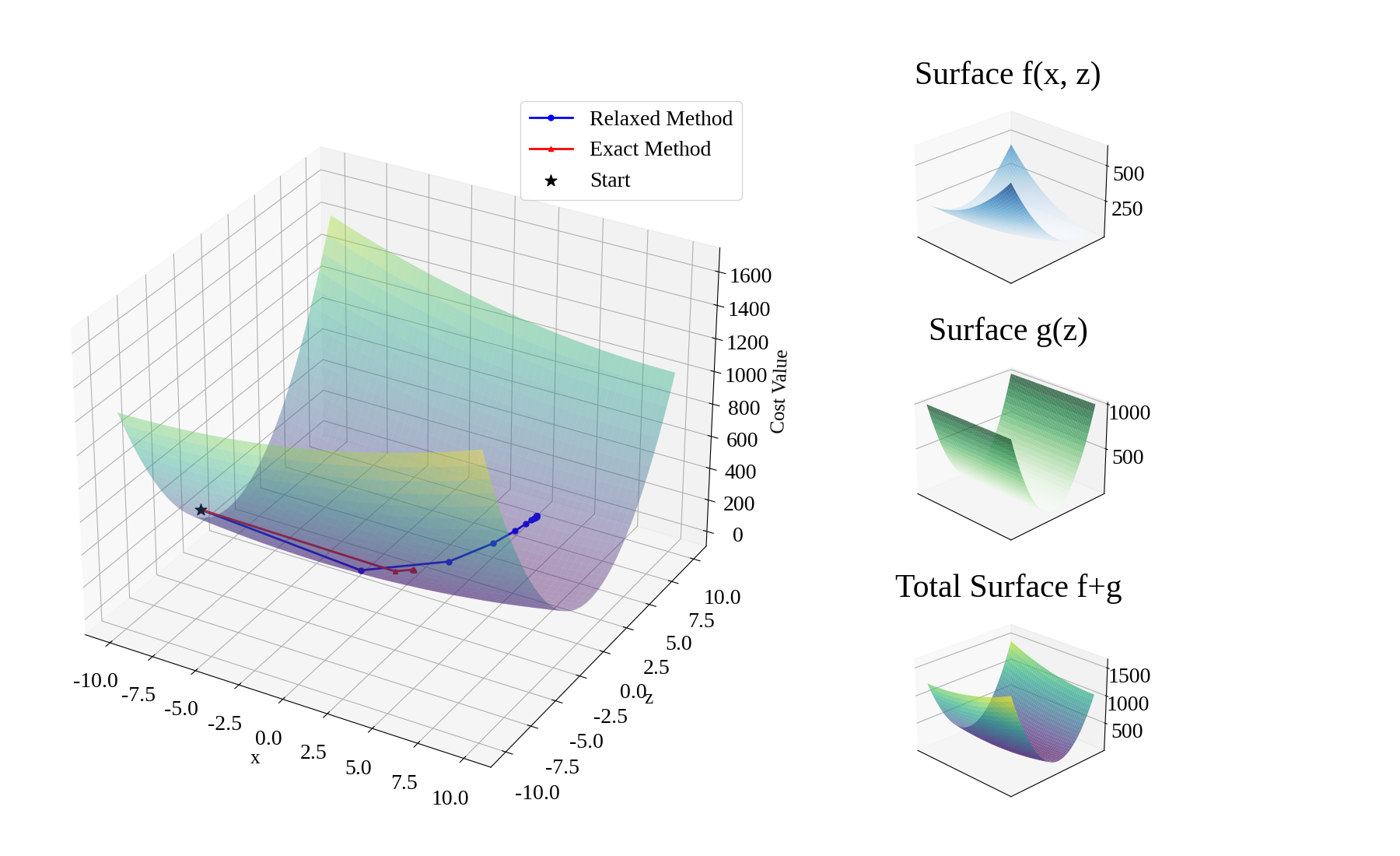}
        \caption{(a) Optimization trajectory 1: scenario A}
        \label{fig:top_left}
    \end{subfigure}
    \begin{subfigure}[b]{0.49\textwidth}
        \centering
        \includegraphics[width=\textwidth, trim= 0 0 100 0, clip]{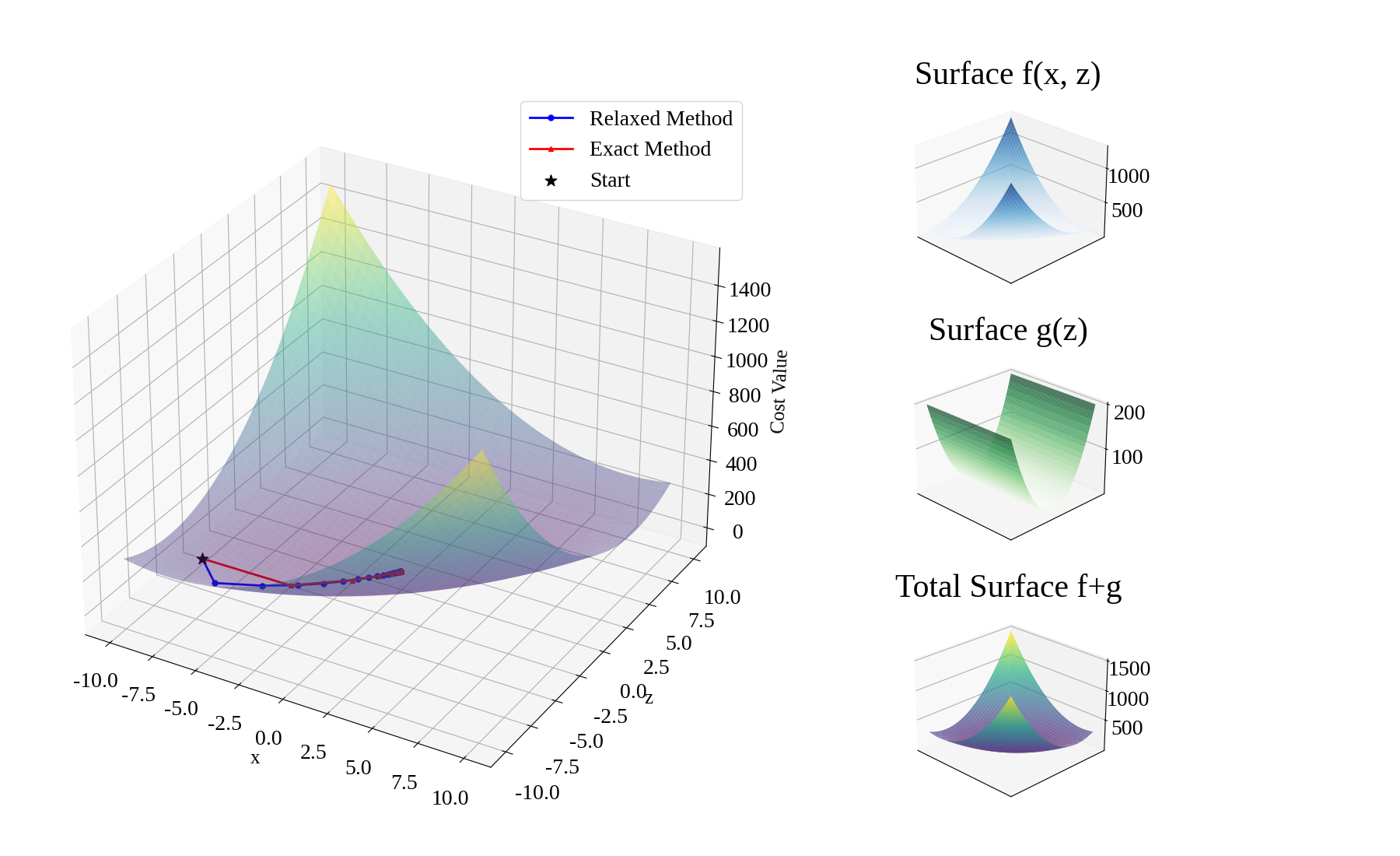}
        \caption{(b) Optimization trajectory 2: scenario B}
        \label{fig:top_right}
    \end{subfigure}

    \vspace{10pt} 

    \begin{subfigure}[b]{0.4\textwidth}
        \centering
        \includegraphics[width=\textwidth]{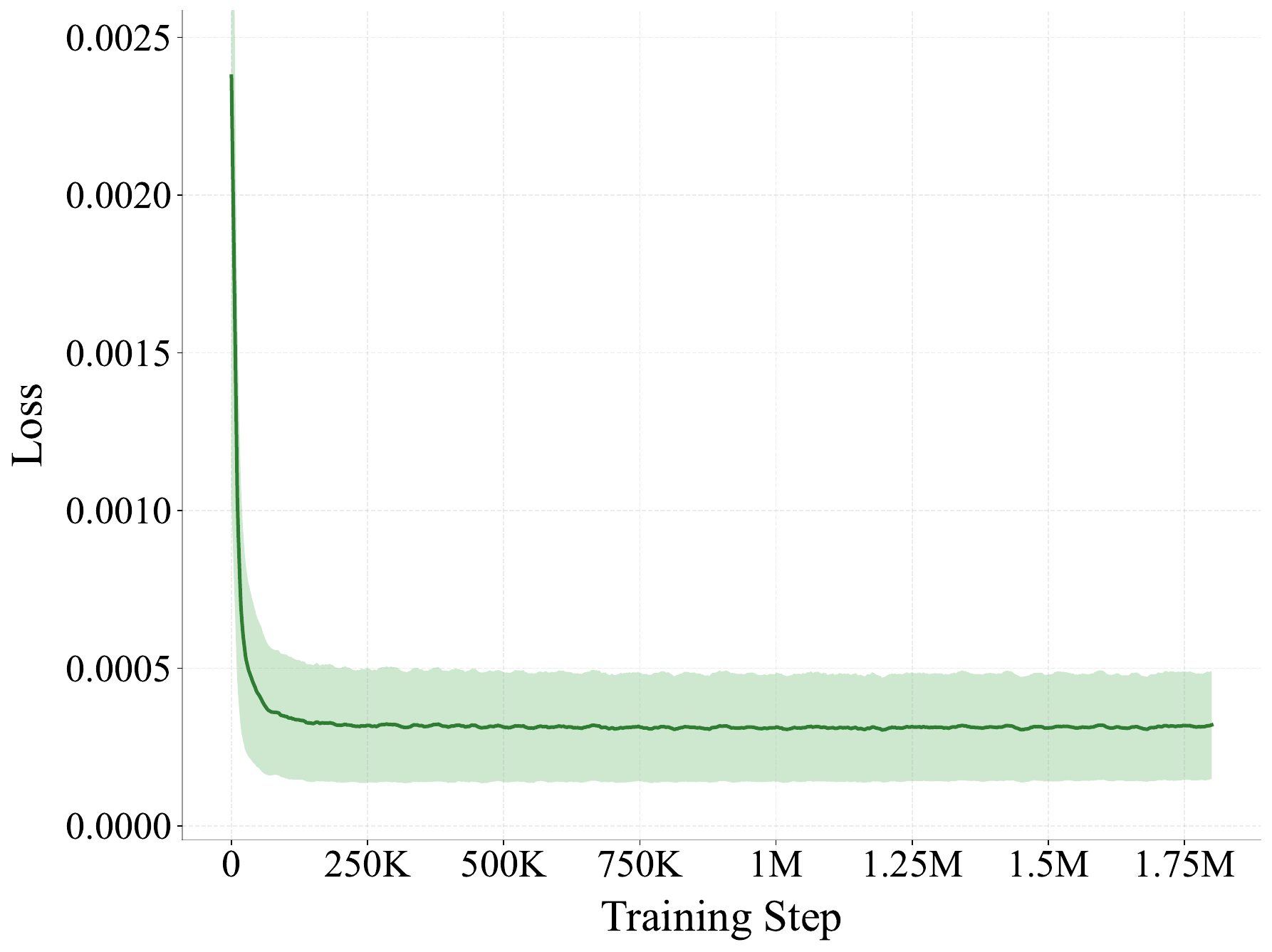}
        \caption{(c) Correspondence loss $f(x, z)$ along training steps}
        \label{fig:bottom_left}
    \end{subfigure}
    \hspace{10mm}
    \begin{subfigure}[b]{0.4\textwidth}
        \centering
        \includegraphics[width=\textwidth]{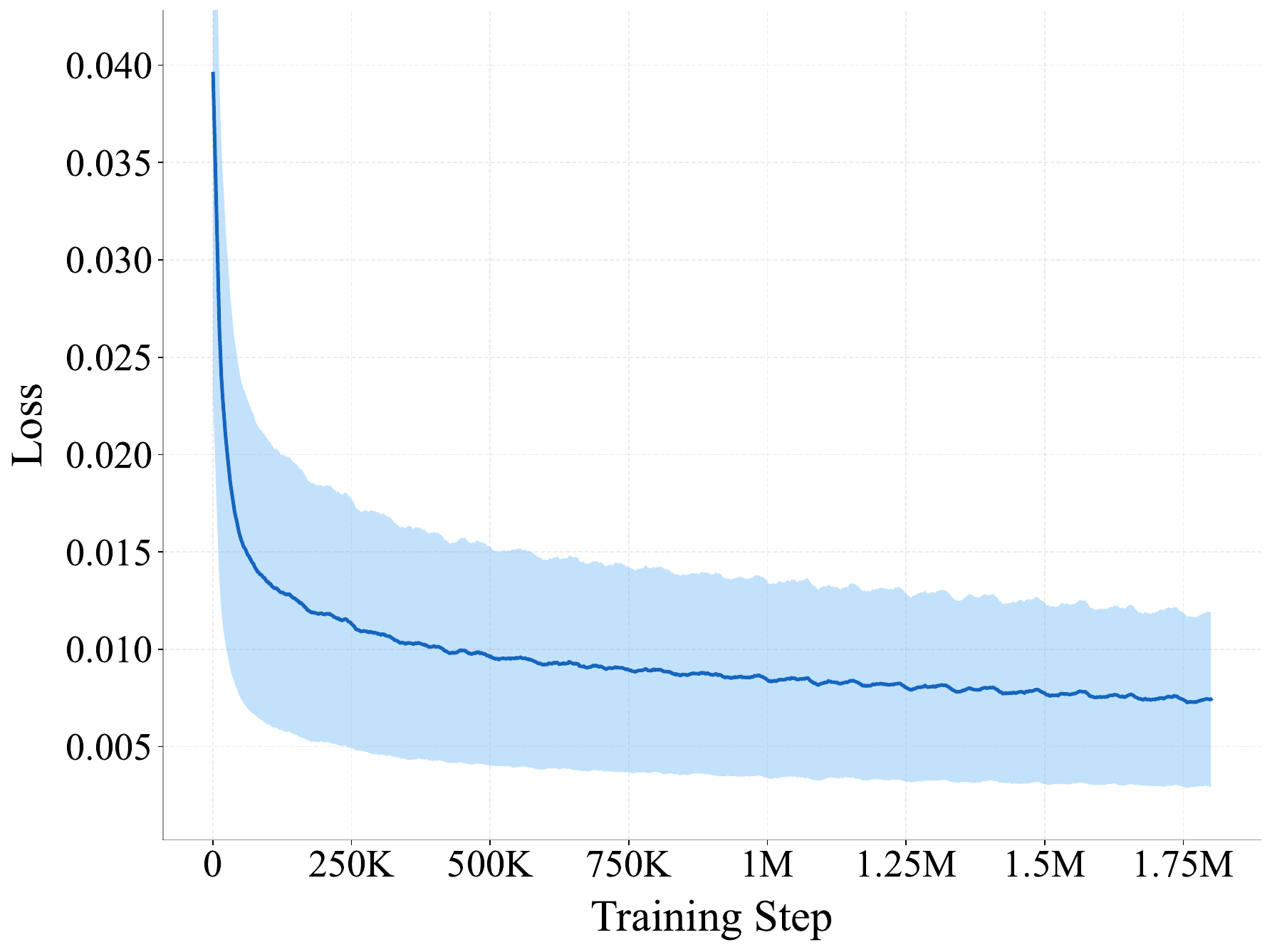}
        \caption{(d) Pose loss $g(z)$ along training steps}
        \label{fig:bottom_right}
    \end{subfigure}

    \caption{Comparison of optimization trajectories between relaxed and exact methods. 
    In (a) Scenario A, where $f(x, z) = (x - z)^2 + (z - 5)^2$ and $g(z) = 10z^2$, the optimal values of $z$ for $f$ and $g$ differ significantly, leading to a large gap between the relaxed and exact methods. 
    In (b) Scenario B, where $f(x, z) = x^2 + 3(z - x)^2$ and $g(z) = 2z^2$, the optimal $z$ values for $f$ and $g$ are consistent at $z \approx 0$, showing that the relaxed method effectively converges to the global optimum. (c) and (d) represent the correspondence loss $f$ and pose loss $g$, respectively. It is observed that both losses gradually decrease when training under the relaxed condition.}
    \label{fig:overall_result}
\end{figure*}

\setcounter{figure}{0}          
\renewcommand{\thefigure}{SD\arabic{figure}}
\setcounter{table}{0}          
\renewcommand{\thetable}{S\arabic{table}}
\clearpage
\newpage
\section{Experimental Details}
\vspace{+3mm}
\label{app:experiment}
\subsection{Object Pose Estimation}

\textbf{Dataset.}
All experiments were conducted using models from the Stanford 3D Scanning Repository~\citep{curless1996volumetric}.
As a training dataset, we selected five object models—\textit{Bunny, Armadillo, Asian Dragon, Lucy, }and \textit{Thai Statue}—and generated synthetic point cloud pairs by independently sampling $N \in \{{500, \text{1,000}}\}$ points from each model. For each object category and for each point count, 300 training samples were generated (i.e., 300 samples for $N=500$ and 300 samples for $N=\text{1,000}$).

For a test dataset, we selected two object models—\textit{Dragon }and \textit{Happy Buddha}—and generated point cloud pairs by independently sampling $N \in \{{300, 400, 500, 600, 700, 800, 900, \text{1,000}}\}$ points from each model. For each object category, 100 samples were generated.

All point clouds were subjected to spherical normalization, after which relative pose transformations were synthesized using random axis–angle rotations with maximum rotation angles of $60^\circ$. Correspondences between scan pairs were established via nearest neighbor search before applying the random axis–angle rotations.

\textbf{Training Setup.}
The network was trained for 10 epochs using the Adam optimizer with a learning rate of $1 \times 10^{-4}$. With a batch size of 1, each epoch required less than 3 minutes. All experiments were performed on a single workstation equipped with an NVIDIA RTX 4060 GPU.

\textbf{Object Pose Estimation with POLI.}
Correspondence estimation using POLI followed the Open3D~\citep{Zhou2018Open3D} FPFH~\citep{rusu2009fast} pipeline. 
Surface normals were computed using POLI, where the normal direction corresponds to the eigenvector of the covariance matrix $\mathbf{C}_{\mathbf{q}}$ associated with its smallest eigenvalue. 
FPFH~\citep{rusu2009fast} features were then extracted using Open3D~\citep{Zhou2018Open3D} with a feature extraction radius of 0.3\;m. 
Inlier correspondences were selected using ROBIN~\citep{shi2021robin} (maximum clique) with a noise bound of 0.1\;m.

\textbf{Baseline Algorithms.}
For the baseline algorithms, KISS-Matcher~\citep{lim2024kiss}, TEASER~\citep{yang2020teaser}, and RPM-Net~\citep{9157132} were used. 
For KISS-Matcher~\citep{lim2024kiss}, a voxel size of 0.03\;m was used, while all other parameters were kept at their default values. 
For TEASER~\citep{yang2020teaser}, we used the same FPFH~\citep{rusu2009fast} radius as above and set the solver parameters to $\texttt{cbar}_{2}=1.0$, $\texttt{GNC}_{factor}=1.4$, and $\texttt{max\mbox{-}iteration}=100$. 
For RPM-Net~\citep{9157132}, the pretrained model provided by the authors was used. 
Due to the scale mismatch introduced by spherical normalization, the normalized point clouds were uniformly rescaled by a factor of 25 before inference. 
This scaling factor was selected empirically as it yielded the best performance among several tested scales.

\textbf{Evaluation Metric.}
A pose estimation was considered successful if the rotation error was less than or equal to $10^\circ$~\citep{lim2024quatro++}.

\clearpage
\newpage 

\subsection{LiDAR Odometry}
\label{app:lo_training_setup}
\textbf{Dataset.}
We used the HeLiPR~\citep{jung2024helipr} dataset for training. 
The dataset was collected using two sensors: an Ouster OS2-128, which provides up to approximately $260,000$ points per scan (10Hz), and a Velodyne VLP-16, which provides up to approximately $30,000$ points per scan (10Hz). 
For both sensors, datasets were constructed with multiple voxel downsampling resolutions. 
Specifically, three voxel sizes were used for each sensor: 0.2\,m, 0.5\,m, and 1.0 for VLP-16, and 1.0\,m, 1.5\,m, and 2.0\,m for OS2-128. 
For training, the \textit{Roundabout, Town, Riverside,} and \textit{Bridge} sequences were used, while the \textit{KAIST} and \textit{DCC} sequences were reserved for evaluation.

During preprocessing, only the 3D coordinates of points were used without intensity information. 
All point coordinates were normalized by dividing by 100 and represented in the native sensor coordinate frame. 
No maximum range filtering or additional point removal was applied. 
Consecutive LiDAR scans were used as point cloud pairs for training. 
Correspondences were obtained by first aligning the two scans using the GNSS–INS transformation and then performing nearest neighbor search.

\textbf{Training Setup.}
Separate models were trained for each sensor–voxel configuration, resulting in a total of six models. 
Training was performed with a batch size of 1, and each epoch required approximately 1.5 hours. 
For the VLP-16 dataset, the models were trained for 12, 12, and 21 epochs for voxel sizes of 0.2\,m, 0.5\,m, and 1.0\,m, respectively. 
For the OS2-128 dataset, the models were trained for 11, 11, and 14 epochs for voxel sizes of 1.0\,m, 1.5\,m, and 2.0\,m, respectively. 
All models were optimized using the Adam optimizer with a learning rate of $1 \times 10^{-4}$. 
All experiments were conducted on a single workstation equipped with an NVIDIA RTX 4080 SUPER GPU.

\textbf{LiDAR Odometry with POLI Augmentation.}
To perform LiDAR odometry with the trained model, we integrated POLI into a GICP~\citep{segal2009generalized}-based framework. 
Using the covariance predicted by POLI, correspondences obtained via nearest neighbor search were refined through iterative Mahalanobis distance minimization, resulting in the final LiDAR odometry pipeline.

\textbf{Baseline Algorithms.}
For comparison, ICP-based methods including Point-to-Point ICP~\citep{besl1992icp}, Point-to-Plane ICP~\citep{rusinkiewicz2001efficient}, GICP~\citep{segal2009generalized}, and VGICP~\citep{koide2021vgicp} were evaluated under the same preprocessing and voxelization settings. 
In addition, two state-of-the-art odometry systems were tested for each sensor. 
For VLP-16, A-LOAM~\citep{zhang2014loam} and F-LOAM~\citep{wang2021floam} were used, while for OS2-128, KISS-ICP~\citep{vizzo2023kiss} and GenZ-ICP~\citep{lee2024genz} were evaluated. 
Since these systems employ their own downsampling strategies, raw point clouds were provided as an input to ensure fair comparison.

\textbf{Evaluation Metric.}
Odometry performance was measured using the relative pose error~\citep{grupp2017evo} (RPE) metric reported in the main text. 

\textbf{Result Summary.}
As shown in Fig.~\ref{fig:odom_comparison}, the proposed method consistently outperformed classical ICP--based approaches across all evaluation metrics, while achieving higher translation accuracy and comparable rotation accuracy relative to dedicated LiDAR odometry systems.
Despite being evaluated on unseen environments (\textit{DCC} and \textit{KAIST}), POLI+GICP demonstrated consistent performance, demonstrating its generalizability.
These results indicate that estimating local geometry with POLI and using it as a metric for scan matching is highly beneficial.

\clearpage
\newpage 

\begin{figure*}[h]
    \centering

    % % ---------- Header ----------
    % \begin{tabular}{p{0.24\textwidth} p{0.24\textwidth} p{0.24\textwidth} p{0.24\textwidth}}
    %     \multicolumn{2}{c}{VLP-16} &
    %     \multicolumn{2}{c}{OS2-128} \\[2pt]
    %     \centering Rotation (deg) &
    %     \centering Translation (m) &
    %     \centering Rotation (deg) &
    %     \centering Translation (m)
    % \end{tabular}
    % ---------- Header ----------
    \begin{tabular}{p{0.24\textwidth} p{0.24\textwidth} p{0.24\textwidth} p{0.24\textwidth}}
        \multicolumn{2}{c}{\hspace{5mm}VLP-16} &
        \multicolumn{2}{c}{\hspace{-9mm}OS2-128} \\[2pt]
        \centering \hspace{7mm} Rotation (deg) &
        \centering \hspace{4mm}Translation (m) &
        \centering \hspace{-5mm} Rotation (deg) &
        \centering \hspace{-10mm} Translation (m)

    \end{tabular}
    % ---------- Images ----------
    \includegraphics[
        height=20cm,
        keepaspectratio,
        trim=0 -65 0 -25,
        clip
    ]{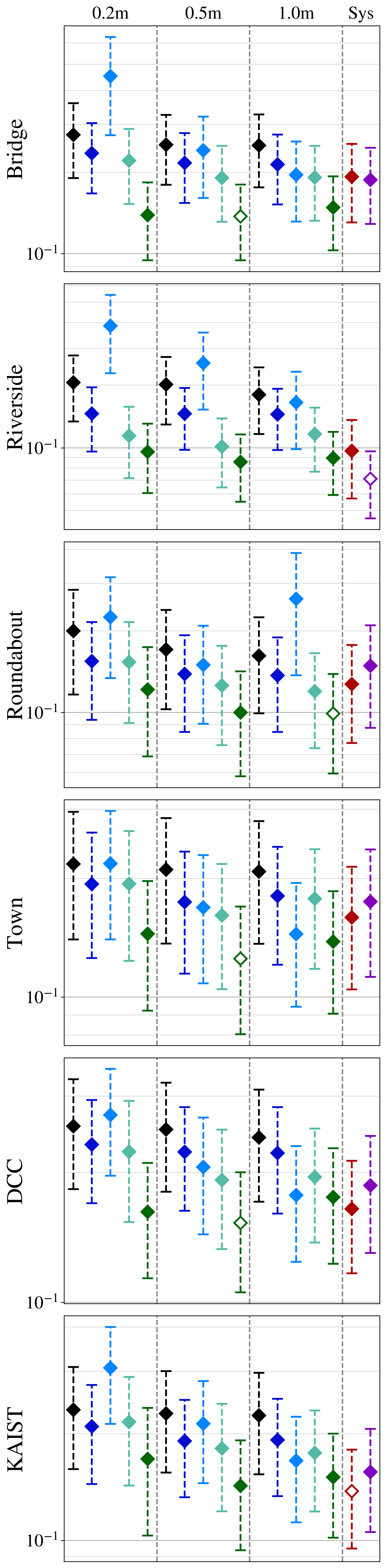}
    \includegraphics[
        height=20cm,
        keepaspectratio,
        trim=30 -65 0 -25,
        clip
    ]{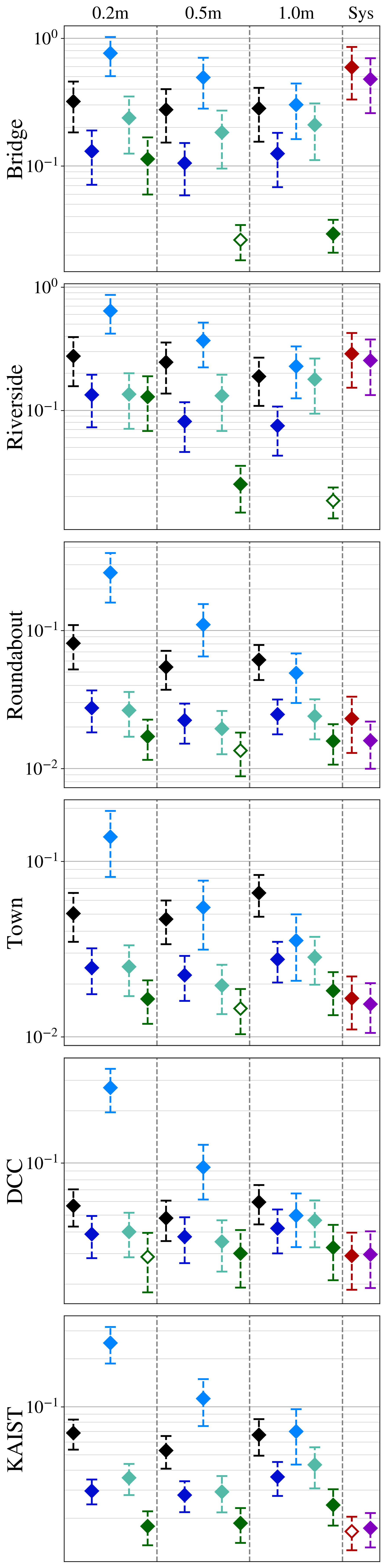}
    \includegraphics[
        height=20cm,
        keepaspectratio,
        trim=30 -65 0 -25,
        clip
    ]{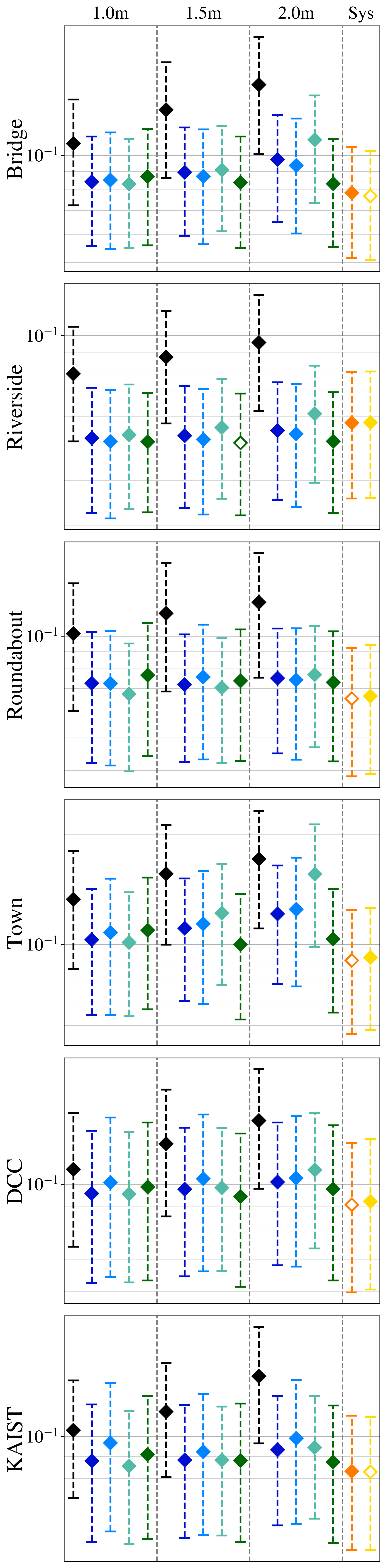}
    \includegraphics[
        height=20cm,
        keepaspectratio,
        trim=30 -65 0 -25,
        clip
    ]{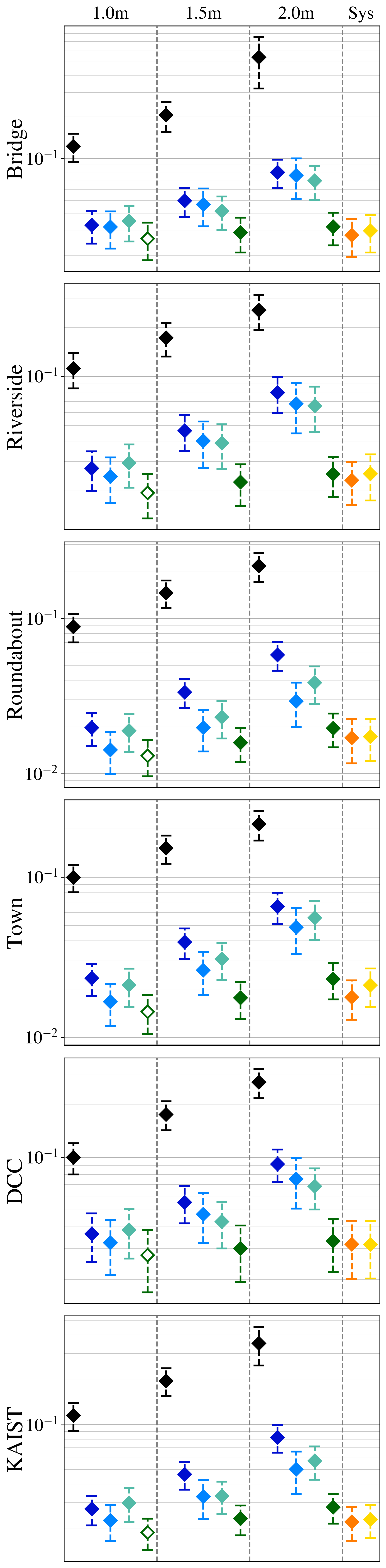}

    \caption[Evaluation of LiDAR odometry performance (VLP-16 / OS2-128)]{
    Evaluation of LiDAR odometry performance using VLP-16 and OS2-128 sensors. Performance is assessed using root mean square error values with error bars representing $\pm 0.5$ standard deviations of RPE (m). For each sensor and sequence, three voxel sizes are evaluated, comparing five ICP variants (\textcolor[RGB]{0,0,0}{\textbf{Point-to-Point}~\protect\citep{besl1992icp}}, \textcolor[RGB]{0, 14, 207}{\textbf{Point-to-Plane}~\protect\citep{rusinkiewicz2001efficient}}, \textcolor[RGB]{0,132,255}{\textbf{GICP}~\protect\citep{segal2009generalized}}, \textcolor[RGB]{82,186,167}{\textbf{VGICP}~\protect\citep{koide2021vgicp}}, and \textcolor[RGB]{0,103,4}{\textbf{POLI+GICP}}). In addition, two LiDAR odometry systems are compared for each sensor (\textcolor[RGB]{173,0,0}{\textbf{A-LOAM}~\protect\citep{zhang2014loam}} and \textcolor[RGB]{129,0,189}{\textbf{F-LOAM}~\protect\citep{wang2021floam}} for VLP-16; \textcolor[RGB]{255,123,0}{\textbf{GenZ-ICP}~\protect\citep{lee2024genz}} and \textcolor[RGB]{247,181,0}{\textbf{KISS-ICP}~\protect\citep{vizzo2023kiss}} for OS2-128). The RMSE of the best-performing method is highlighted by hollow diamond for each sequence.}
    \label{fig:odom_comparison}
\end{figure*}

\clearpage
\newpage

\subsection{Scan Augmentation: LiDAR Odometry}

\textbf{Dataset.}
We conduct experiments on two LiDAR scan datasets: HeLiPR~\citep{jung2024helipr} and KITTI~\citep{geiger2012we}. From HeLiPR~\citep{jung2024helipr}, we use all sequences of data collected with a Velodyne VLP-16 sensor. We additionally evaluate on KITTI~\citep{geiger2012we} odometry sequences 00–10. For this evaluation, the original KITTI~\citep{geiger2012we} scans are converted into a 16-channel LiDAR configuration by selecting one ring every four vertical scan lines.

\textbf{Training Setup.}
POLI directly adopts the weights pretrained in HeLiPR~\citep{jung2024helipr} dataset with a voxel size of 0.2 m.

\textbf{LiDAR Odometry with POLI.}
The trained model predicts point-wise covariance that represents local geometric structures. For each covariance, additional points are sampled to augment the original scan, producing a geometry-enhanced point cloud. The augmented scans are then integrated into the KISS-ICP~\citep{vizzo2023kiss} LiDAR odometry pipeline for pose estimation. Since accurate geometric representation is critical for odometry, point sampling is restricted to highly reliable regions of the Gaussian ellipsoidal distributions estimated by the network. Specifically, seven points are sampled within the 0.05$\sigma$ region of each distribution to reduce noise and improve registration precision.

\textbf{Baseline Algorithms.}
We also construct LiDAR odometry systems using augmented scans obtained from two baselines: principal component analysis (PCA), which is widely used for point cloud geometric estimation, and TULIP~\citep{yang2024tulip}, a range image-based scan densification method.

For the PCA baseline, local covariance is estimated using PCA with the 24 nearest neighbors for each point. Points are then sampled from the resulting Gaussian distribution in the same manner as POLI, i.e., seven points within the 0.05$\sigma$ region. The sampled points are used to augment the original scans and are subsequently integrated into the KISS-ICP~\citep{vizzo2023kiss}LiDAR odometry pipeline.

For TULIP~\citep{yang2024tulip} we use the publicly released pretrained model trained on KITTI~\citep{geiger2012we}. Following the original pipeline, LiDAR scans are converted into 16 × 1024 range images and fed into the network. The network outputs 64 × 1024 range images, which are reprojected into 3D point clouds. The resulting points are likewise incorporated into the LiDAR odometry system for pose estimation.

\textbf{Evaluation Metric.}
We evaluate all methods using standard LiDAR odometry metrics by comparing trajectory estimation performance across all HeLiPR~\citep{jung2024helipr} sequences and KITTI~\citep{geiger2012we} sequences 00–10. Specifically, we evaluate RPE with a delta of 100 frames.

\subsection{Scan Augmentation: Global Registration}

\textbf{Dataset.}
We evaluate global registration on both HeLiPR~\citep{jung2024helipr} and KITTI~\citep{geiger2012we}. All six sequences of HeLiPR~\citep{jung2024helipr} are used, and KITTI~\citep{geiger2012we} odometry sequences 00–10 are included for evaluation. For each sequence, 50 point-cloud pairs are randomly sampled with an inter-scan distance of up to 12~m. This results in a total of 300 pairs for HeLiPR~\citep{jung2024helipr} and 550 pairs for KITTI~\citep{geiger2012we}.

\textbf{Training Setup.}
POLI directly adopts the weights pretrained in HeLiPR~\citep{jung2024helipr} dataset with a voxel size of 0.2 m.

\textbf{Registration with POLI Augmentation.}
POLI samples points from high-confidence regions defined by the estimated Gaussian covariance to improve geometric reliability. For HeLiPR~\citep{jung2024helipr} 50 points are sampled within $0.3\sigma$ of each distribution, whereas for KITTI~\citep{geiger2012we}, 40 points are sampled within a tighter $0.1\sigma$ region. The sampled points are used to augment the original scans, and the augmented scans are subsequently integrated into the KISS-Matcher~\citep{lim2024kiss} registration system for transformation estimation.

\textbf{Baseline Algorithms.}
For TULIP~\citep{yang2024tulip}, we use the publicly released pretrained model trained on KITTI~\citep{geiger2012we}. For the PCA baseline, point clouds are first downsampled with a voxel size of 0.2~m, and PCA is performed within a neighborhood radius of 0.5~m to estimate local covariance. To ensure a fair comparison, both the number of sampled points and the sampling regions are set identical to those used in POLI. For both baselines, the resulting augmented points are likewise incorporated into the same registration system. Even with extensive tuning of PCA parameters, POLI consistently outperforms the PCA baseline.

\textbf{Evaluation Protocol.}
Performance is measured by evaluating registration accuracy across all sampled scan pairs from both datasets, enabling a comprehensive comparison under varying environments and sensor configurations.

\subsection{Computation Time}

In robotic perception, real-time performance is not optional but rather a fundamental system requirement.
Perception modules operate within tight control loops and must continuously process streaming sensory data with minimal latency.
This constraint is particularly critical in applications such as localization and mapping, where delayed geometric interpretation can directly degrade state estimation accuracy and overall system stability.

Accordingly, computational efficiency becomes a primary evaluation criterion when selecting or designing perception algorithms.
In this study, we systematically analyze the computation time of several algorithms used for fundamental geometric structure estimation.
The reported runtime corresponds to the sum of the algorithm execution time and the point augmentation time, and both components are jointly visualized.
The augmentation process—Gaussian sampling for PCA and POLI, and 3D reprojection of the range image for TULIP~\citep{yang2024tulip}—was implemented with CUDA acceleration.
In practice, the sampling process from the covariance matrix is highly parallelizable and executes orders of magnitude faster than the dominant computational components, such as neural network inference or PCA eigen-decomposition.
As a result, the augmentation overhead is negligible and does not affect the overall runtime trends.

Specifically, we compare the following methods:
\begin{itemize}
\item PCA
\begin{itemize}
\item CUDA-accelerated GPU implementation
\item CPU-based implementation provided by Open3D~\citep{Zhou2018Open3D}
\end{itemize}
\item TULIP~\citep{yang2024tulip}, a baseline method introduced in prior work
\item POLI, the proposed method
\end{itemize}

To evaluate scalability with respect to input size, experiments were conducted on point clouds containing
5{,}000, 10{,}000, 15{,}000, 20{,}000, 25{,}000, and 30{,}000 points.
For each configuration, the execution time was measured and compared across all methods.
However, since TULIP~\citep{yang2024tulip} relies on a range-image representation with a fixed input resolution of $16 \times 1024$, runtime comparisons were only feasible for point sets containing up to approximately 15{,}000 points.

\begin{figure}[t]
\includegraphics[width=0.95\linewidth, trim = 0 0 0 0, clip]{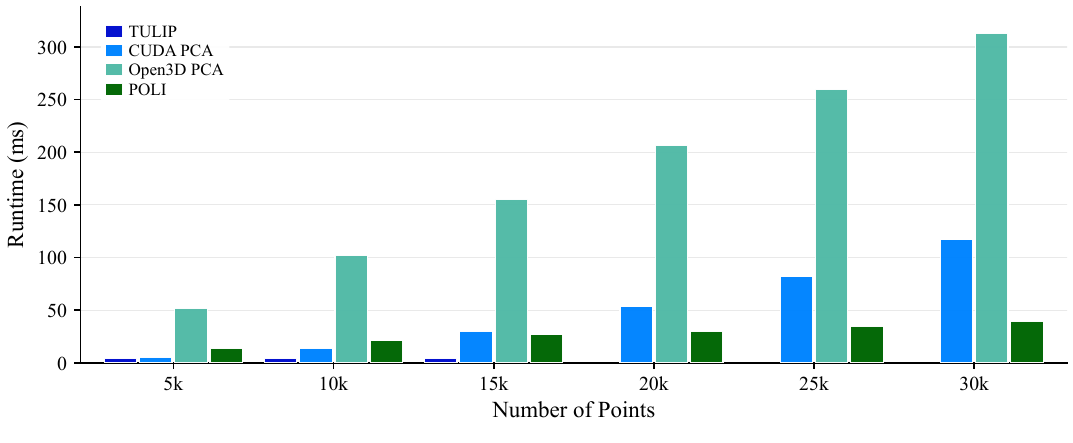}
\caption{Runtime comparison of PCA (CPU/GPU), TULIP~\citep{yang2024tulip}, and the proposed POLI with respect to the number of input points.}
\end{figure}

The results show that TULIP~\citep{yang2024tulip} exhibits strong real-time performance with almost constant execution time across all evaluated cases.
This behavior is primarily attributed to its fixed-resolution range-image–based processing pipeline, which decouples runtime from the number of input points.
However, this design also imposes an inherent limitation: the number of points that can be processed is bounded by the image resolution.
Handling denser point clouds would therefore require increasing the input image size and retraining the network accordingly, reducing scalability and flexibility.

The CUDA-accelerated PCA achieves significantly faster performance than its CPU-based counterpart; however, its runtime increases with the number of points and is eventually surpassed by the proposed POLI method at larger scales.

\setcounter{figure}{0}          
\renewcommand{\thefigure}{SE\arabic{figure}}
\clearpage
\newpage
\section{Detailed Discussion on Probabilistic Modeling}
\vspace{+3mm}
\label{app:modeling}
The overall training objective of the neural network can be decomposed into a pose likelihood term and a correspondence likelihood term:
\begin{align}
 \arg\min_{\mathbf{C}}
\underbrace{
\frac{1}{2}\log\!\big|H(\hat{\mathbf T})\big|
+
\frac{1}{2}\,\xi(\hat{\mathbf{T}})^\top
\Gamma^{-1}
\xi(\hat{\mathbf{T}})
}_{\text{pose likelihood}}
+
\underbrace{
\frac{1}{2}\sum_{i=1}^N
\Big[
\log\!\big|2\mathbf{C}_{i}\big|
+
\mathbf{d}_{i}^\top
(2\mathbf{C}_{i})^{-1}
\mathbf{d}_{i}
\Big]
}_{\text{correspondence likelihood}}.
\tag{MLE of T-T-Q-C}
\end{align}
In this section, we discuss the necessity and validity of our probabilistic formulation (shown in Fig.~\ref{fig:prob_compare}(b), T-T-Q-C), whose MLE objective consists of two components: the correspondence likelihood and the pose likelihood. As shown in the main paper, incorporating the pose likelihood term introduces additional mathematical complexity. Therefore, it is important to justify the necessity of this term. From the above formulation, a natural question arises. If the transformation between $\mathcal{P}$ and $\mathcal{Q}$ is already known exactly, i.e., $\hat{\mathbf{T}} = \tilde{\mathbf{T}}$, the uncertainty in the pose likelihood vanishes, and only the correspondence likelihood remains. In this situation, \emph{would it not be sufficient to optimize only the correspondence likelihood?} Furthermore, even if the transformation is not perfectly accurate but highly reliable (e.g., provided by modern GNSS--INS systems), \emph{could we approximately rely solely on the correspondence likelihood?} Following this line of reasoning, one may alternatively adopt a simplified probabilistic model that treats the transformation as a fixed and known quantity, i.e., $\mathbf{T}=\tilde{\mathbf{T}}$, rather than as a latent variable. In this case, $\mathbf{T}$ is no longer an unknown to be inferred, which substantially simplifies the objective; this corresponds to the formulation shown in Fig.~\ref{fig:prob_compare}(a), denoted as T-Q-C. Motivated by this observation, we conduct comparative experiments between the two probabilistic models, T-Q-C and T-T-Q-C.
In contrast to the MLE objective of T-T-Q-C, the MLE objective of T-Q-C admits a much simpler form, which can be written as follows:
\begin{equation}
\arg\min_{\mathbf{C}}
\frac{1}{2}\sum_{i=1}^N
\Big[
\log |2\mathbf{C}_{i}|
+ \mathbf{d}_i^\top (2\mathbf{C}_{i})^{-1}\mathbf{d}_i
\Big]
\tag{MLE of T-Q-C}
\end{equation}
which corresponds to considering only the correspondence likelihood, without incorporating the pose likelihood. At first glance, this simplification suggests that, given access to the transformation that is close to ground-truth transformation, the learning problem reduces to estimating correspondence covariances alone, making the pose likelihood seemingly unnecessary. To investigate whether this intuition holds in practice, we compare the performance of two strategies (i.e., MLE of T-Q-C and MLE T-T-Q-C) under various conditions.
\begin{figure*}[htbp]
    \centering
    \captionsetup[subfigure]{labelformat=empty}
    \begin{minipage}{0.98\linewidth}
        \centering
        \begin{subfigure}{0.49\linewidth}
            \centering
            \includegraphics[width=\linewidth,keepaspectratio,trim= 0 400 0 400,clip]{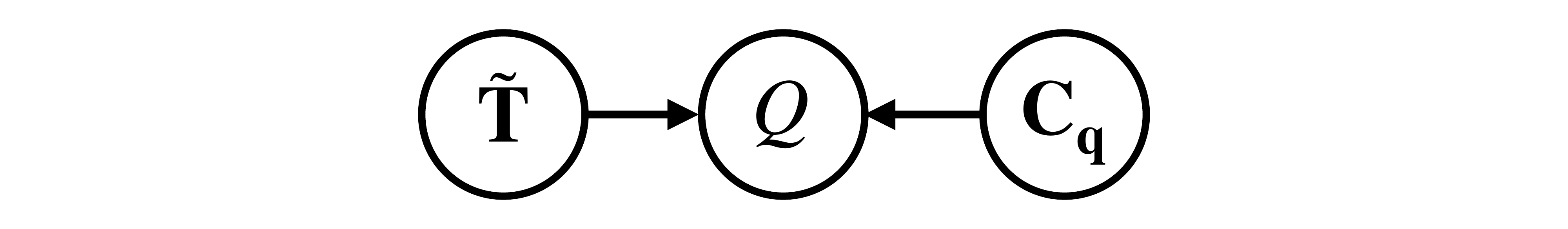}
            \caption{(a) Directed Acyclic Graph of T-Q-C Formulation}
            \label{fig:5}
        \end{subfigure}
        \hfill
        \begin{subfigure}{0.49\linewidth}
            \centering
            \includegraphics[width=\linewidth,keepaspectratio,trim=0 400 0 400,clip]{figures/factor.pdf}
            \caption{(b) Directed Acyclic Graph of T-T-Q-C Formulation}
            \label{fig:30}
        \end{subfigure}
    \end{minipage}
    \caption{(a) corresponds to a probabilistic model, referred to as T–Q–C, that assumes an exact (noise-free) transformation, whereas (b) corresponds to a probabilistic model, referred to as T–T–Q–C, that does not assume an exact transformation.}
    \label{fig:prob_compare}
\end{figure*}

\textbf{Case 1: Exact ground-truth transformation available.}
We use dense and highly accurate point clouds from the Stanford 3D Scanning Repository~\citep{curless1996volumetric} and randomly sample a fixed number of points. This sampling procedure is repeated to generate two independent point clouds sharing the same underlying geometry. 
Arbitrary rigid transformations are applied to both scans to create a pair of point clouds
$(\mathcal{P}, \mathcal{Q})$, with the relative rigid transformation between them serving as the known
ground-truth transformation $\mathbf{T}^*$.
Two models are trained on this data: one using only the correspondence (i.e., MLE of T-Q-C) and the other using both likelihood terms (i.e., MLE of T-T-Q-C). During evaluation, the exact transformation $\mathbf{T}^*$ is used for likelihood computation. Result is shown in Fig.~\ref{fig:shooooot}(a).

\textbf{Case 2: Inaccurate transformation available.}
We follow the same procedure as above, except that the transformation used during likelihood computation is perturbed by injecting zero-mean Gaussian noise into the translation component of $\mathbf{T}^*$, with standard deviations of 0.01, 0.02, and 0.03 for Fig.~\ref{fig:shooooot}(b), (c), and (d), respectively. This setting reflects practical robotic scenarios where only noisy pose measurements are available rather than true ground-truth transformations.

\begin{figure*}[t]
    \centering
    \captionsetup[subfigure]{labelformat=empty}

    \begin{subfigure}{0.24\linewidth}
        \centering
        \includegraphics[width=\linewidth]{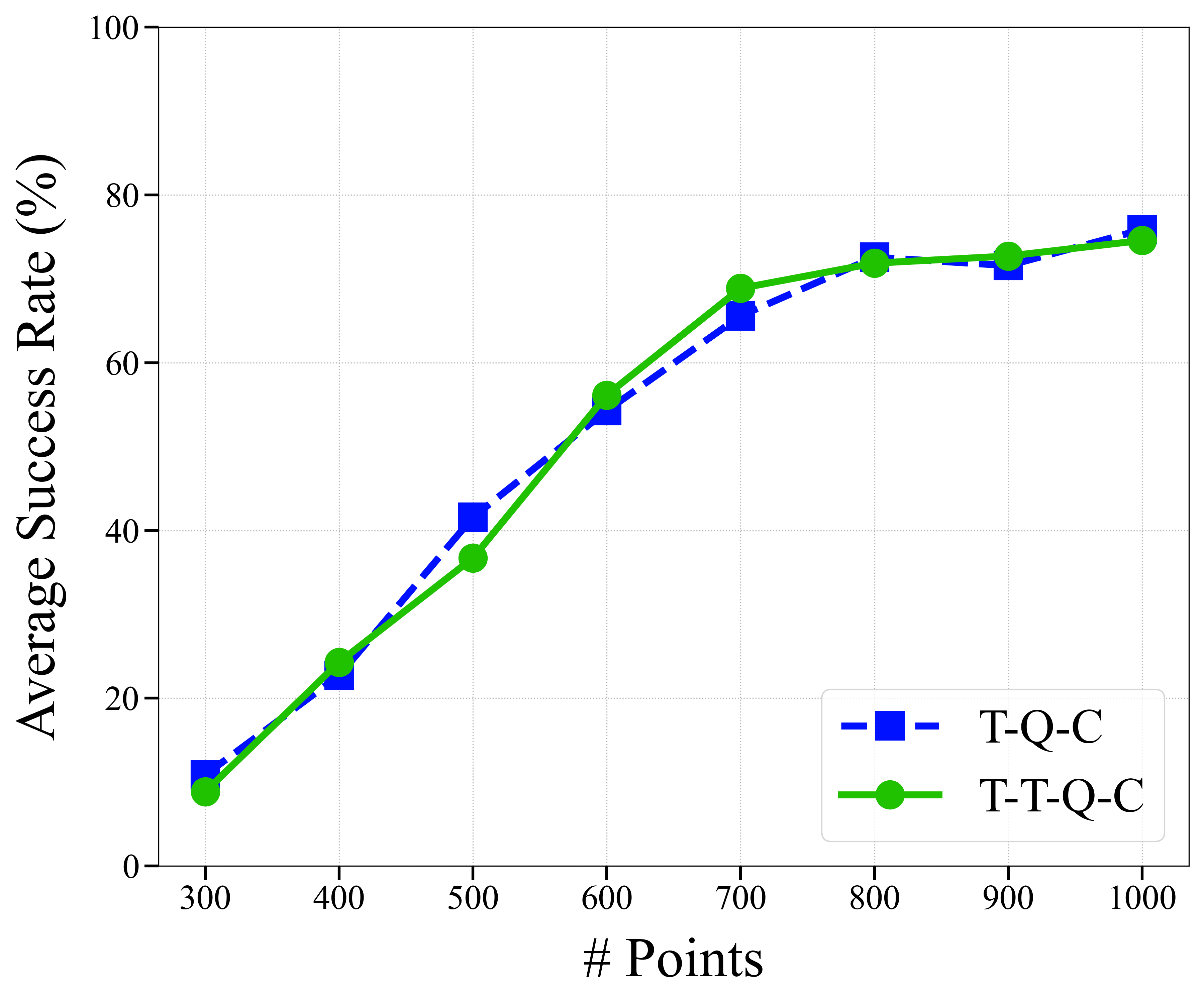}
        \caption{(a) With exact ground-truth}
    \end{subfigure}
    \hfill
    \begin{subfigure}{0.24\linewidth}
        \centering
        \includegraphics[width=\linewidth]{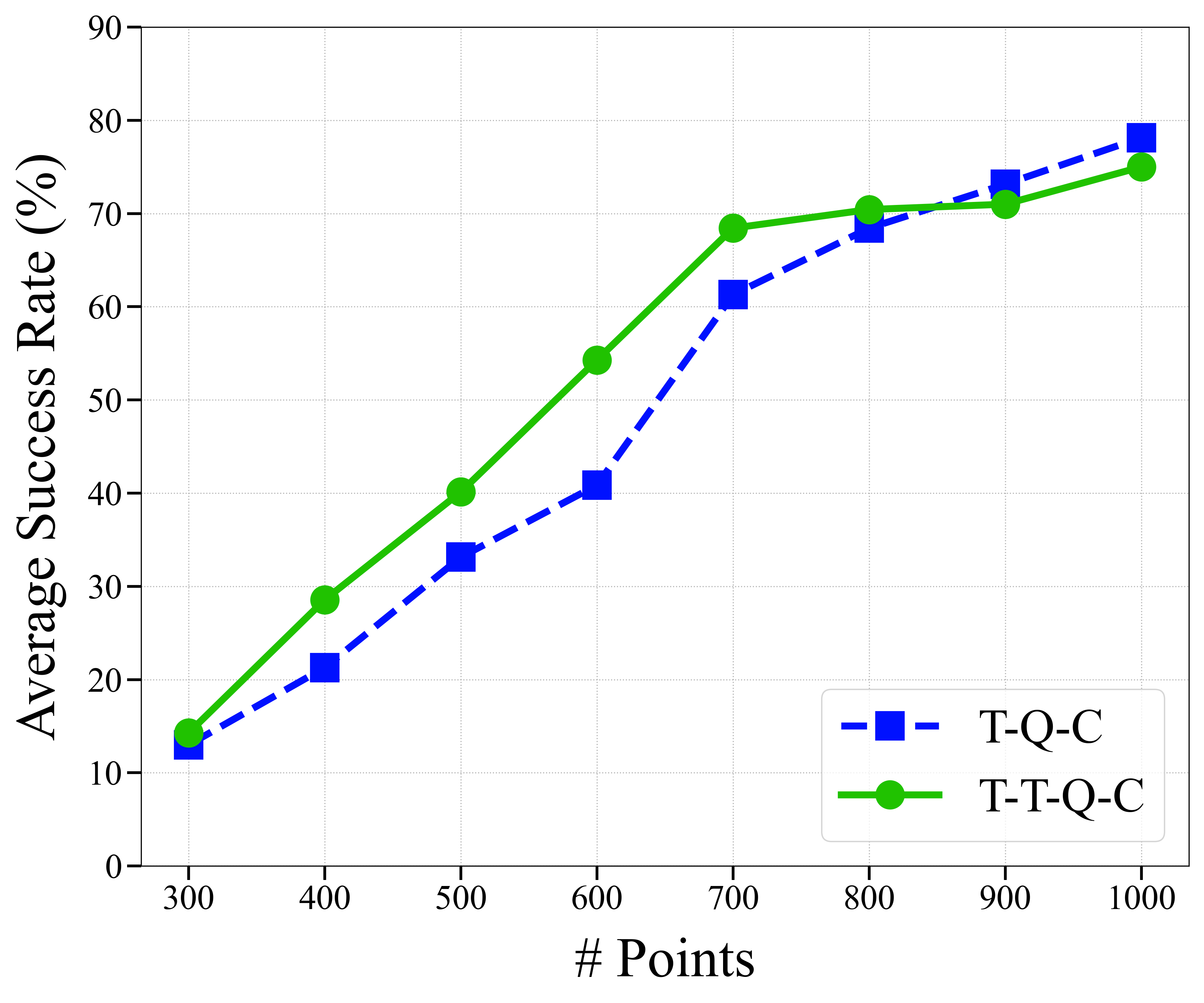}
        \caption{(b) $\mathcal{N}(0,0.01)$}
    \end{subfigure}
    \hfill
    \begin{subfigure}{0.24\linewidth}
        \centering
        \includegraphics[width=\linewidth]{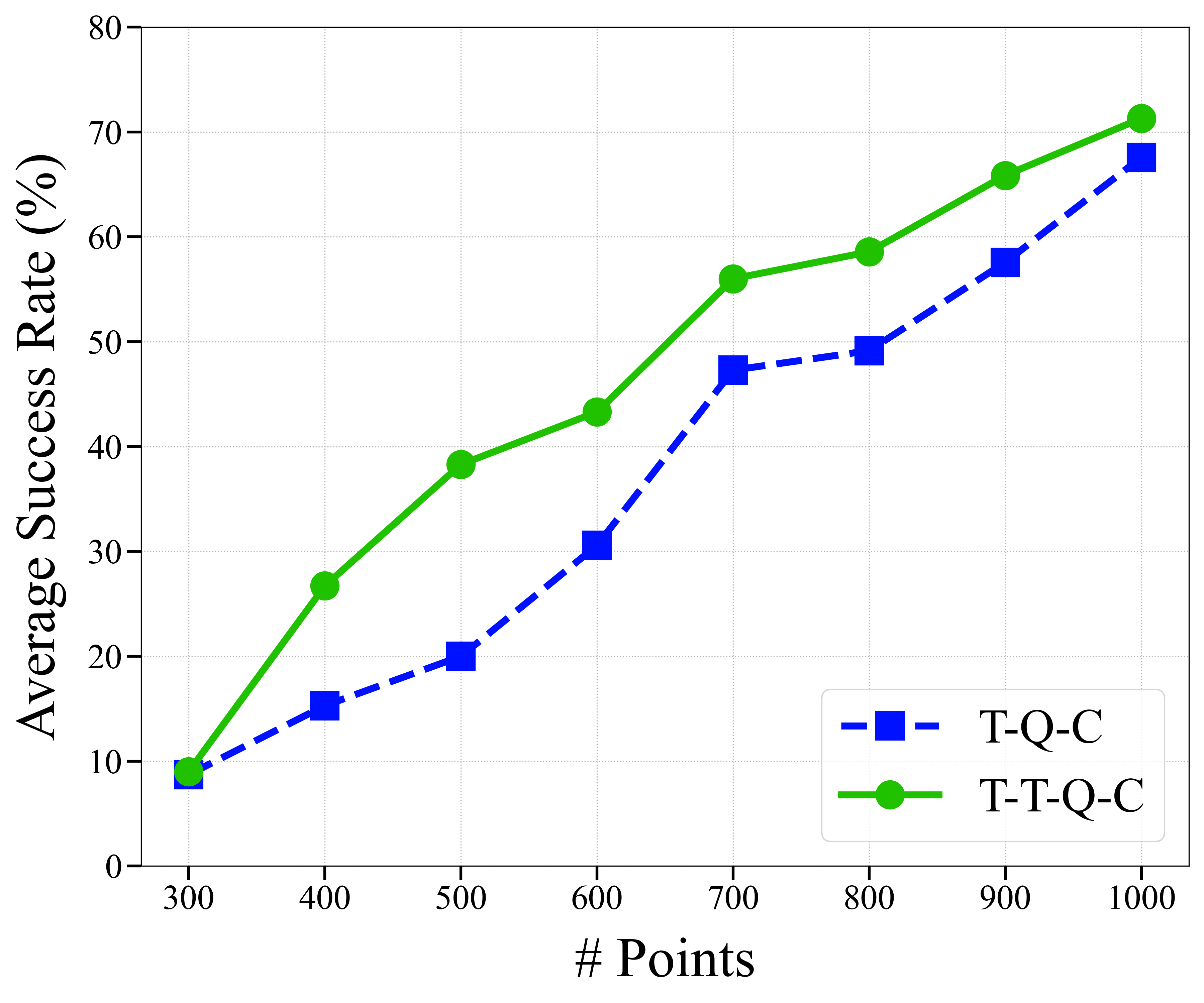}
        \caption{(c) $\mathcal{N}(0,0.02)$}
    \end{subfigure}
    \hfill
    \begin{subfigure}{0.24\linewidth}
        \centering
        \includegraphics[width=\linewidth]{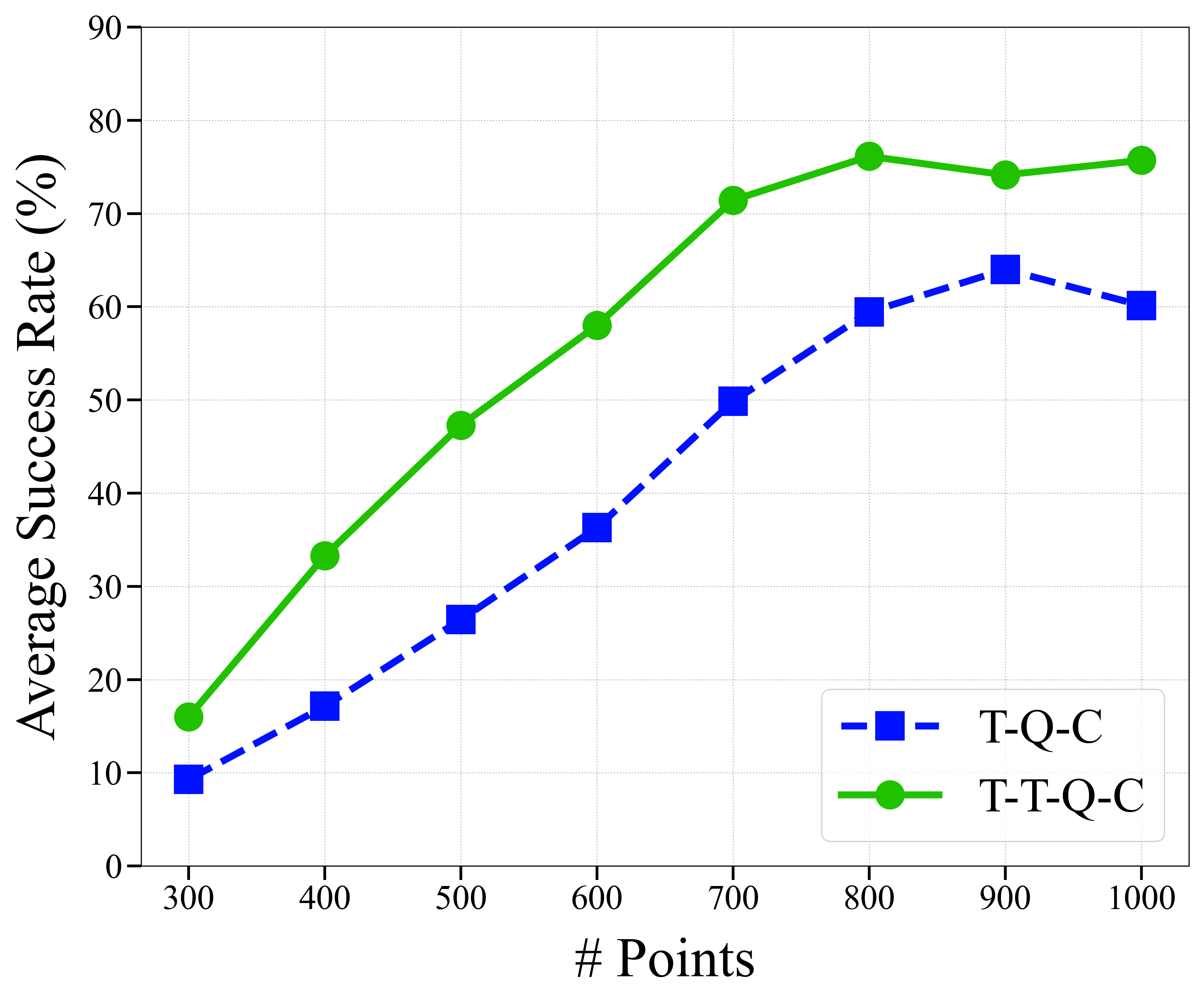}
        \caption{(d) $\mathcal{N}(0,0.03)$}
    \end{subfigure}

    \caption{(a) corresponds to the model trained using the exact ground-truth transformation, whereas (b), (c), and (d) show the results obtained when Gaussian noise, as indicated in each caption, is injected into the translation component of the transformation during training.}
    \label{fig:shooooot}
\end{figure*}

The experimental results show that when the ground-truth transformation is available, the performance gap between the two training strategies is negligible across different point cloud resolutions. In contrast, when an inaccurate transformation is used, the difference becomes substantial, and jointly leveraging both likelihood (i.e., MLE of T-T-Q-C) terms consistently yields significantly better performance. These findings confirm that our probabilistic modeling is both reasonable and well-justified. In particular, the proposed formulation combining correspondence and pose likelihoods is crucial when the ground-truth transformation is unavailable.

Although this distinction may appear minor under idealized conditions, it becomes significant in practical robotic scenarios where accurate ground-truth transformations are often unavailable. Moreover, the proposed formulation enables training with quasi–ground-truth poses from LiDAR odometry, avoiding reliance on expensive sensing suites such as GNSS–INS.

To further evaluate the impact of this limitation under realistic conditions where only quasi-ground-truth poses are available, we conduct additional experiments on a scene-level point cloud dataset. We use models trained for 12 epochs on the HeLiPR~\citep{jung2024helipr} VLP-16 dataset with a voxel size of 0.5,m, following the training protocol described in Section~\ref{app:lo_training_setup}. We compare the point cloud geometry predicted by each training strategy. As illustrated in Fig.~\ref{fig:augaug}, clear qualitative differences can be observed. When optimizing only the correspondence likelihood, the estimated geometry tends to be conservative, particularly on planar structures such as walls, where the reconstructed surfaces appear under-expressed and incomplete. In contrast, when the pose likelihood is jointly incorporated during training, the model captures structural elements such as walls and floors more aggressively and produces geometrically more faithful representations.

Beyond these qualitative observations, we further evaluate how the differences in learned geometry affect downstream tasks. Specifically, we assess both models on a LiDAR odometry task to quantify the practical impact of the learned geometric representations. During evaluation, the covariances predicted by the model are incorporated into Generalized-ICP (GICP)~\citep{segal2009generalized}, and performance is measured on the HeLiPR~\citep{jung2024helipr} \textit{KAIST} sequence using relative pose error (RPE) across different temporal intervals $\Delta$. The results in Table~\ref{tab:foo__} show that the proposed T-T-Q-C formulation consistently outperforms T-Q-C, emphasizing the importance of probabilistic modeling that accounts for uncertainty in transformation labels.

\begin{figure*}[t]
    \centering
    \captionsetup[subfigure]{labelformat=empty}
    \begin{minipage}{0.98\linewidth}
        \centering
        \begin{subfigure}{0.4\linewidth}
            \centering
            \includegraphics[width=\linewidth,keepaspectratio,trim=850 400 750 350,clip]{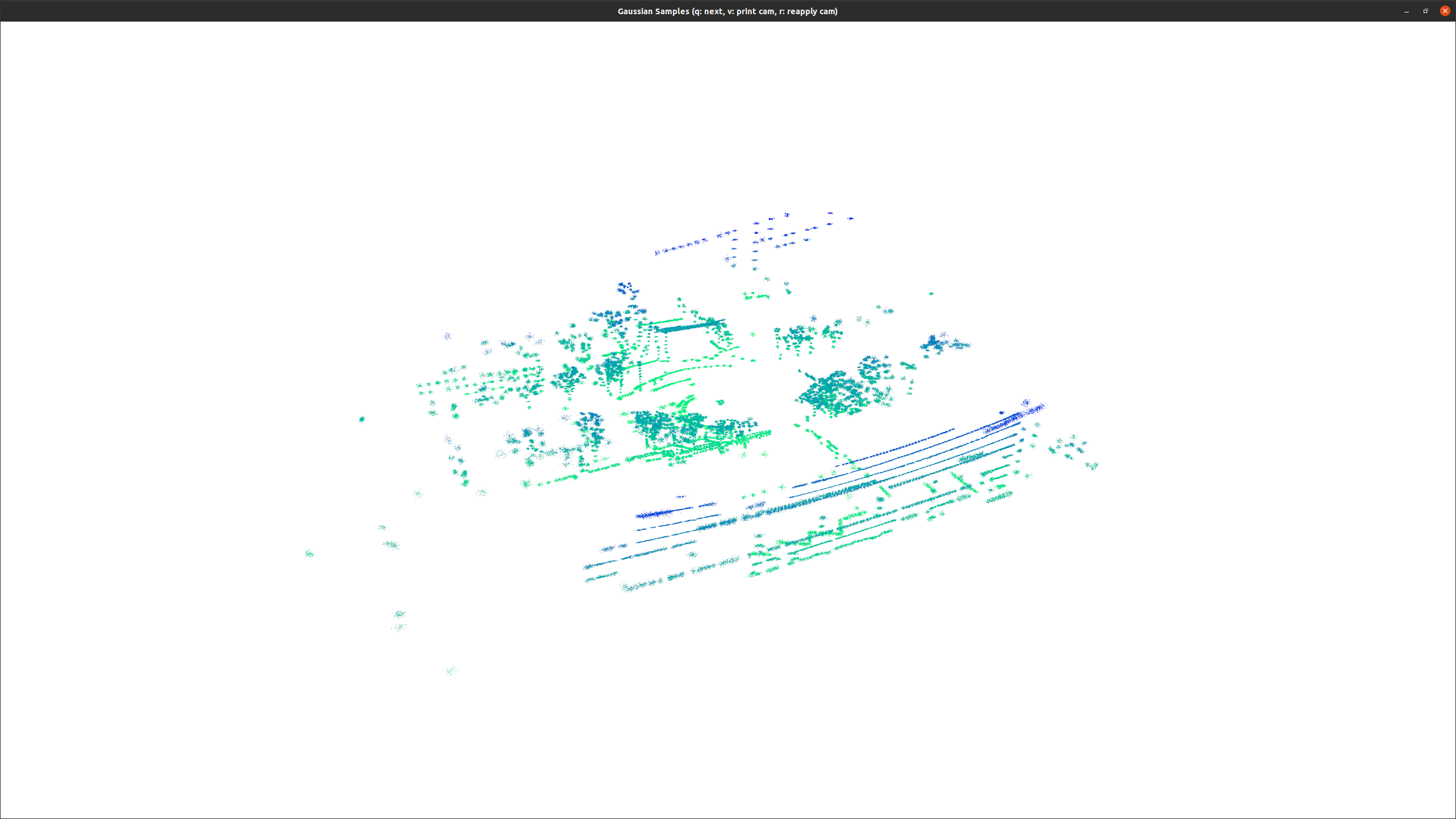}
            \caption{(a) Assuming perfect pose label (T-Q-C)}
            \label{fig:5}
        \end{subfigure}
        \hspace{10mm}
        \begin{subfigure}{0.4\linewidth}
            \centering
            \includegraphics[width=\linewidth,keepaspectratio,trim=850 400 750 350,clip]{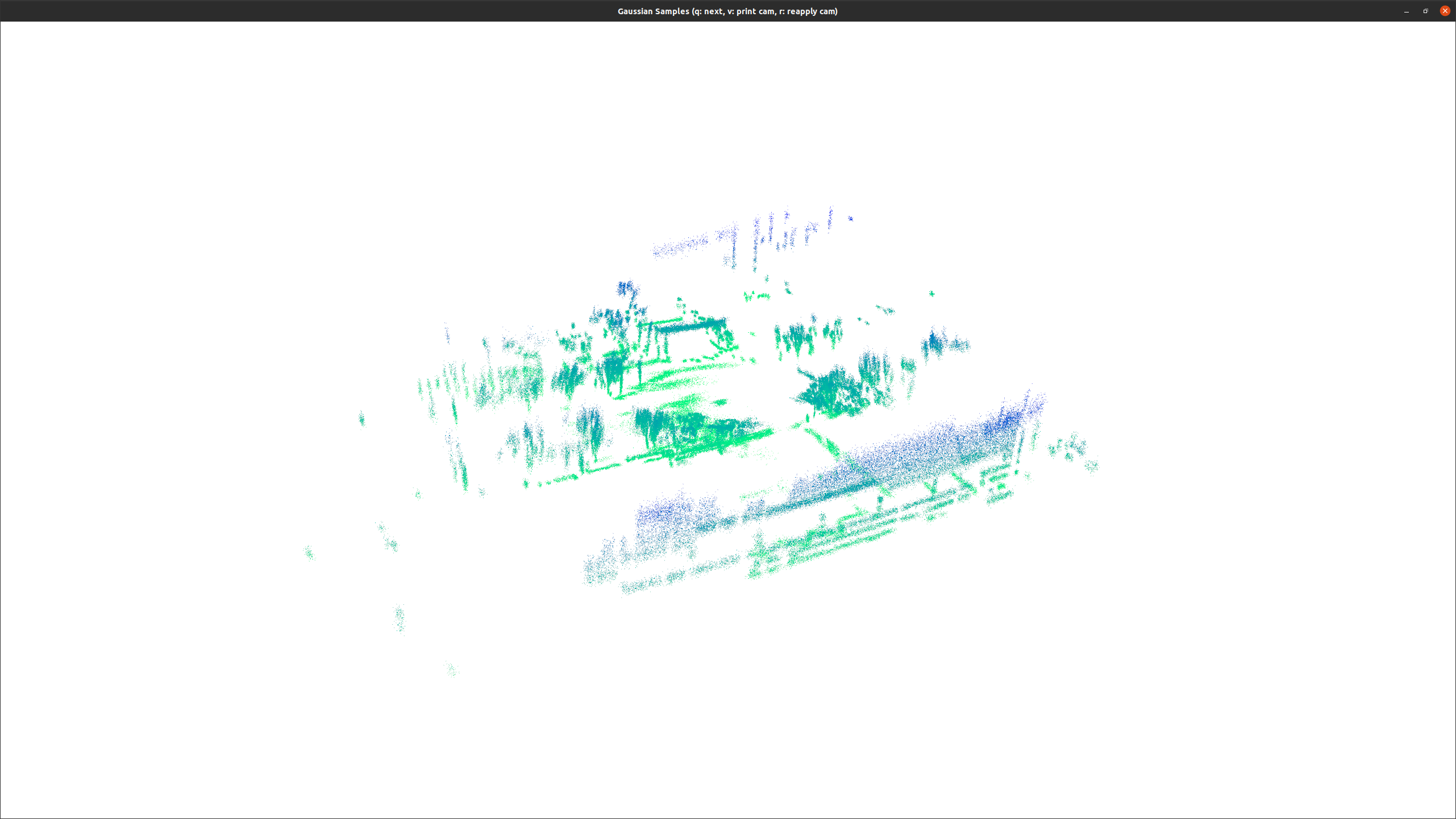}
            \caption{(b) Assuming imperfect pose label (T-T-Q-C)}
            \label{fig:30}
        \end{subfigure}
    \end{minipage}

    \caption{(a) shows the result obtained by training under a probabilistic model that assumes an exact (noise-free) transformation, whereas (b) shows the result obtained by training under a probabilistic model that does not assume an exact transformation.}
    \label{fig:augaug}
\end{figure*}

\begin{table}[htbp]
\centering
\caption{LiDAR odometry evaluation on HeLiPR \textit{KAIST04} using RPE (m) (RMSE at different $\Delta$).}
\label{tab:foo__}
\begin{tabular}{lc|cc|cc|cc}
\hline
\textbf{Method} 
& \textbf{GICP~\citep{segal2009generalized}} 
& \textbf{+ T-Q-C} 
& \textbf{+ T-T-Q-C} 
& \textbf{+ T-Q-C} 
& \textbf{+ T-T-Q-C} 
& \textbf{+ T-Q-C} 
& \textbf{+ T-T-Q-C} \\

& 
& \multicolumn{2}{c|}{($\Delta = 1$)} 
& \multicolumn{2}{c|}{($\Delta = 50$)} 
& \multicolumn{2}{c}{($\Delta = 100$)} \\
\hline
RMSE
& -- 
& 0.0385 
& \textbf{0.0186} 
& 1.8846 
& \textbf{0.6747} 
& 5.4912 
& \textbf{2.0053} \\
\hline
\end{tabular}
\end{table}

\clearpage
\newpage
\section{Advantages of Ellipsoidal Representation}
\vspace{+3mm}
\label{app:gaussian_advantage}

\subsection{Advantages of Gaussian Ellipsoidal Representation}
Gaussian ellipsoids provide both \emph{efficiency} and \emph{controllability} as a local geometric representation. First, they are computationally efficient: the network only needs to predict a compact covariance parameterization (e.g., six degrees of freedom for a symmetric $3\times 3$ covariance), so the output space is not overly complex. Empirically, we observe that training typically converges within a few hours in most settings. Second, they offer controllability without distorting the raw sensor geometry: the network does not move the raw measurements, but instead learns how much to trust each point locally by predicting its covariance. At inference time, the user can control the effective strictness of the representation by choosing a confidence scale (e.g., using a $1\sigma$ or $0.1\sigma$). 

\subsection{Scene Completion vs. Local Geometric Abstraction}
While generative models often aim to \emph{complete} an entire scene from partial scans, full-scene completion is not a well-posed objective.  The unobserved geometry can be fundamentally ambiguous, and any hallucinated completion may introduce structures that do not exist in the real world, which can be detrimental in downstream robotic perception and decision-making. In contrast, our goal is not to reconstruct every unobserved detail, but to predict the \emph{local geometric structure} around the observed measurements in an abstract and physically grounded form.  Rather than requiring globally consistent scene synthesis, the model only needs to capture how local neighborhoods behave---e.g., the directionality, anisotropy, and uncertainty of local surface geometry around each measurement. In many geometric perception pipelines, fine-grained global completeness is unnecessary; what matters is reliable local geometry.  Therefore, Gaussian ellipsoidal representations provide a pragmatic alternative to scene-completion generative modeling, offering a controllable abstraction aligned with robotic geometric perception.

\subsection{Towards Sensor-Agnostic Geometric Perception}
One of the fundamental problems geometric perception must address is reliable operation under real-world diversity.
In practical deployment, the sensing process is rarely stationary: sensor hardware and configurations vary widely, measurement density and angular resolution change, fields-of-view differ, and the level of noise, occlusion, and partial observability fluctuates across environments. As a consequence, a method that performs well under a specific sensing regime often fails to maintain the same level of stability and accuracy once the sensing conditions deviate from those assumed during its design or calibration. A central driver of this gap is the quality of geometric evidence available from observations. When measurements are sparse---for instance, in low-channel LiDAR settings---local geometry becomes under-sampled, and the information required for reliable association and constraint formation becomes fundamentally weaker. This scarcity amplifies ambiguity in correspondence, reduces constraint informativeness, and increases sensitivity to outliers and partial visibility. In contrast, when observations are sufficiently dense to richly capture the underlying surfaces, geometric perception problems often enter a markedly easier regime: local neighborhoods are clearly defined, the induced constraints are better conditioned, and optimization tends to be less fragile. These considerations motivate representations that reduce the discrepancy between sensing regimes at the geometric abstraction level. Rather than forcing downstream algorithms to compensate for sparsity through ad-hoc retuning, a more principled approach is to provide a local representation that captures the structure of geometry around each observation. Gaussian ellipsoidal representations offer such an interface: they do not hallucinate unobserved surfaces, yet they enrich sparse measurements with an uncertainty-aware description of local shape and anisotropy. By transforming raw measurements into a physically grounded and controllable local abstraction, they can improve performance across diverse sensing conditions.

\setcounter{figure}{0}          
\renewcommand{\thefigure}{S\arabic{figure}}

\clearpage
\newpage
\label{app:ellipsoidal_structure}
\begin{figure*}[t]
    \centering
    % ---------- Two Columns ----------
    \begin{minipage}[t]{0.48\textwidth}
        \centering
        \includegraphics[width=\linewidth, trim = 50 0 50 0, clip]{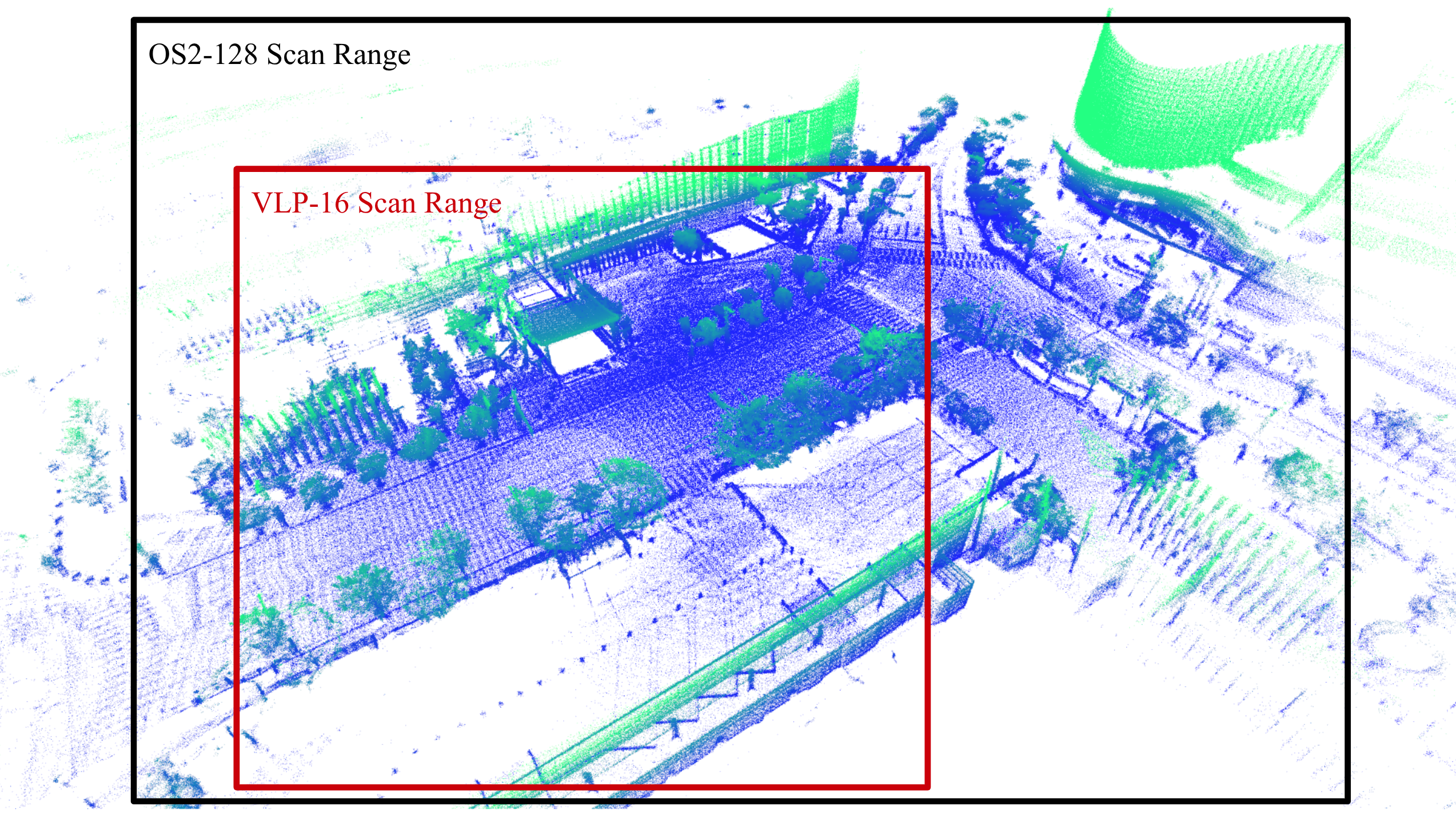}\\
        Ground-truth geometry

        \vspace{0.4em}
        \includegraphics[width=\linewidth]{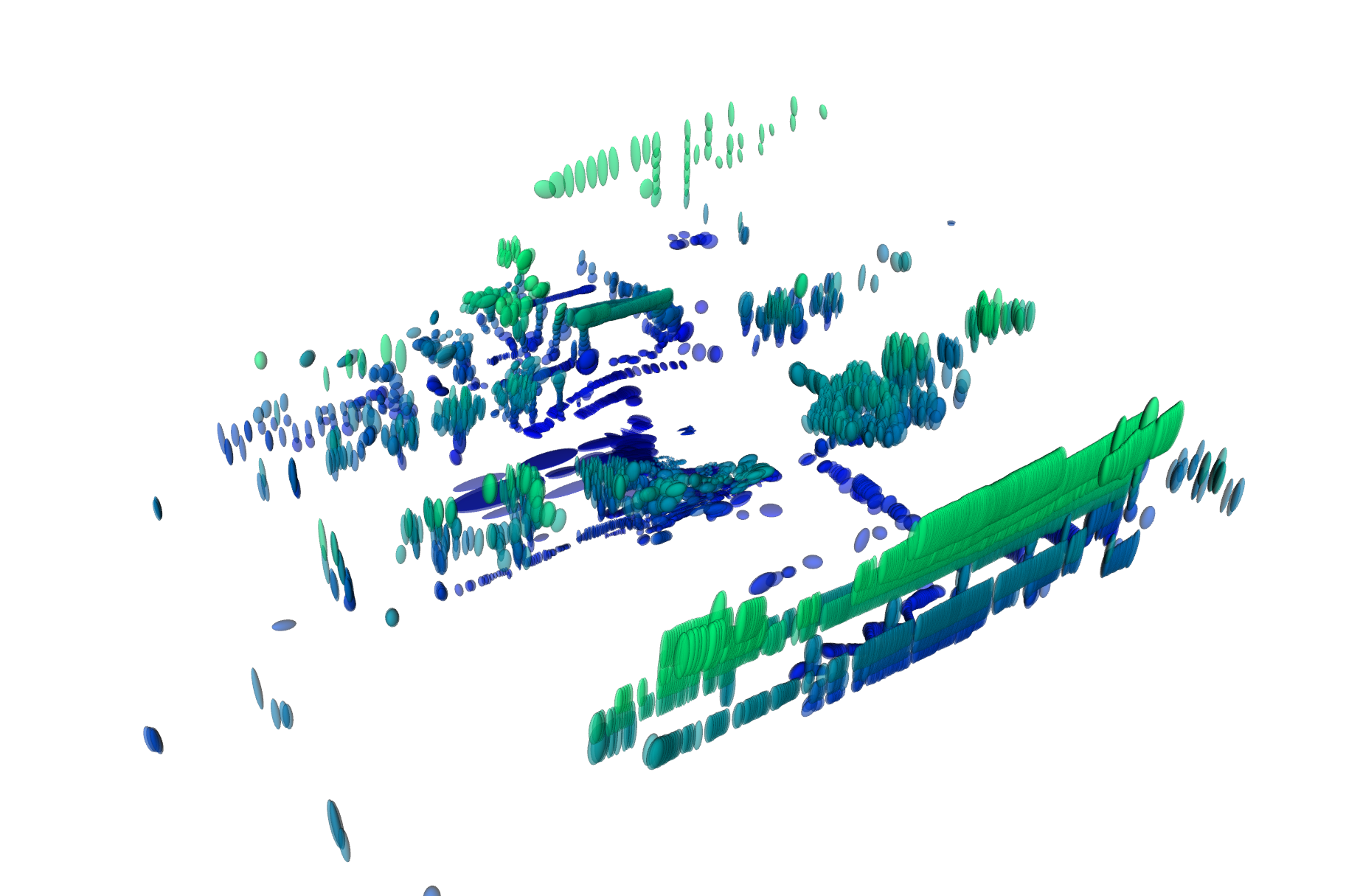}\\
        POLI-estimated geometry (VLP-16, voxel size 0.2\,m) 

        \vspace{0.4em}
        \includegraphics[width=\linewidth]{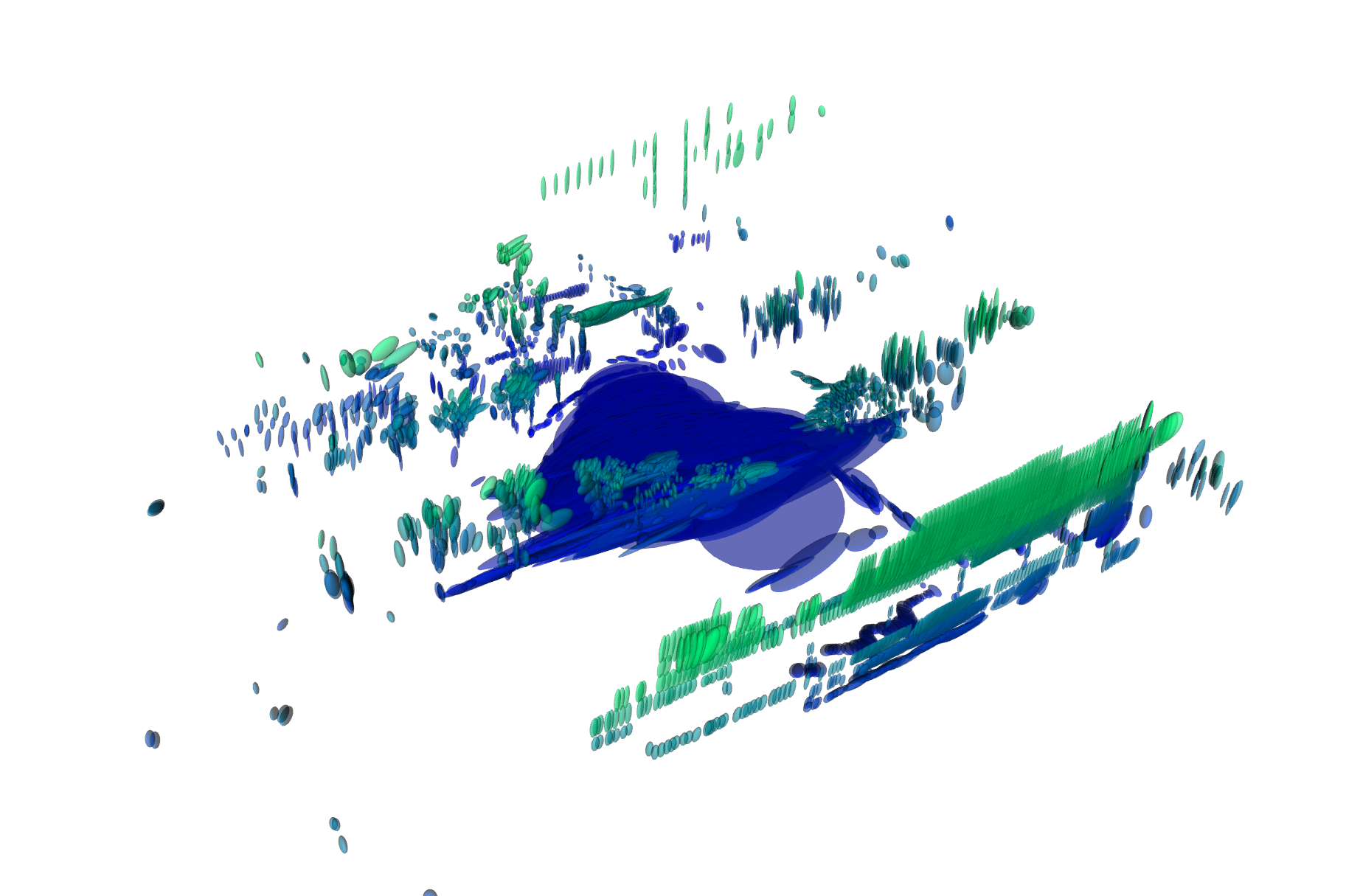}\\
        POLI-estimated geometry (VLP-16, voxel size 0.5\,m)
    \end{minipage}
    \hfill
    \begin{minipage}[t]{0.48\textwidth}
        \centering
        \includegraphics[width=\linewidth]{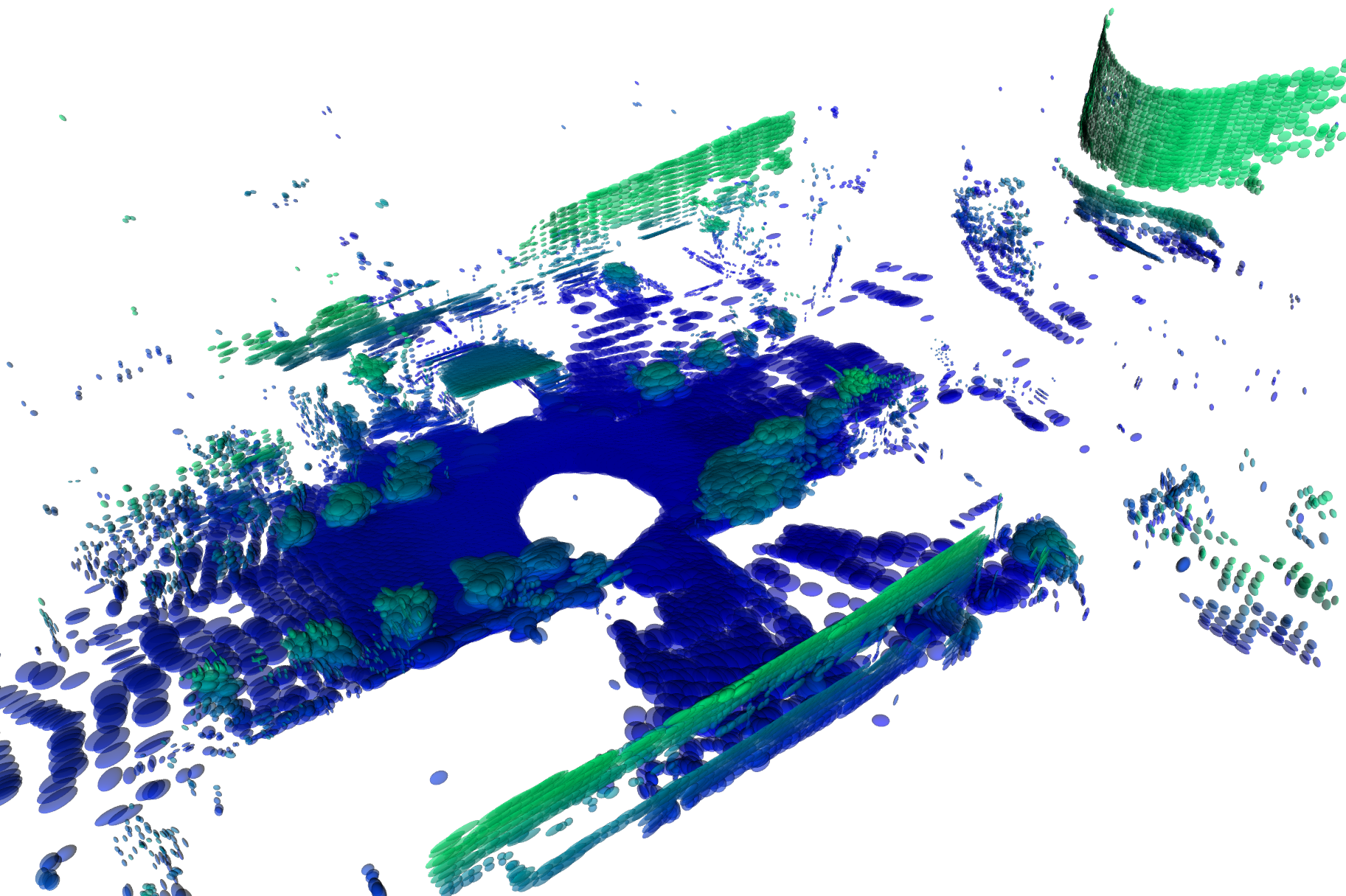}\\
        POLI-estimated geometry (OS2-128, voxel size 1.0\,m)

        \vspace{0.4em}
        \includegraphics[width=\linewidth]{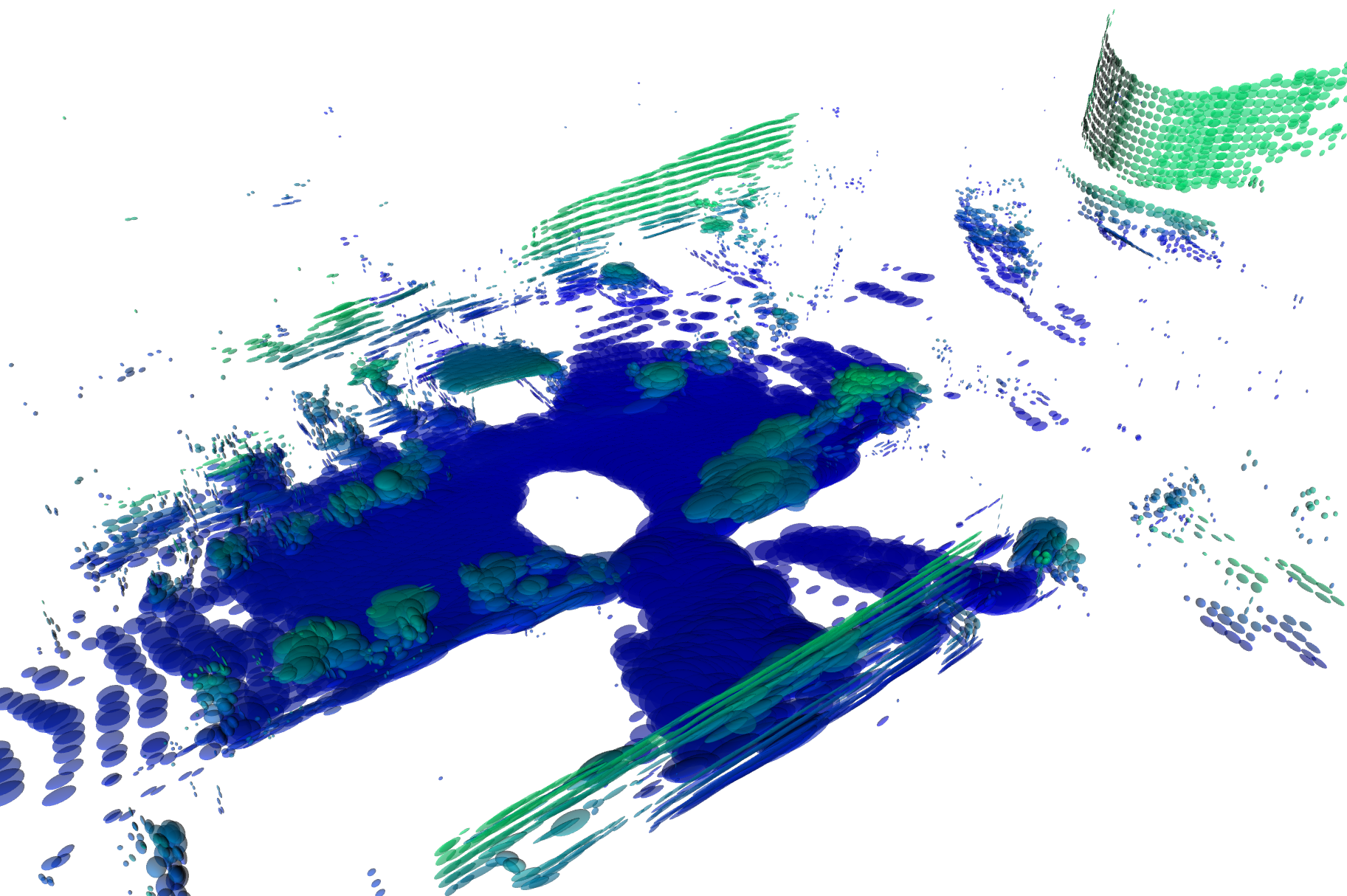}\\
        POLI-estimated geometry (OS2-128, voxel size 1.5\,m)

        \vspace{0.4em}
        \includegraphics[width=\linewidth, trim =  0 5 0 0, clip]{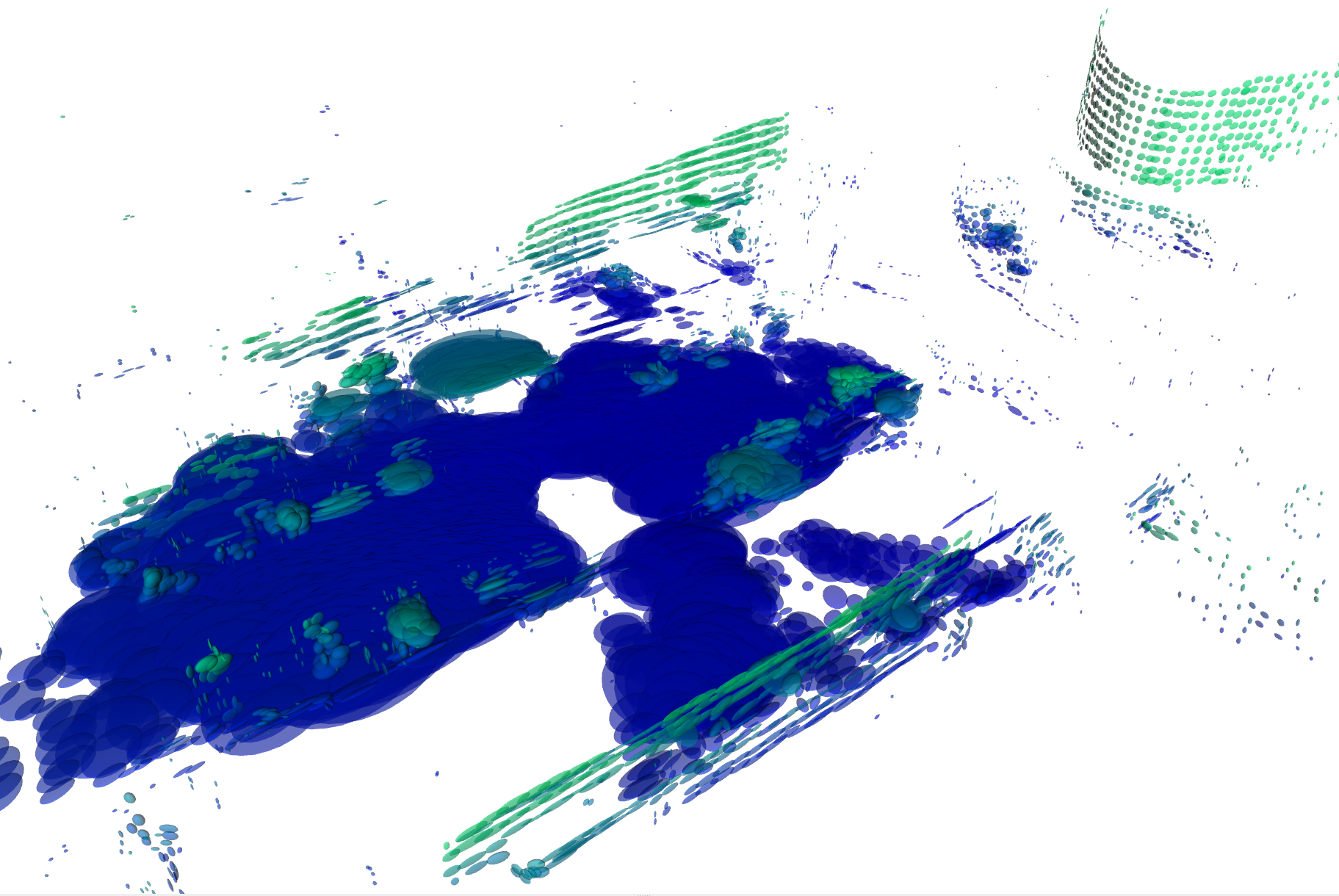}\\
        POLI-estimated geometry (OS2-128, voxel size 2.0\,m)
    \end{minipage}

    \caption{\textbf{Ellipsoidal geometric structure estimation using POLI.} Qualitative visualization of point cloud geometry estimated by POLI under varying sensor modalities and voxel resolutions. The point-wise covariance matrices produced by POLI are represented as ellipsoids, highlighting the underlying local geometric structures.}
    \label{fig:qualitative_comparison}
\end{figure*}

\clearpage
\newpage
\label{app:local}
\begin{figure*}[t]
\centering
\setlength{\tabcolsep}{2pt}

% ================= Block 1 : GLIM =================
\noindent
\hbox{
% ---- Rotated text ----
\begin{minipage}[c]{0.04\textwidth}
    \centering
    \rotatebox{90}{\small GLIM~\citep{koide2024glim}}
\end{minipage}
\hfill
% ---- Images ----
\begin{minipage}{0.94\textwidth}
    \centering
    % Row 1
    \begin{minipage}[t]{0.32\linewidth}
        \includegraphics[width=\linewidth]{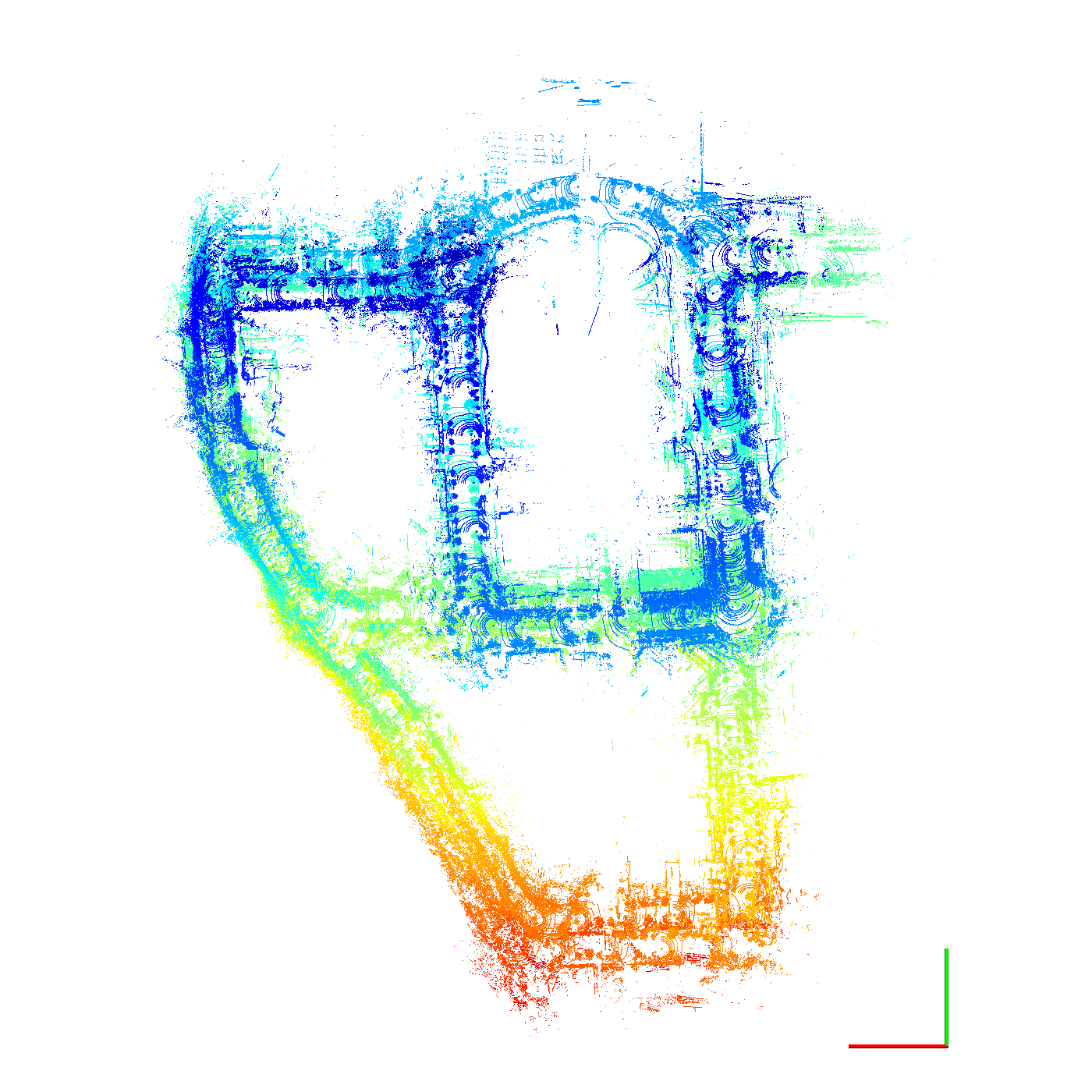}
    \end{minipage}\hfill
    \begin{minipage}[t]{0.32\linewidth}
        \includegraphics[width=\linewidth]{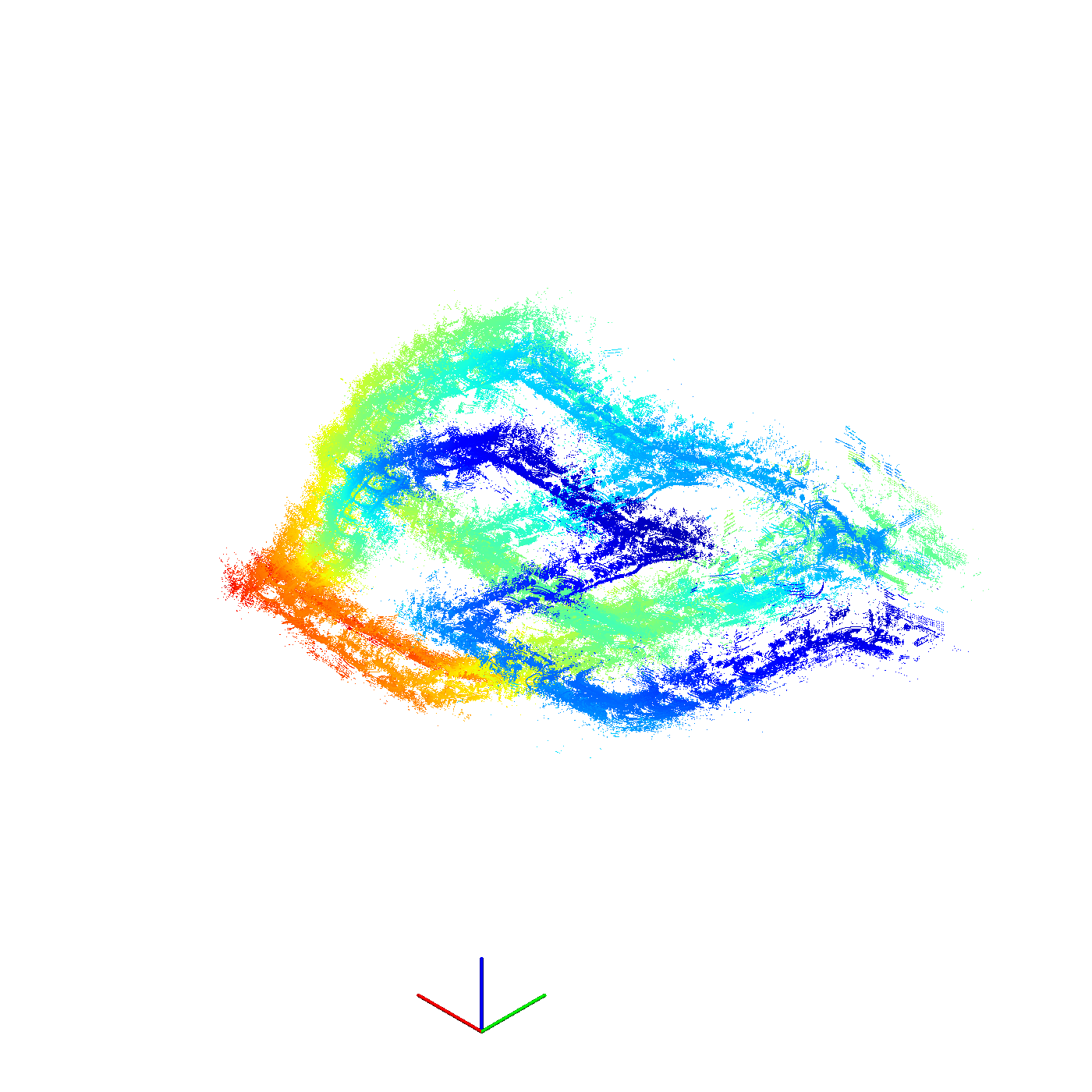}
    \end{minipage}\hfill
    \begin{minipage}[t]{0.32\linewidth}
        \includegraphics[width=\linewidth]{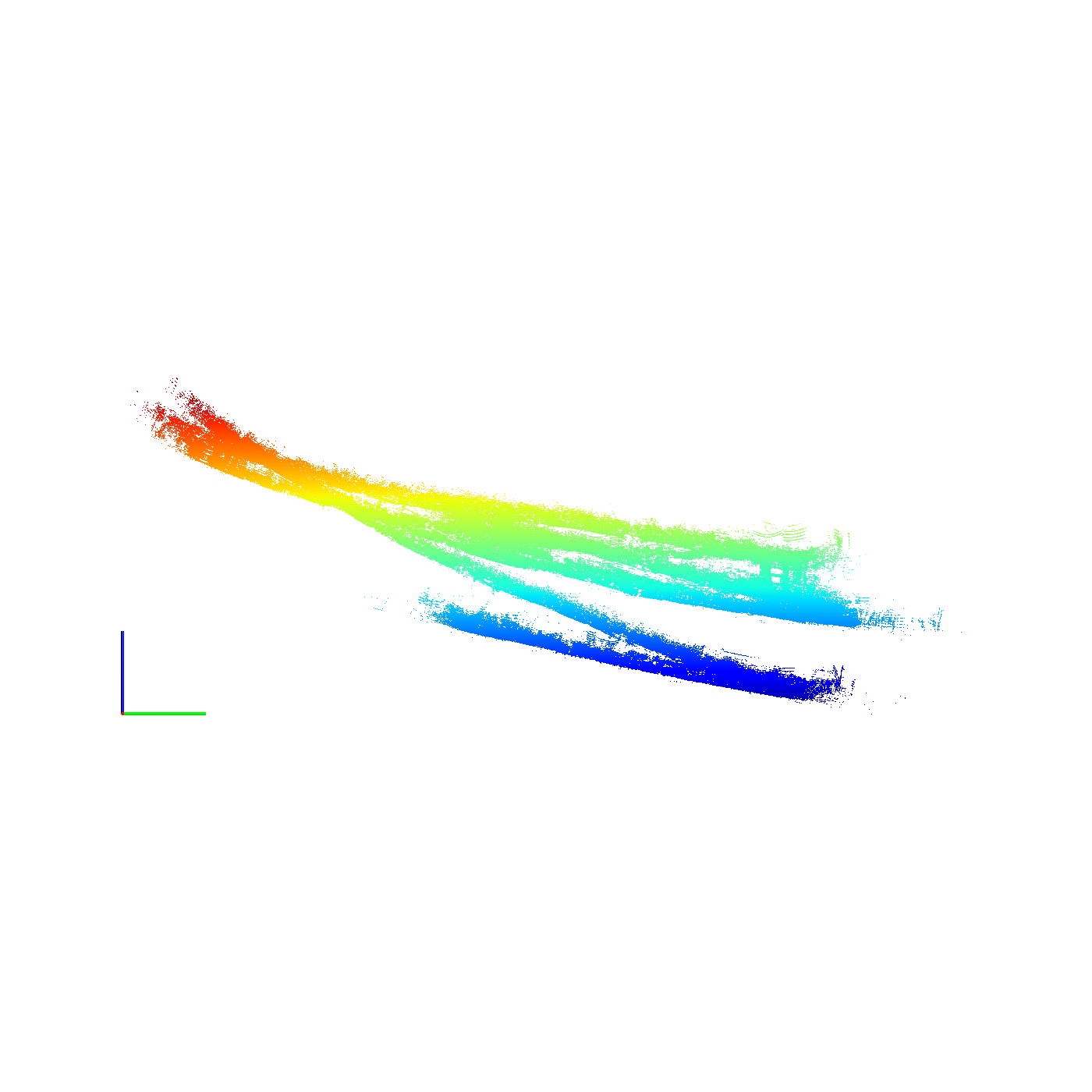}
    \end{minipage}

    % Row 2
    \begin{minipage}[t]{0.32\linewidth}
        \includegraphics[width=\linewidth]{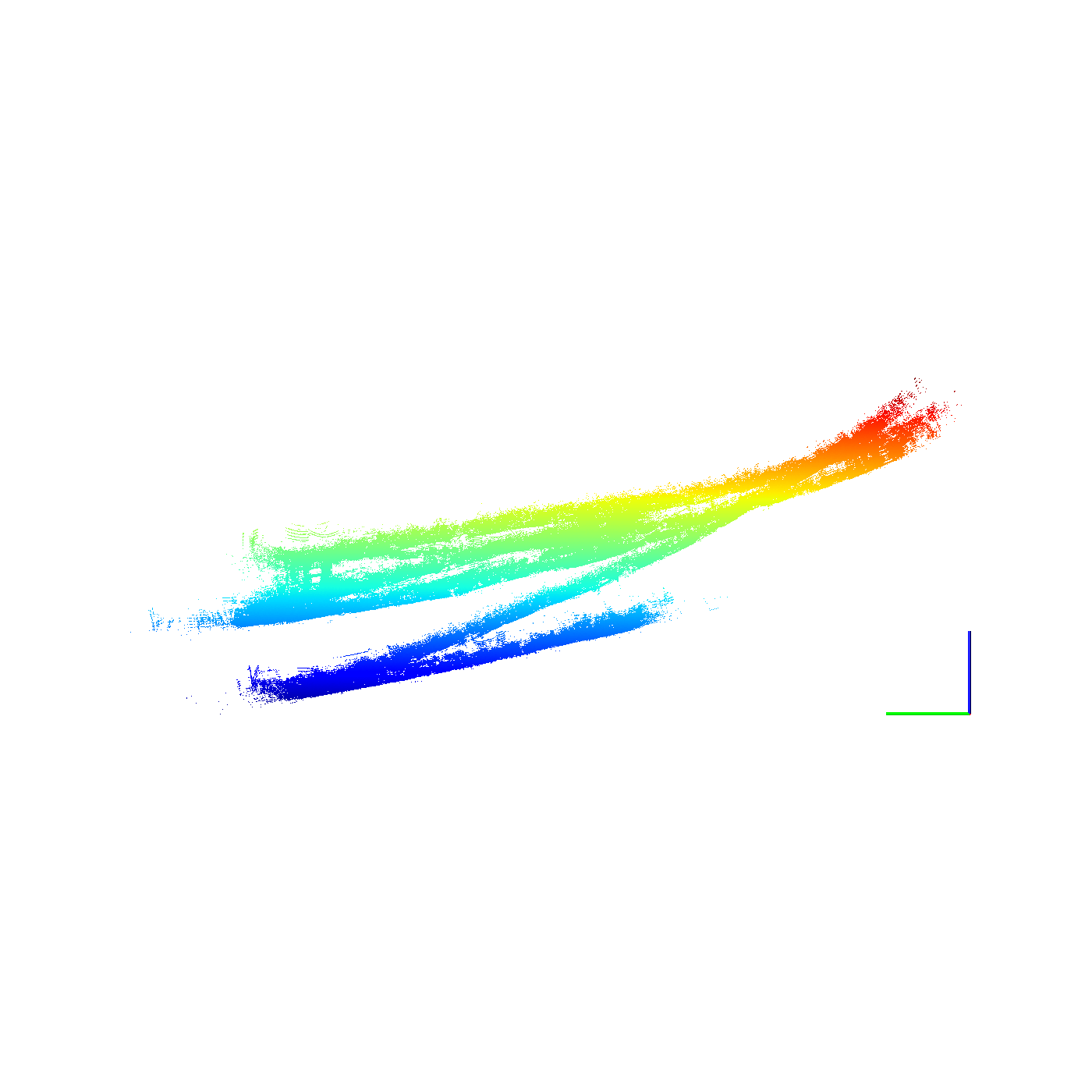}
    \end{minipage}\hfill
    \begin{minipage}[t]{0.32\linewidth}
        \includegraphics[width=\linewidth]{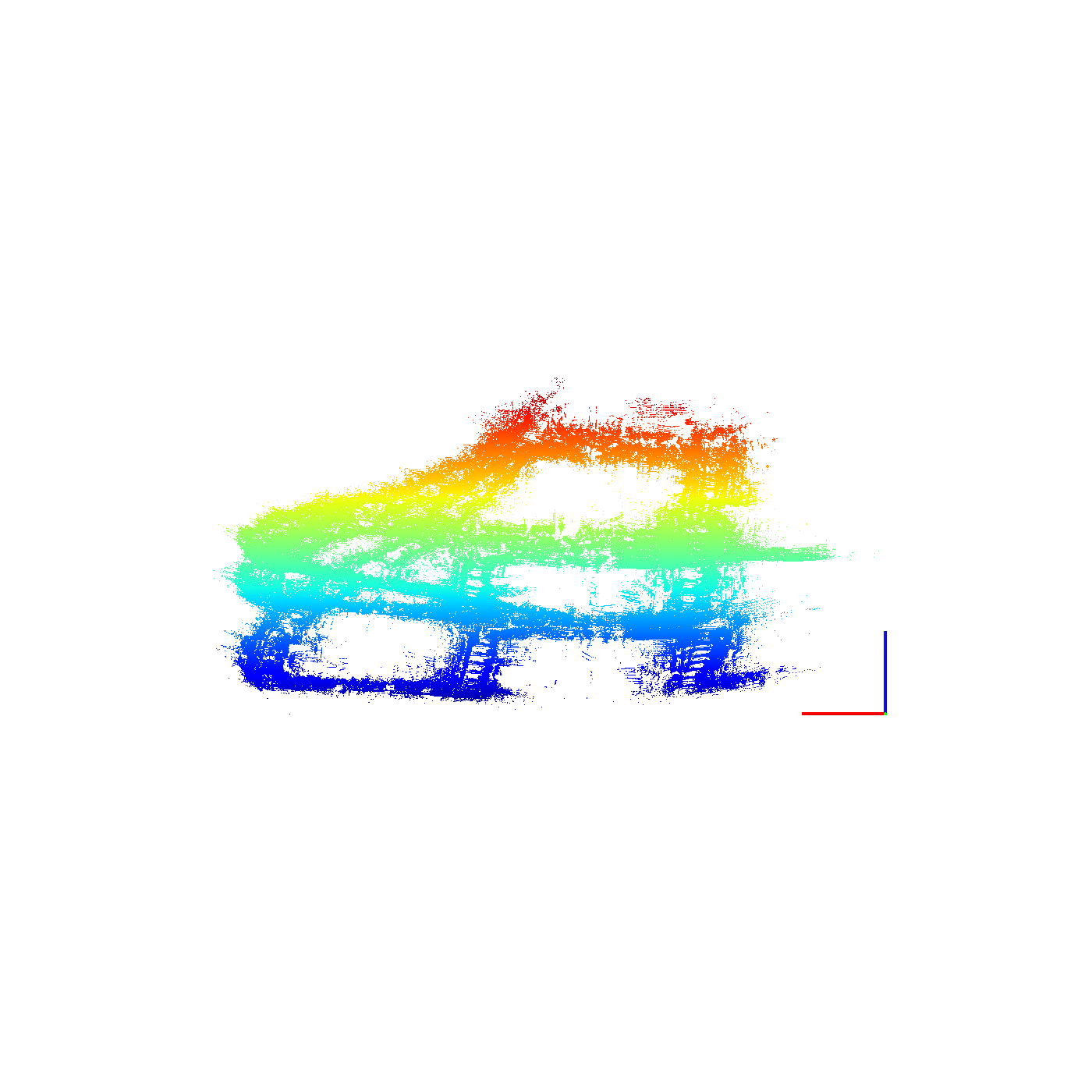}
    \end{minipage}\hfill
    \begin{minipage}[t]{0.32\linewidth}
        \includegraphics[width=\linewidth]{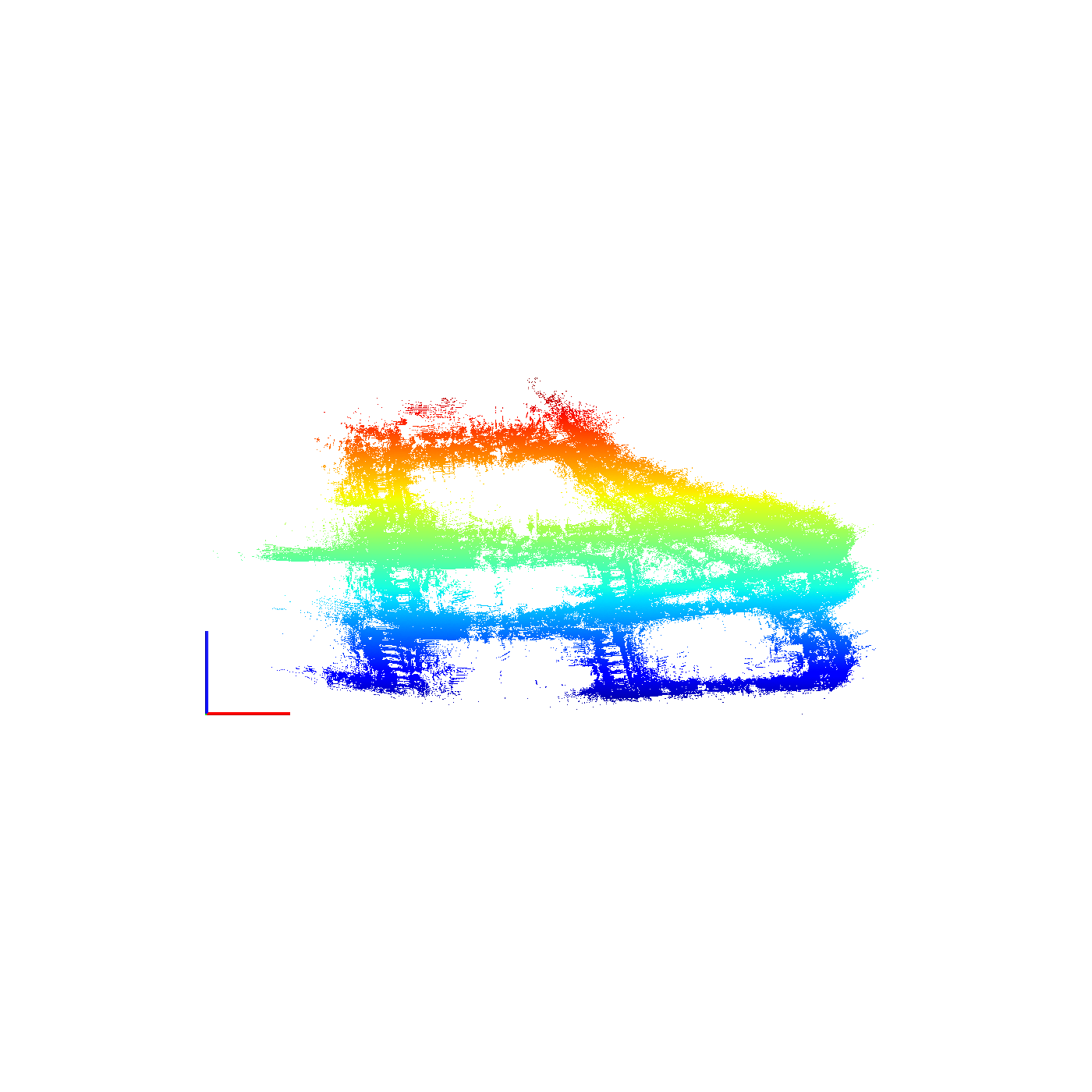}
    \end{minipage}
\end{minipage}
}

% ================= Block 2 : GLIM + POLY =================
\noindent
\hbox{
% ---- Rotated text ----
\begin{minipage}[c]{0.04\textwidth}
    \centering
    \rotatebox{90}{\small GLIM~\citep{koide2024glim} + POLI}
\end{minipage}
\hfill
% ---- Images ----
\begin{minipage}{0.94\textwidth}
    \centering
    % Row 3
    \begin{minipage}[t]{0.32\linewidth}
        \includegraphics[width=\linewidth]{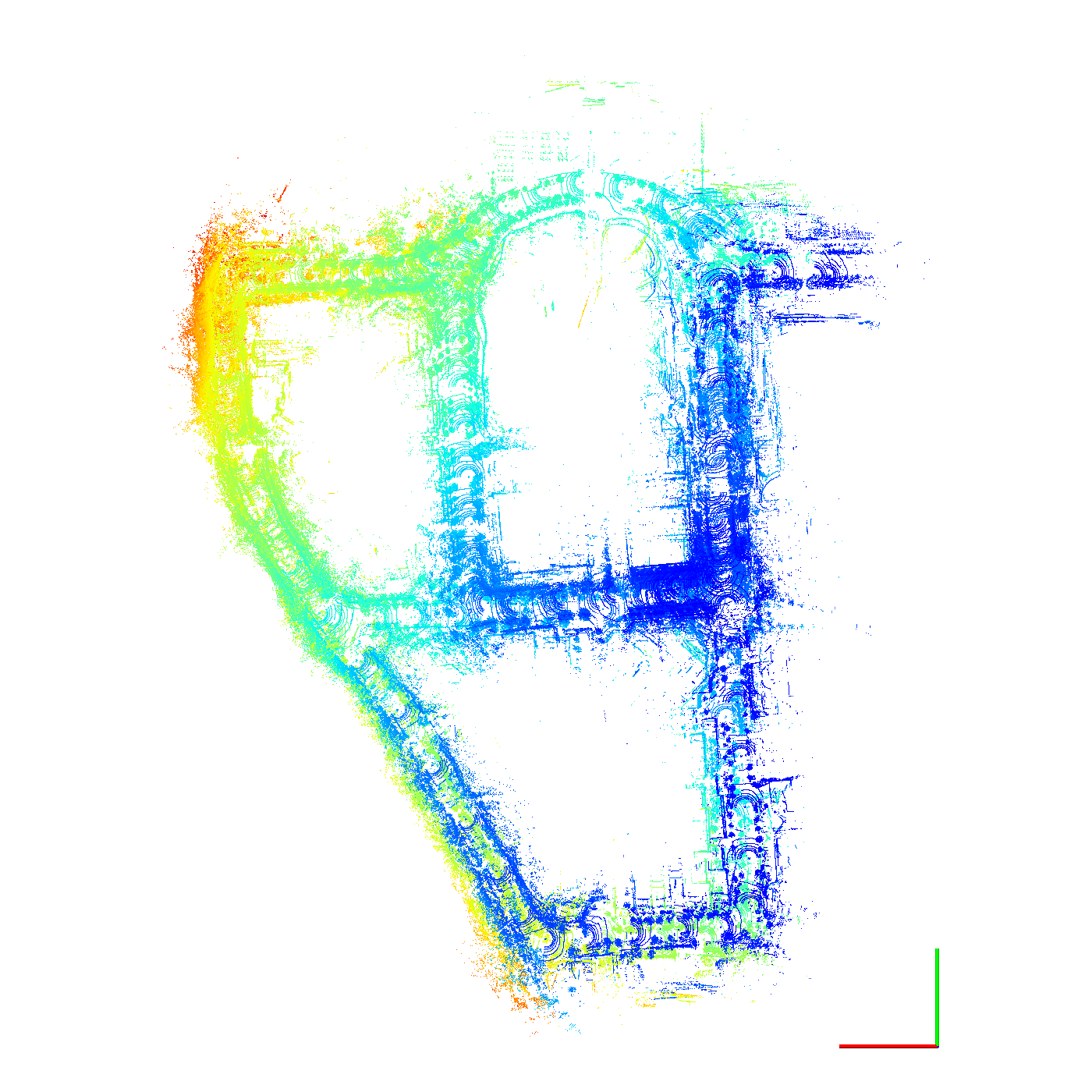}
    \end{minipage}\hfill
    \begin{minipage}[t]{0.32\linewidth}
        \includegraphics[width=\linewidth]{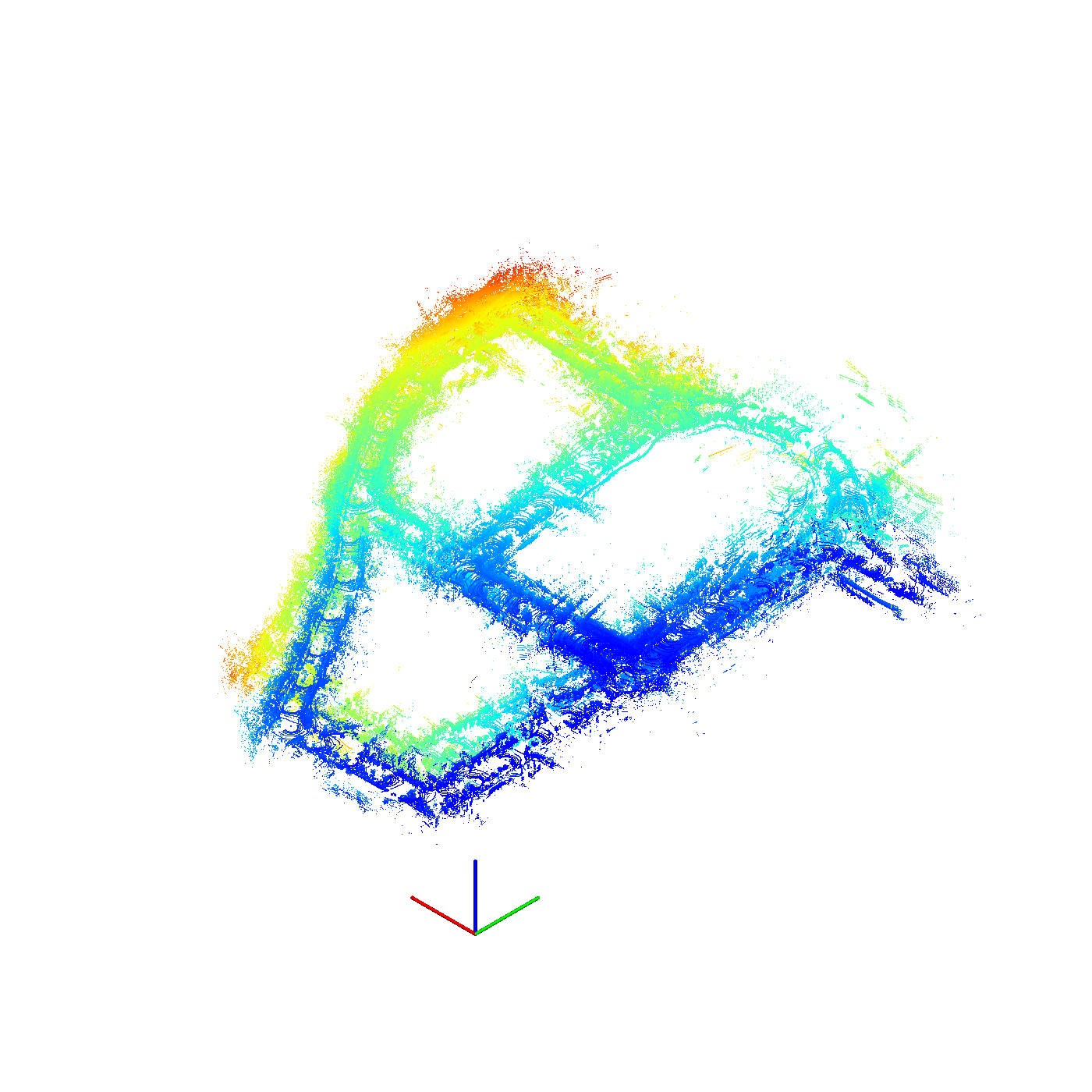}
    \end{minipage}\hfill
    \begin{minipage}[t]{0.32\linewidth}
        \includegraphics[width=\linewidth]{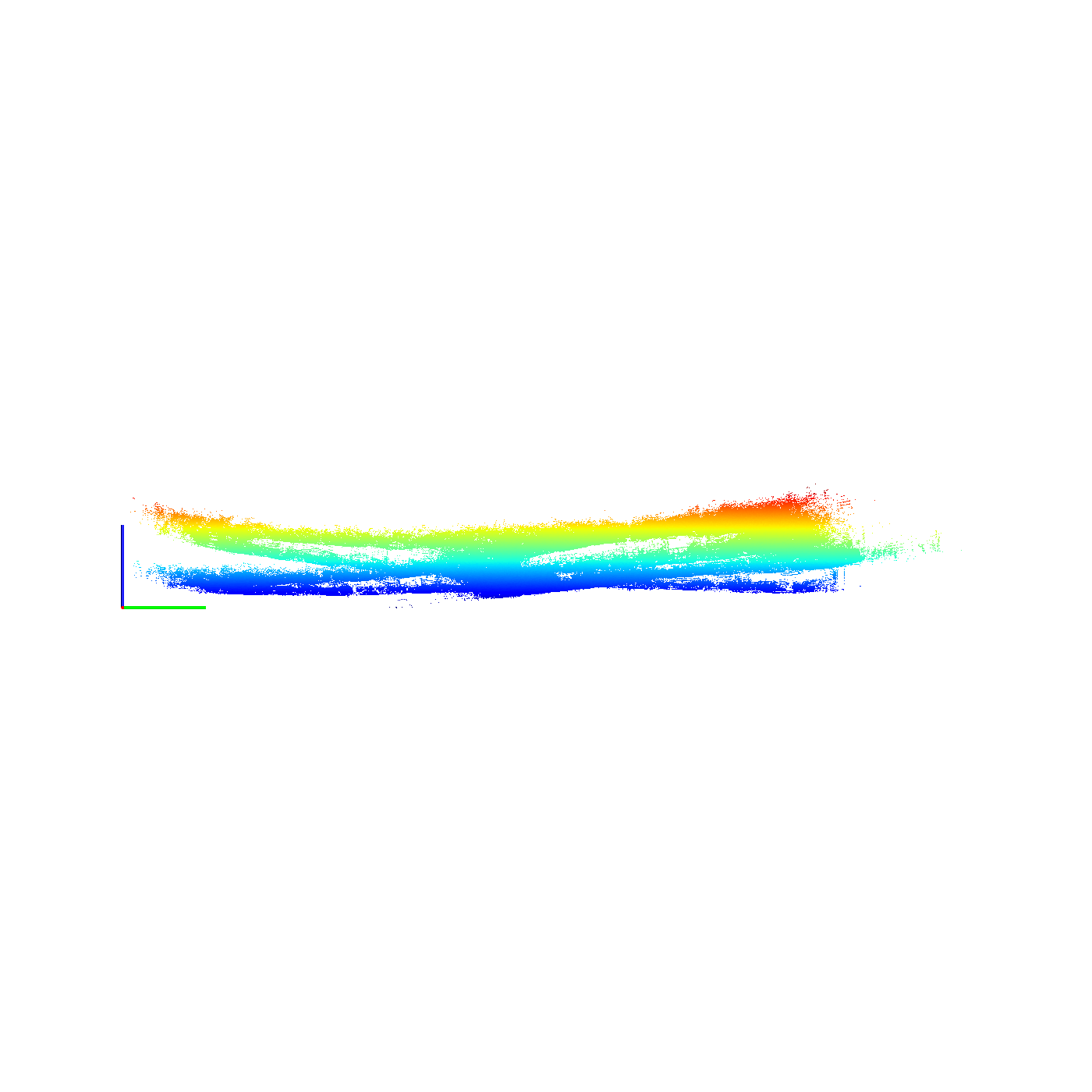}
    \end{minipage}

    % Row 4
    \begin{minipage}[t]{0.32\linewidth}
        \includegraphics[width=\linewidth]{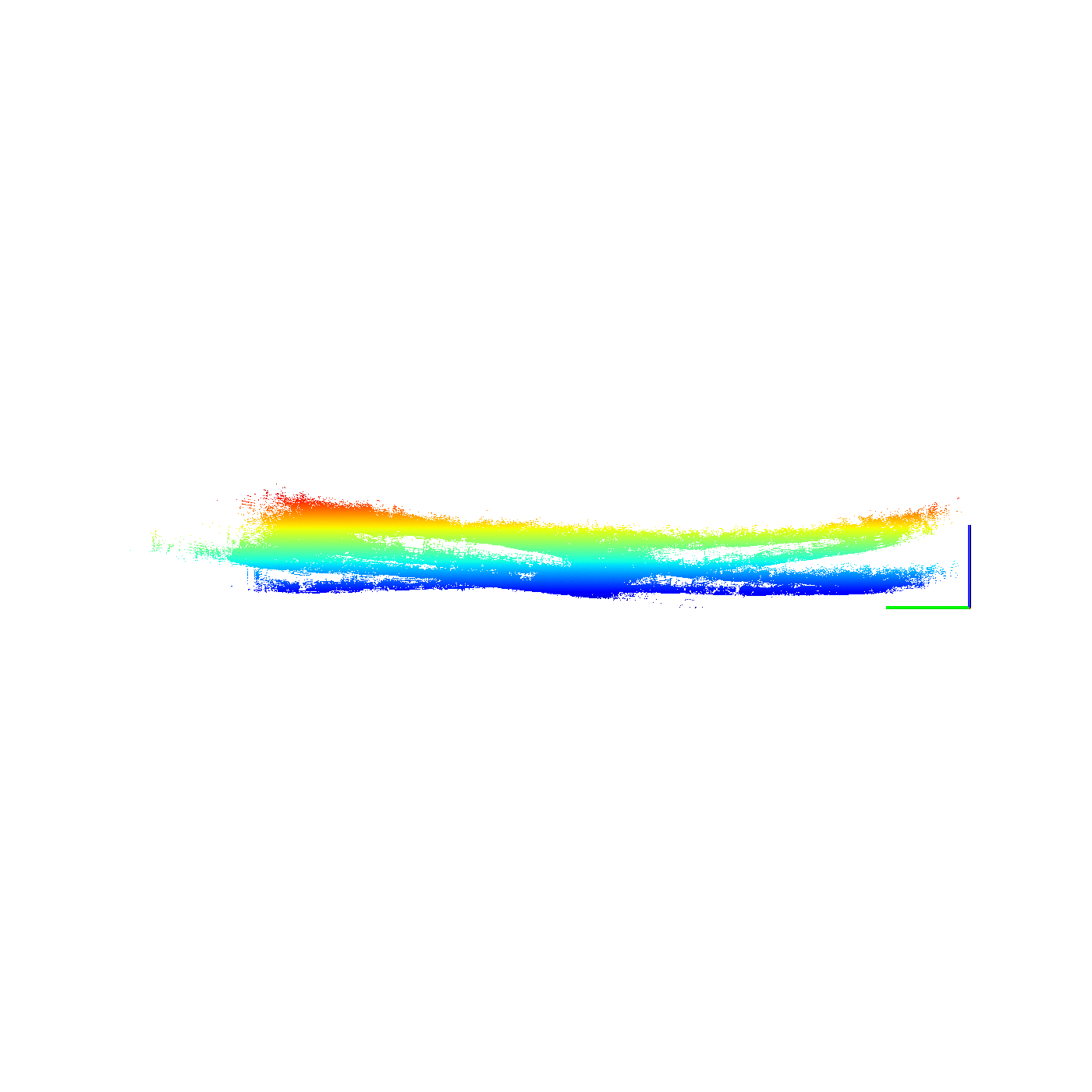}
    \end{minipage}\hfill
    \begin{minipage}[t]{0.32\linewidth}
        \includegraphics[width=\linewidth]{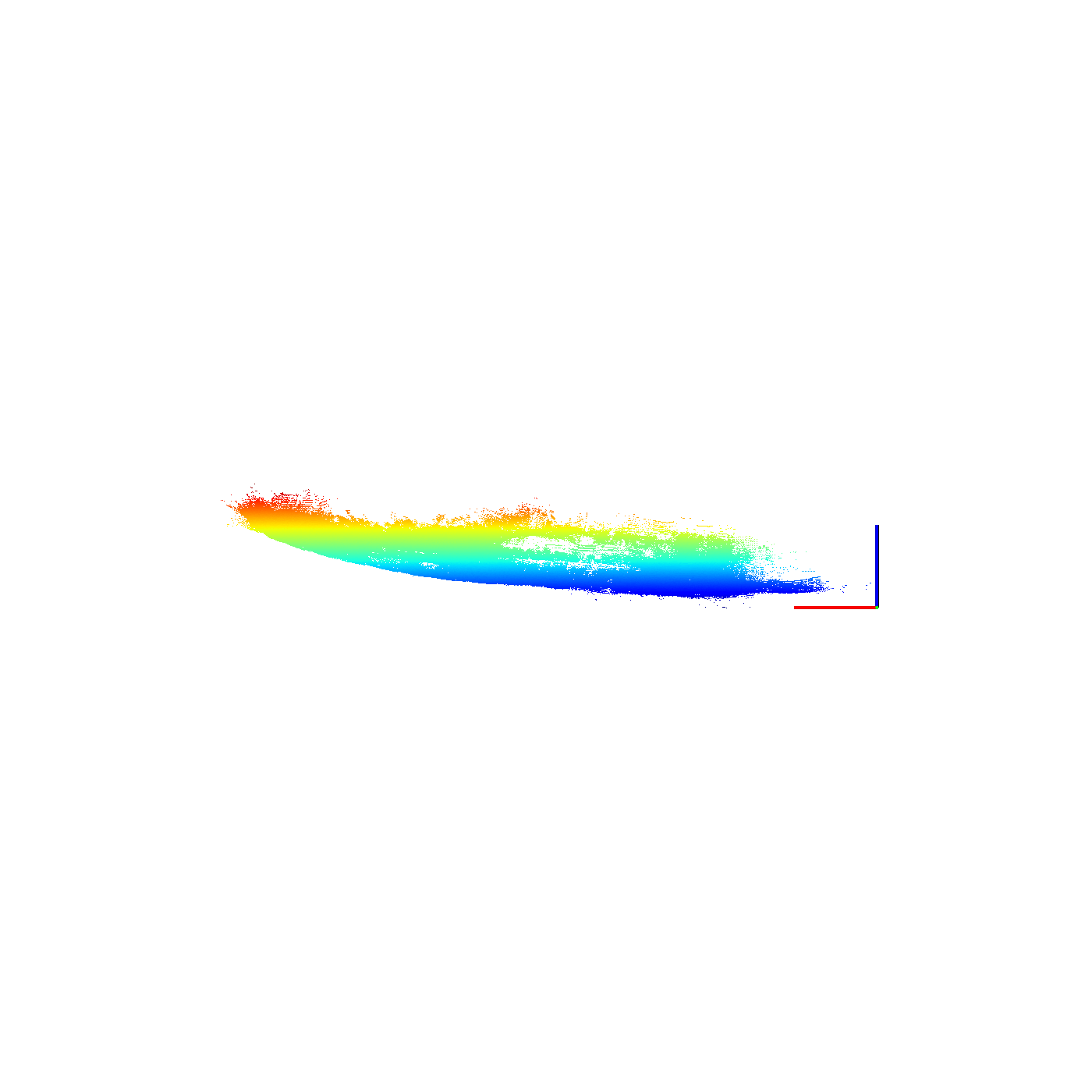}
    \end{minipage}\hfill
    \begin{minipage}[t]{0.32\linewidth}
        \includegraphics[width=\linewidth]{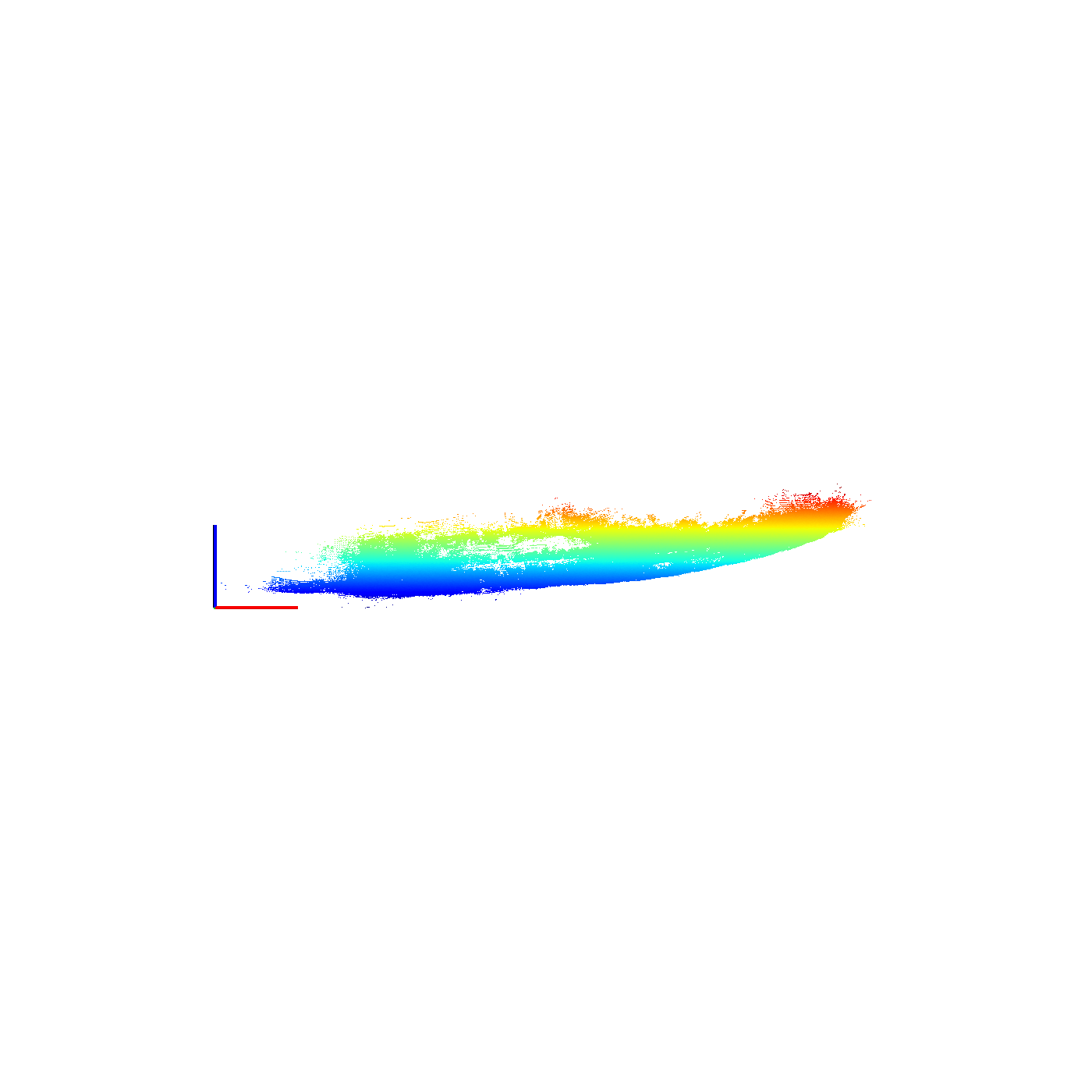}
    \end{minipage}
\end{minipage}
}

\caption{\textbf{Localization and mapping with POLI-augmented scan.} Qualitative visualization of localization and mapping results on the \textit{DCC} sequence from the HeLiPR~\citep{jung2024helipr} VLP-16 dataset. \textbf{Top:} results obtained using the state-of-the-art LiDAR odometry system GLIM~\citep{koide2024glim}. \textbf{Bottom:} results obtained using GLIM~\citep{koide2024glim} augmented with POLI.}
\label{fig:poly_odom}
\end{figure*}

\clearpage
\newpage
\label{app:global}
\begin{figure*}[t]
    \centering
    \setlength{\tabcolsep}{2pt}

    % ---------------- Row 1: batch_0007 (3 images) ----------------
    \begin{minipage}[t]{0.32\textwidth}
        \centering
        \includegraphics[width=\linewidth]{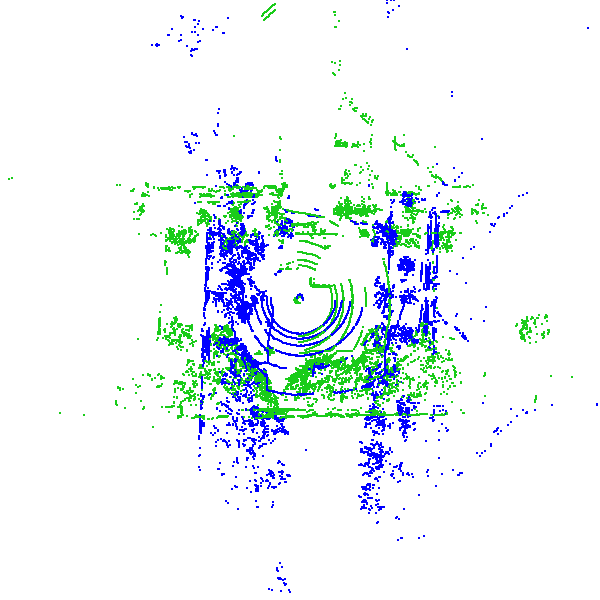}
    \end{minipage}\hfill
    \begin{minipage}[t]{0.32\textwidth}
        \centering
        \includegraphics[width=\linewidth]{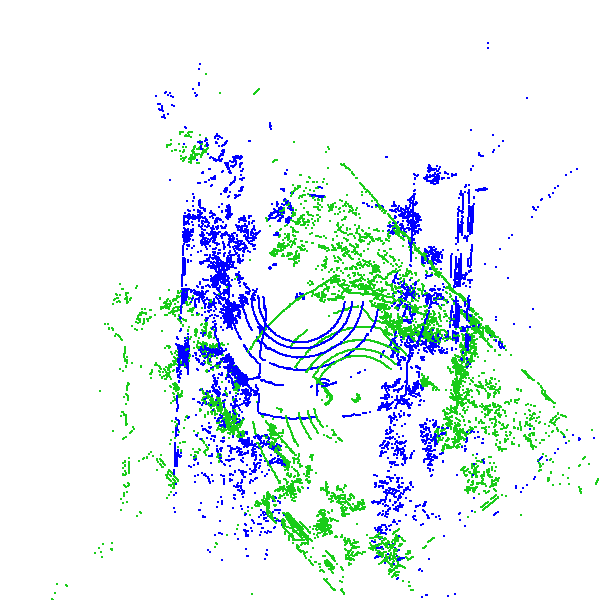}
    \end{minipage}\hfill
    \begin{minipage}[t]{0.32\textwidth}
        \centering
        \includegraphics[width=\linewidth]{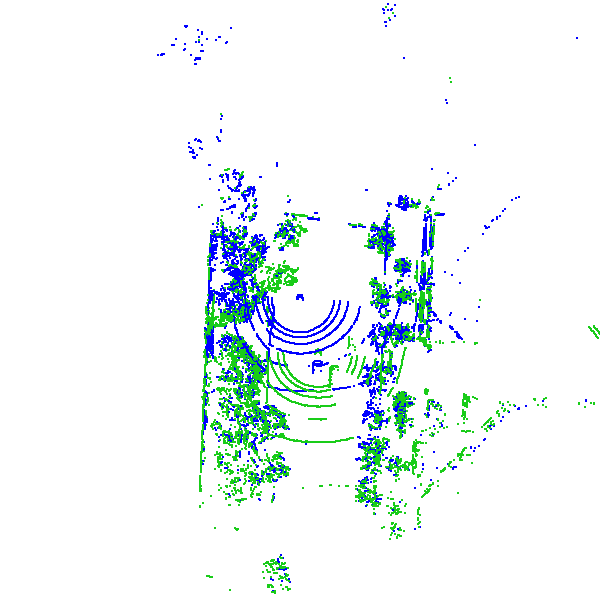}
    \end{minipage}

    % ---------------- Row 2: batch_0010 (3 images) ----------------
    \begin{minipage}[t]{0.32\textwidth}
        \centering
        \includegraphics[width=\linewidth]{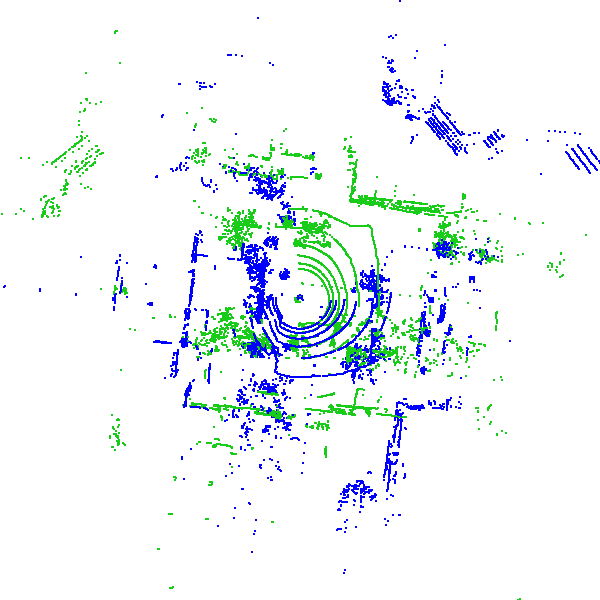}
    \end{minipage}\hfill
    \begin{minipage}[t]{0.32\textwidth}
        \centering
        \includegraphics[width=\linewidth]{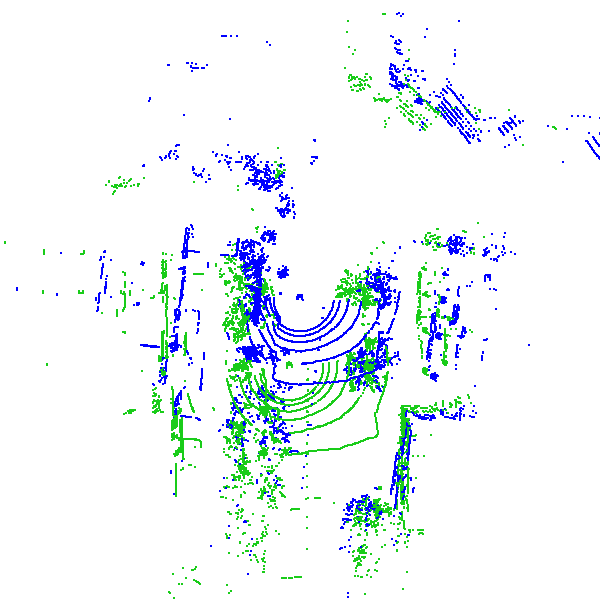}
    \end{minipage}\hfill
    \begin{minipage}[t]{0.32\textwidth}
        \centering
        \includegraphics[width=\linewidth]{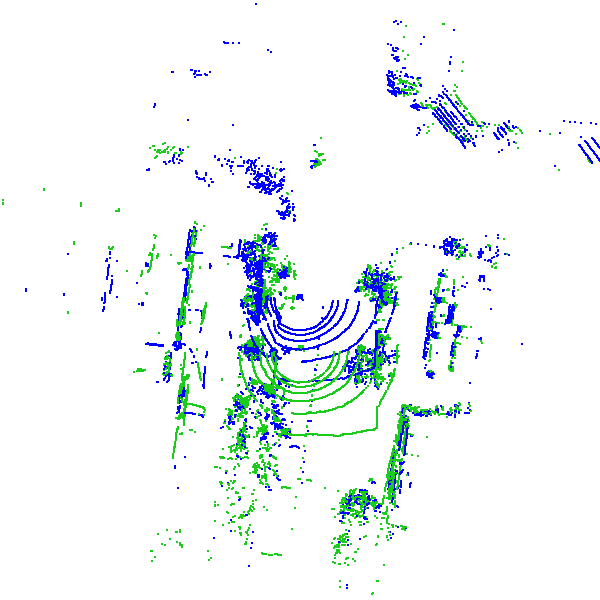}
    \end{minipage}

    % ---------------- Row 3: batch_0021 (3 images) ----------------
    \begin{minipage}[t]{0.32\textwidth}
        \centering
        \includegraphics[width=\linewidth]{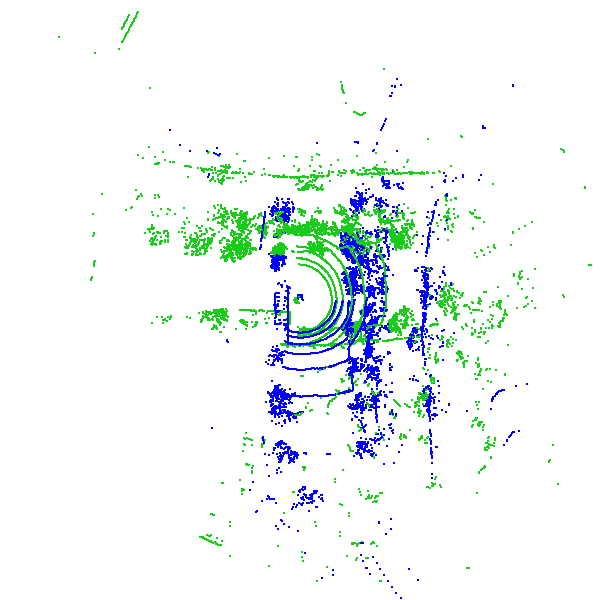}
    \end{minipage}\hfill
    \begin{minipage}[t]{0.32\textwidth}
        \centering
        \includegraphics[width=\linewidth]{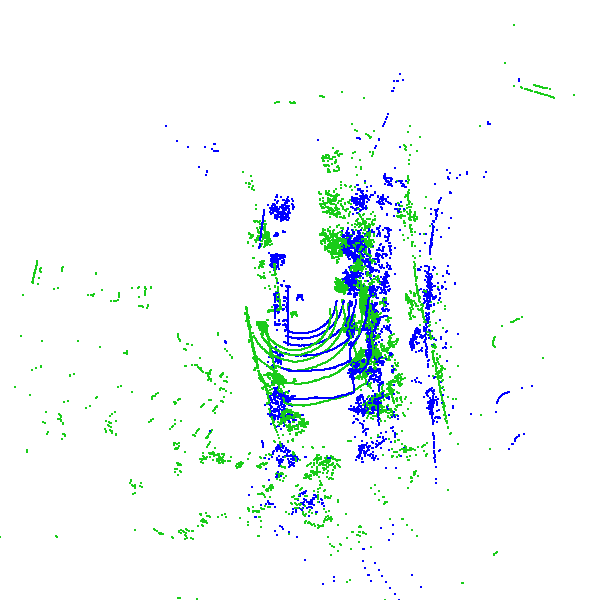}
    \end{minipage}\hfill
    \begin{minipage}[t]{0.32\textwidth}
        \centering
        \includegraphics[width=\linewidth]{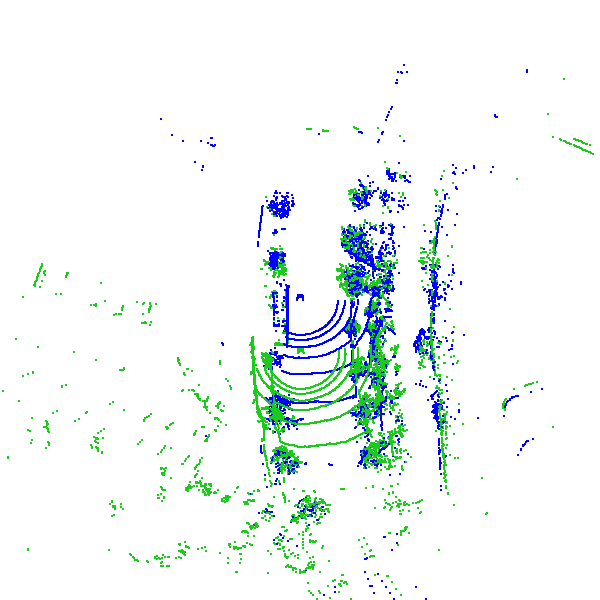}
    \end{minipage}

    % ---------------- Row 4: batch_0024 (3 images) ----------------
    \begin{minipage}[t]{0.32\textwidth}
        \centering
        \includegraphics[width=\linewidth]{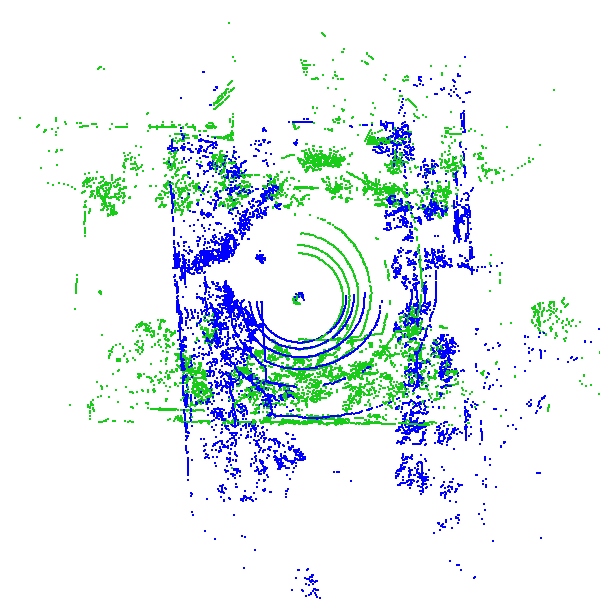}
        Before Registration
    \end{minipage}\hfill
    \begin{minipage}[t]{0.32\textwidth}
        \centering
        \includegraphics[width=\linewidth]{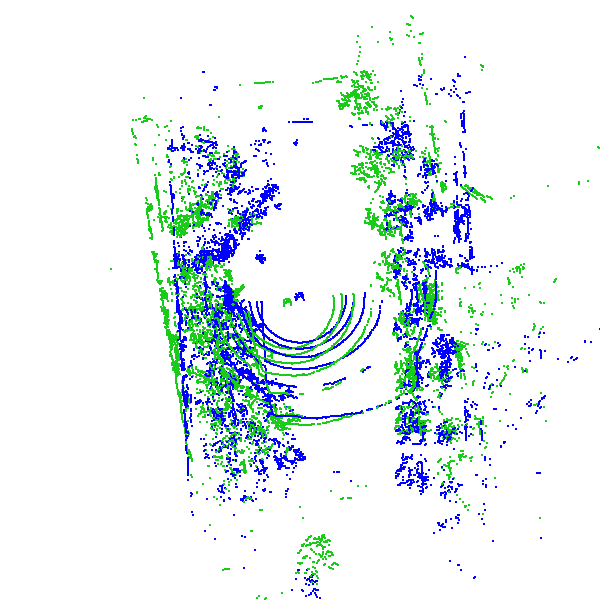}
        KISS-Matcher~\citep{lim2024kiss}
    \end{minipage}\hfill
    \begin{minipage}[t]{0.32\textwidth}
        \centering
        \includegraphics[width=\linewidth]{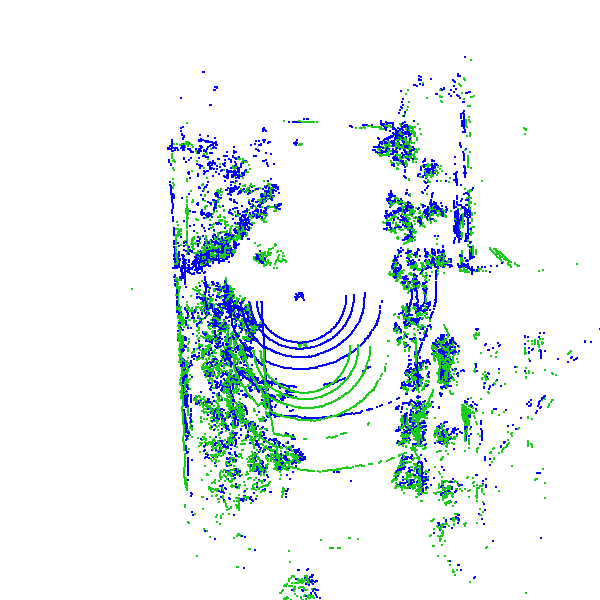}
        KISS-Matcher~\citep{lim2024kiss} + POLI
    \end{minipage}

    \caption{\textbf{Global registration with POLI-augmented scan.} Qualitative visualization of global registration results using POLI-augmented scans. \textbf{Left}: input point cloud pair. \textbf{Middle}: results from KISS-Matcher~\citep{lim2024kiss}. \textbf{Right}: results from KISS-Matcher~\citep{lim2024kiss} augmented with POLI.}
    \label{fig:poly_kiss_qual_12}
\end{figure*}

\clearpage
\newpage
\label{app:aug}

\begin{figure*}[t]
    \centering
    \setlength{\tabcolsep}{0pt}
    \renewcommand{\arraystretch}{1}

    % ---- 튜닝 파라미터 ----
    \newcommand{\colgap}{10pt}
    \newcommand{\rowgap}{4mm}
    \newcommand{\imgh}{0.170\textheight}
    \newcommand{\imgw}{0.31\textwidth}
    \newcommand{\rowlabelw}{18pt}
    % ----------------------
    \begin{tabular}{@{}p{\rowlabelw}@{\hspace{6pt}}c@{\hspace{\colgap}}c@{\hspace{\colgap}}c@{}}

        % ---------- Row 1 ----------
        \rotatebox{90}{\textit{DCC}} &
        \includegraphics[width=\imgw,height=\imgh,keepaspectratio,trim=650 300 650 300,clip]{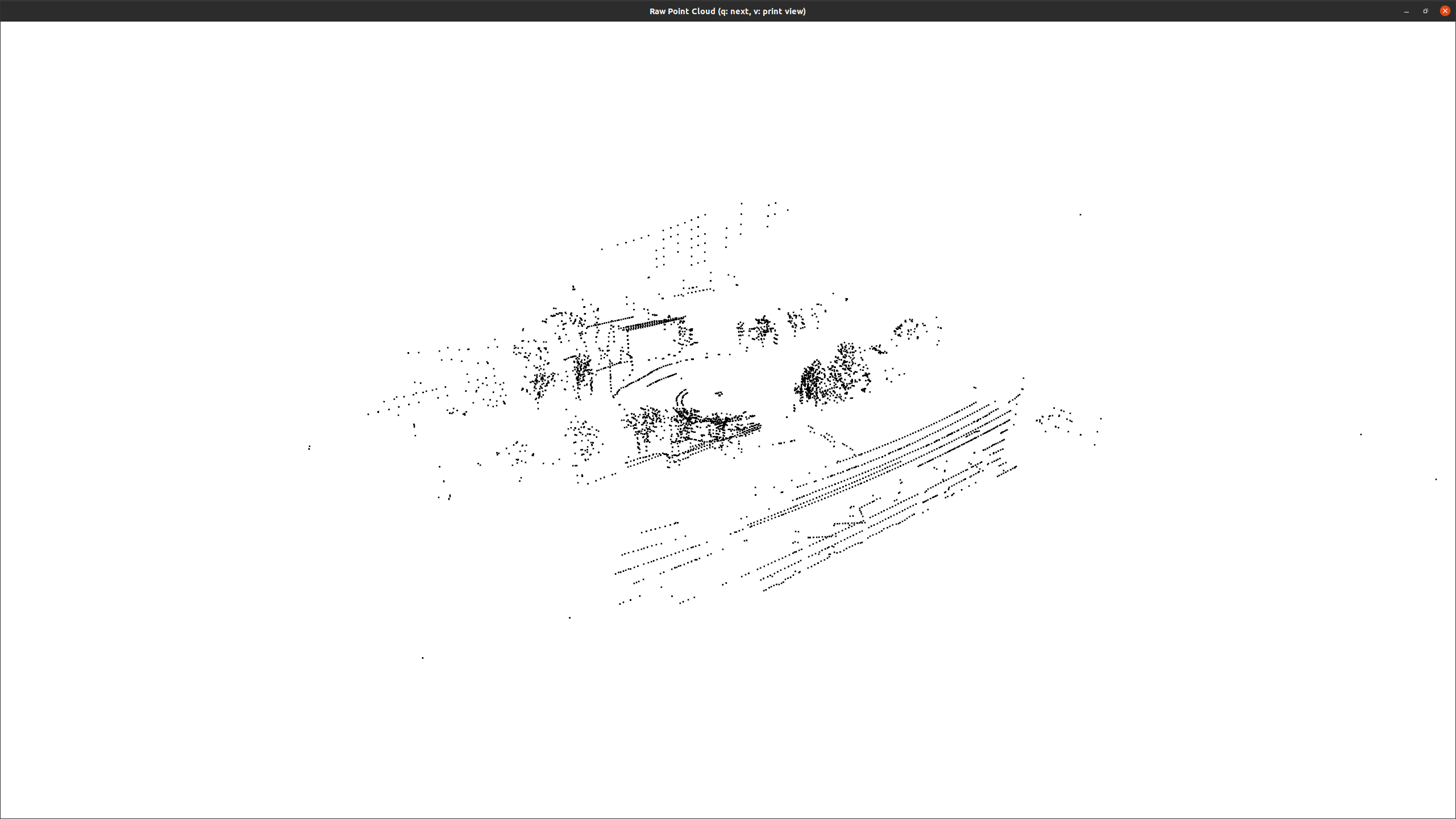} &
        \includegraphics[width=\imgw,height=\imgh,keepaspectratio,trim=650 300 650 300,clip]{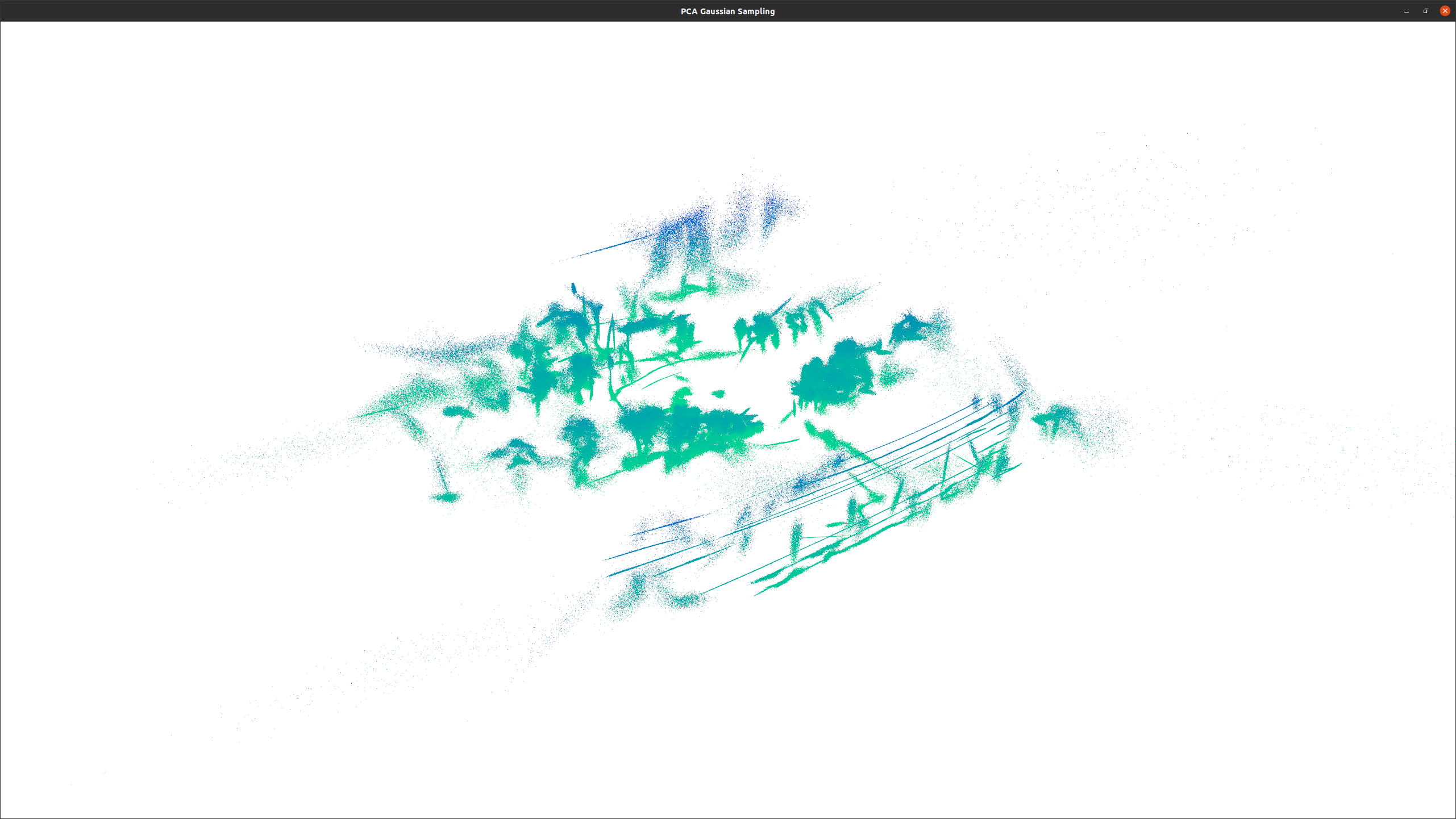} &
        \includegraphics[width=\imgw,height=\imgh,keepaspectratio,trim=650 300 650 300,clip]{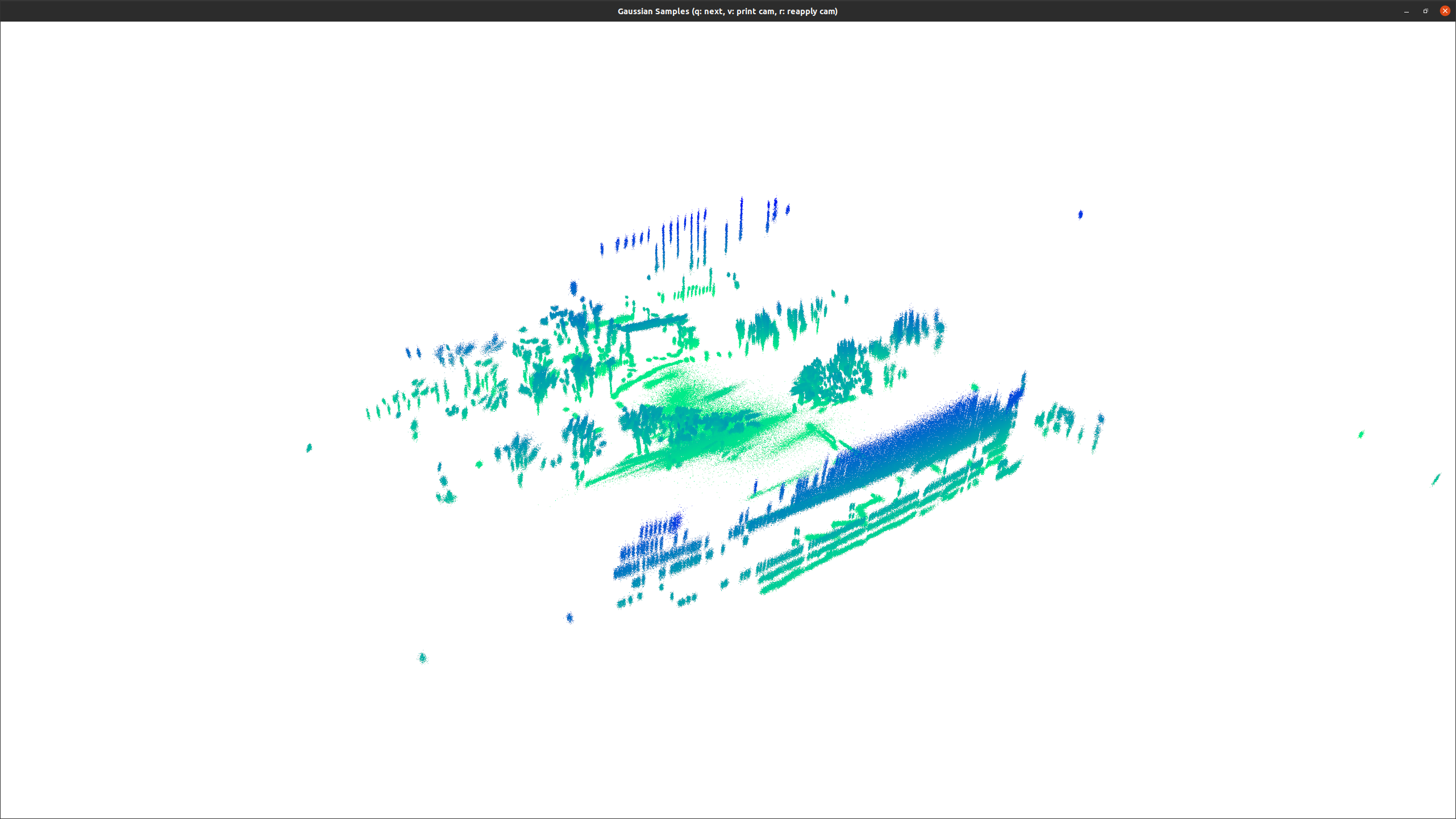} \\[\rowgap]

        % ---------- Row 2 ----------
        \rotatebox{90}{\textit{Riverside}} &
        \includegraphics[width=\imgw,height=\imgh,keepaspectratio,trim=750 400 650 500,clip]{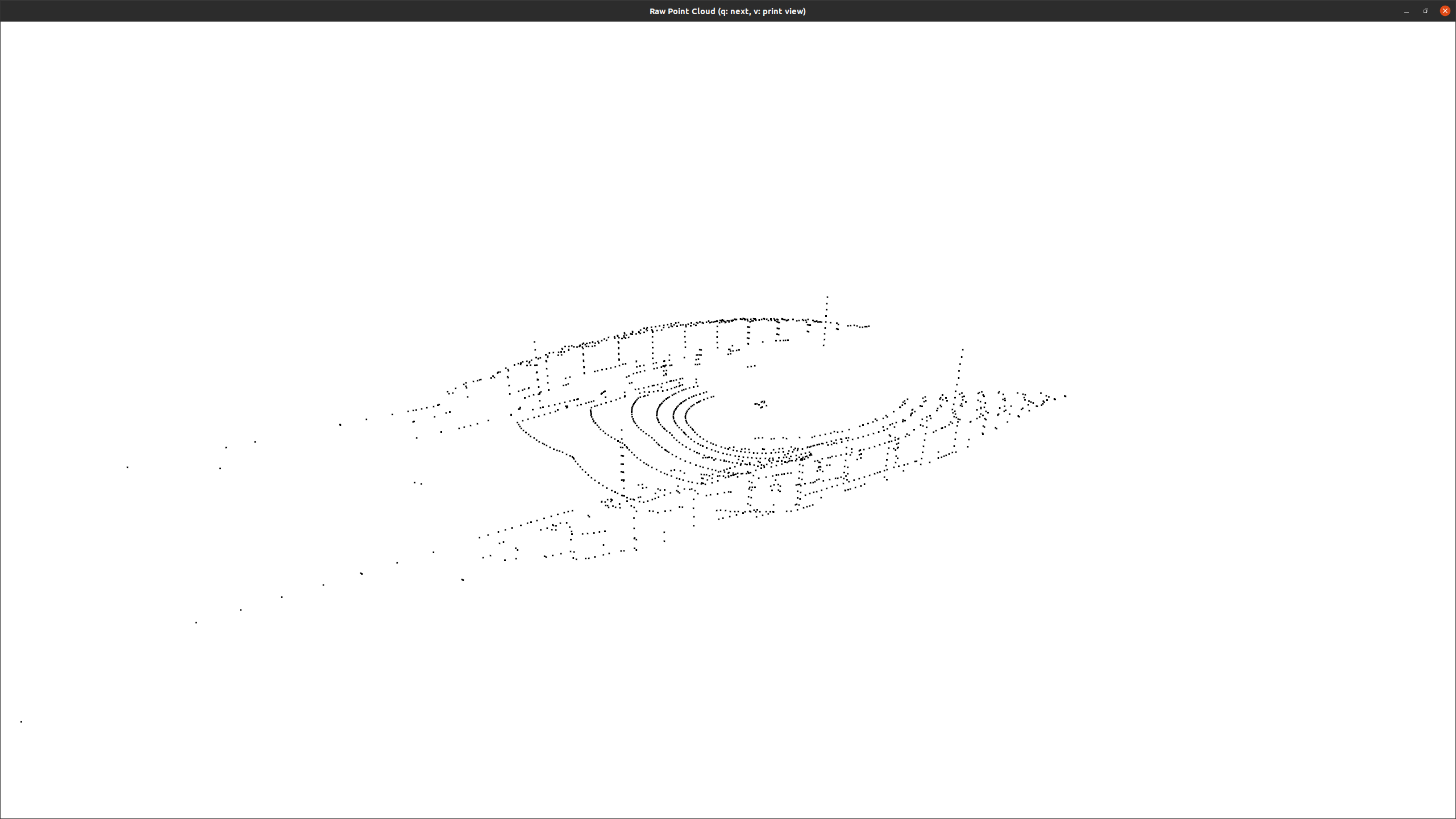} &
        \includegraphics[width=\imgw,height=\imgh,keepaspectratio,trim=750 400 650 500,clip]{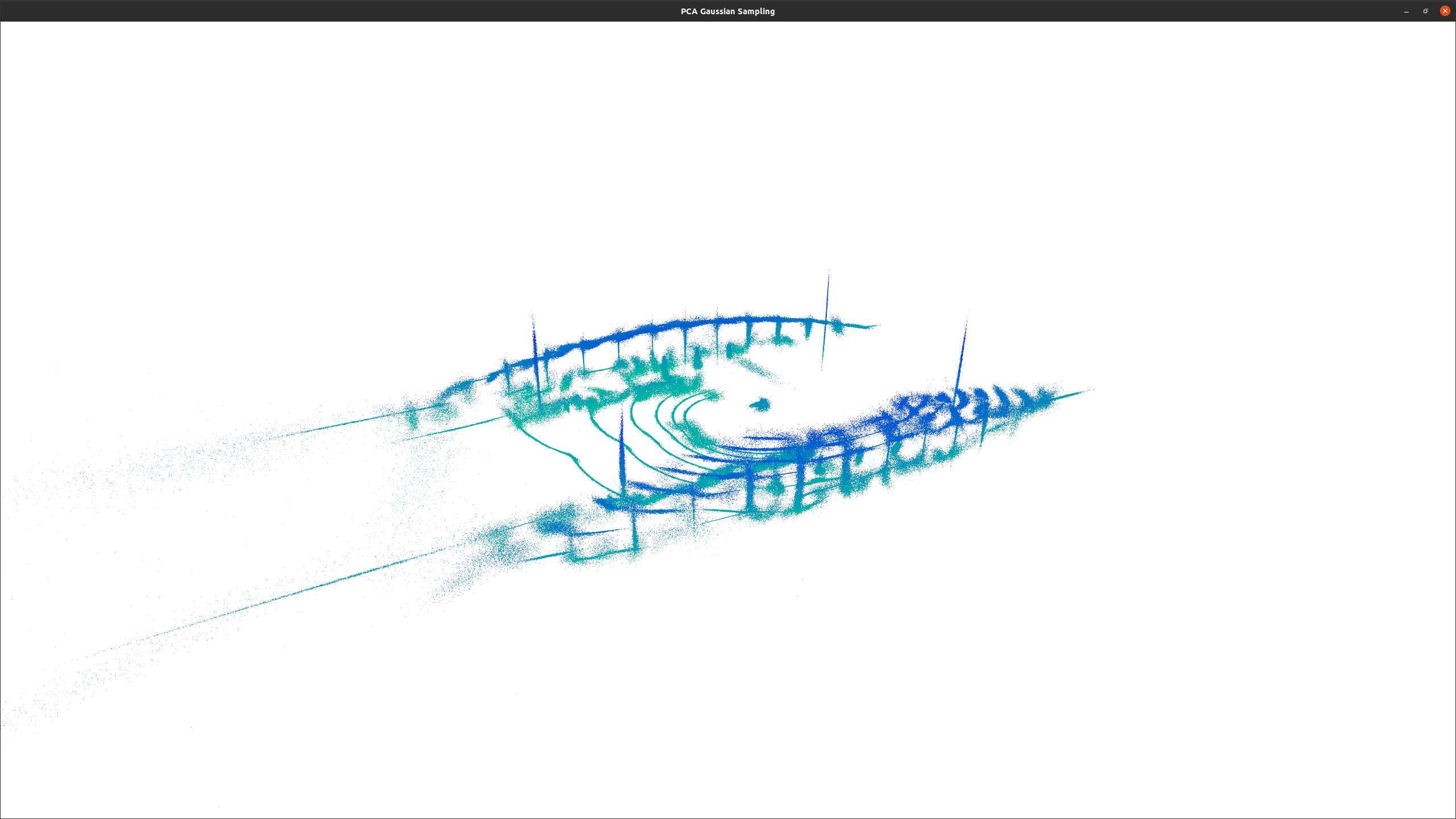} &
        \includegraphics[width=\imgw,height=\imgh,keepaspectratio,trim=750 400 650 500,clip]{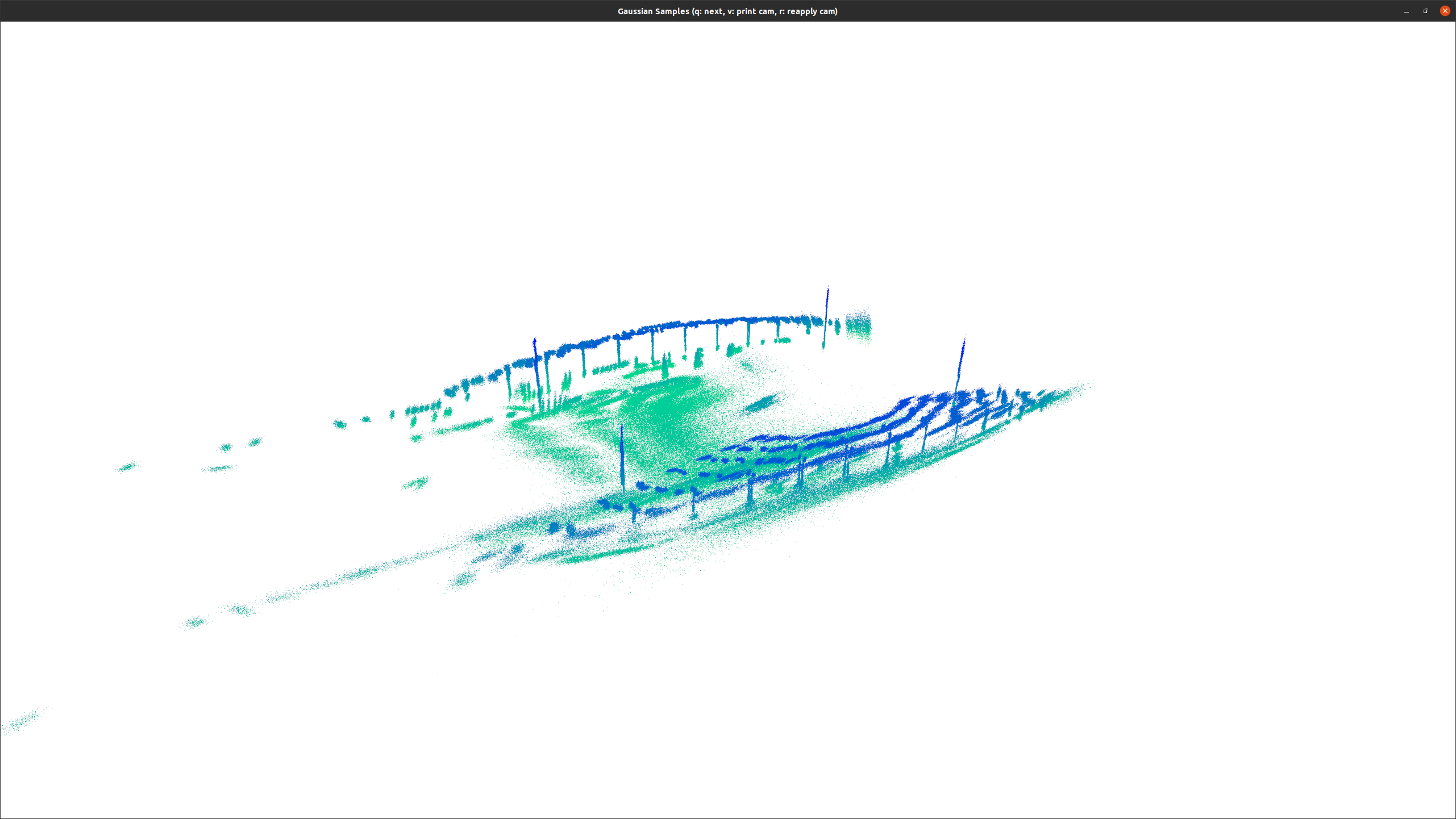} \\[\rowgap]

        % ---------- Row 3 ----------
        \rotatebox{90}{\textit{Bridge}} &
        \includegraphics[width=\imgw,height=\imgh,keepaspectratio,trim=650 400 850 400,clip]{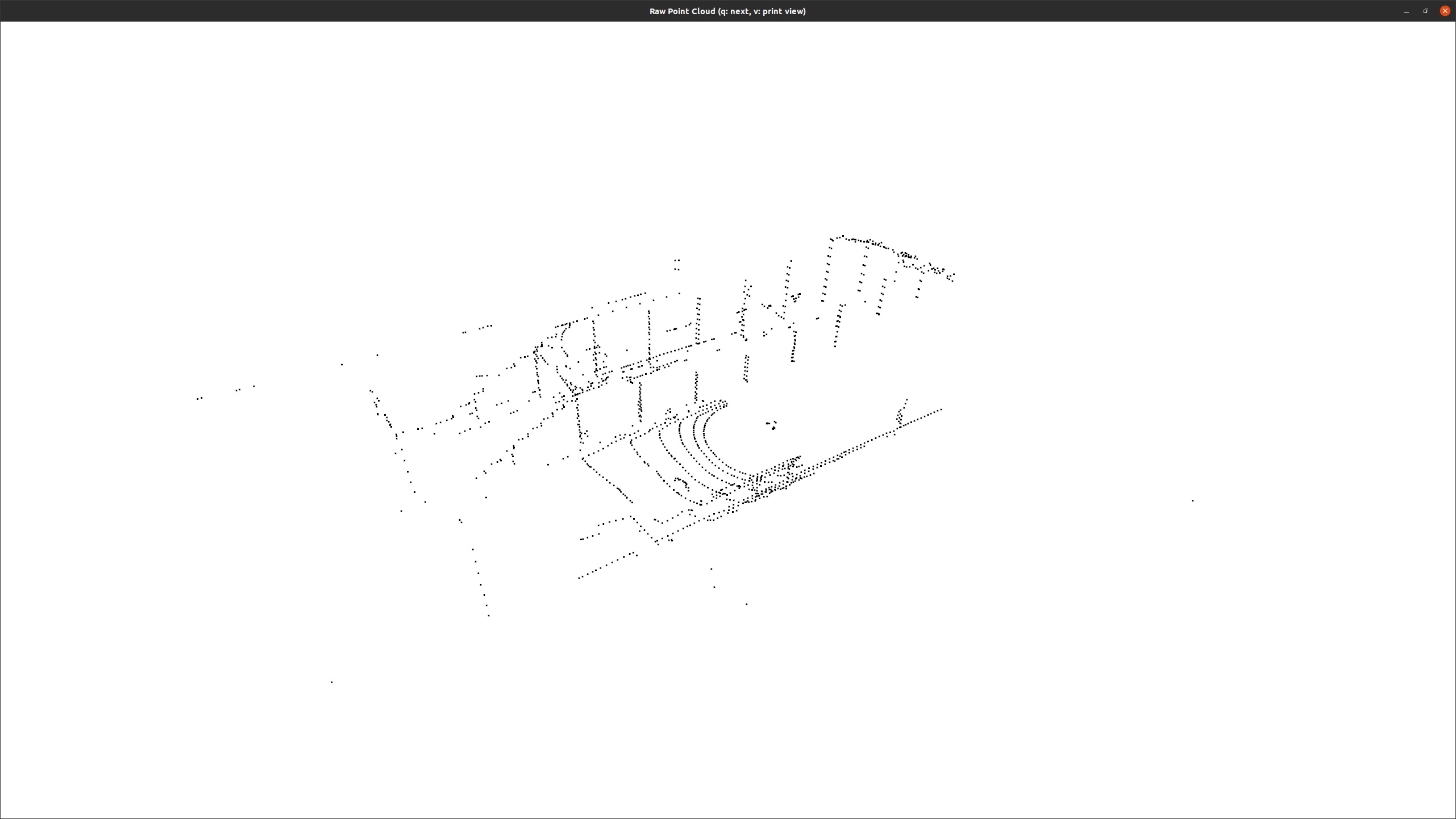} &
        \includegraphics[width=\imgw,height=\imgh,keepaspectratio,trim=650 400 850 400,clip]{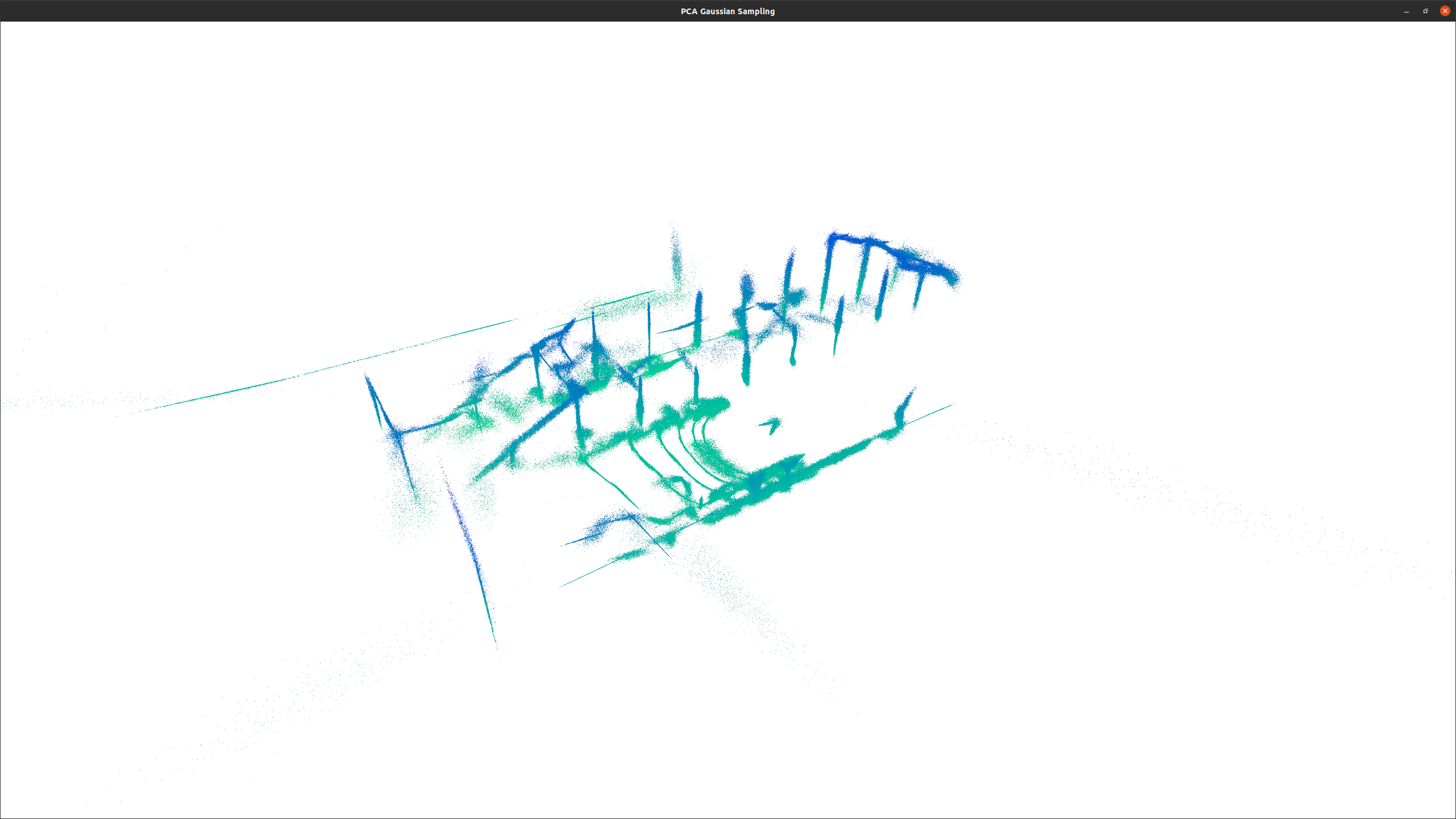} &
        \includegraphics[width=\imgw,height=\imgh,keepaspectratio,trim=650 400 850 400,clip]{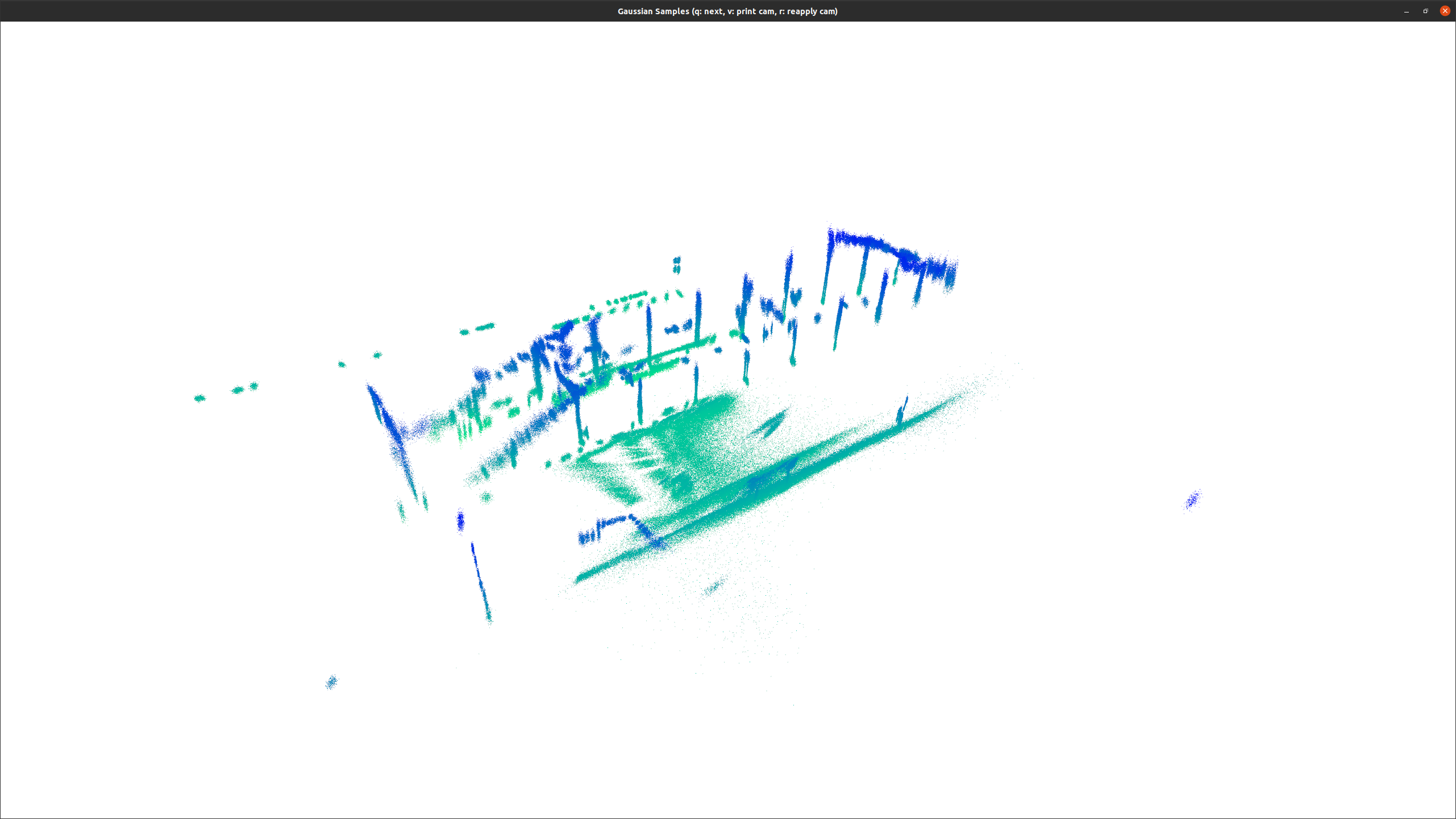} \\[\rowgap]

        % ---------- Row 4 ----------
        \rotatebox{90}{\textit{Town}} &
        \includegraphics[width=\imgw,height=\imgh,keepaspectratio,trim=750 300 550 400,clip]{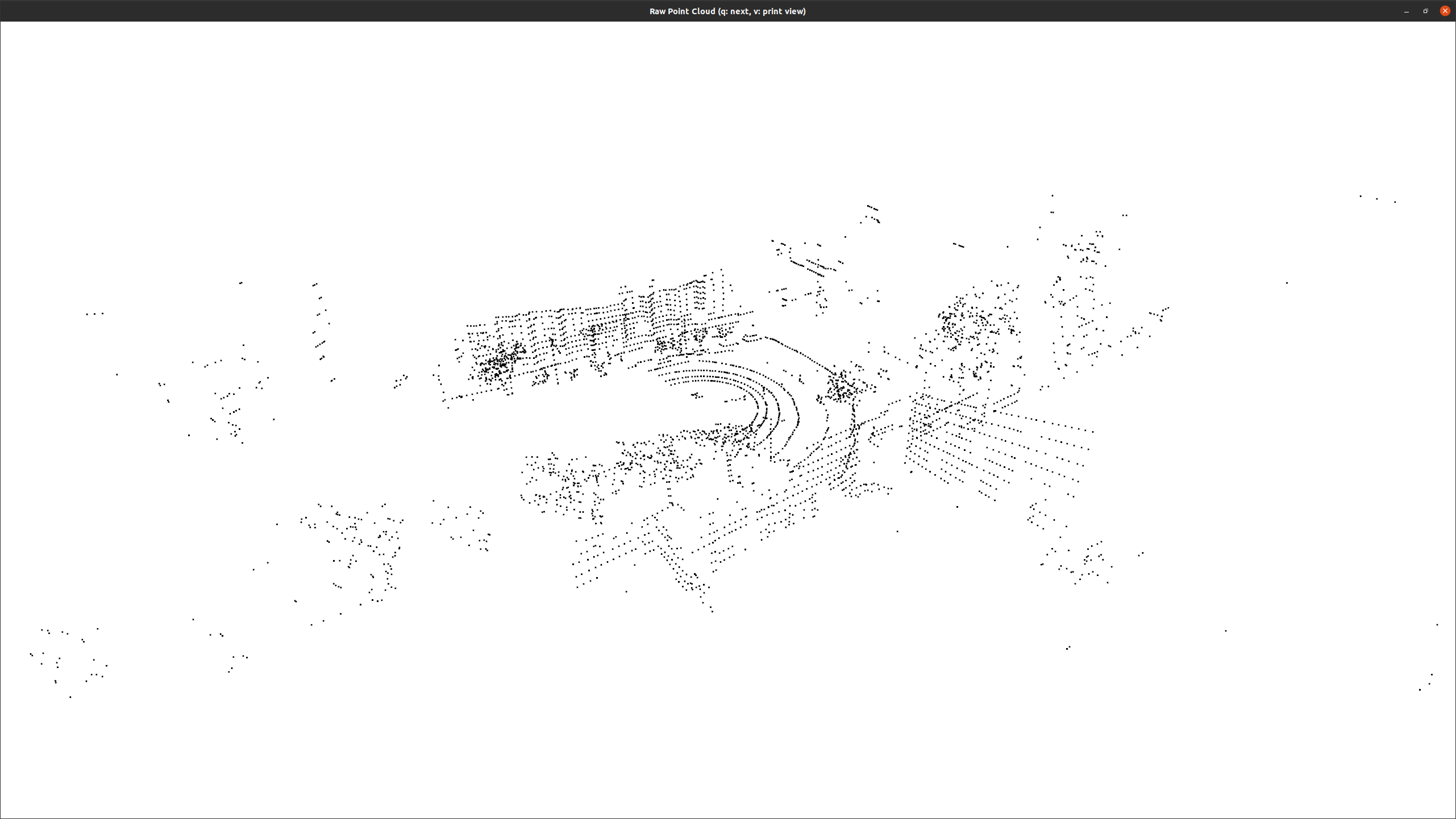} &
        \includegraphics[width=\imgw,height=\imgh,keepaspectratio,trim=750 300 550 400,clip]{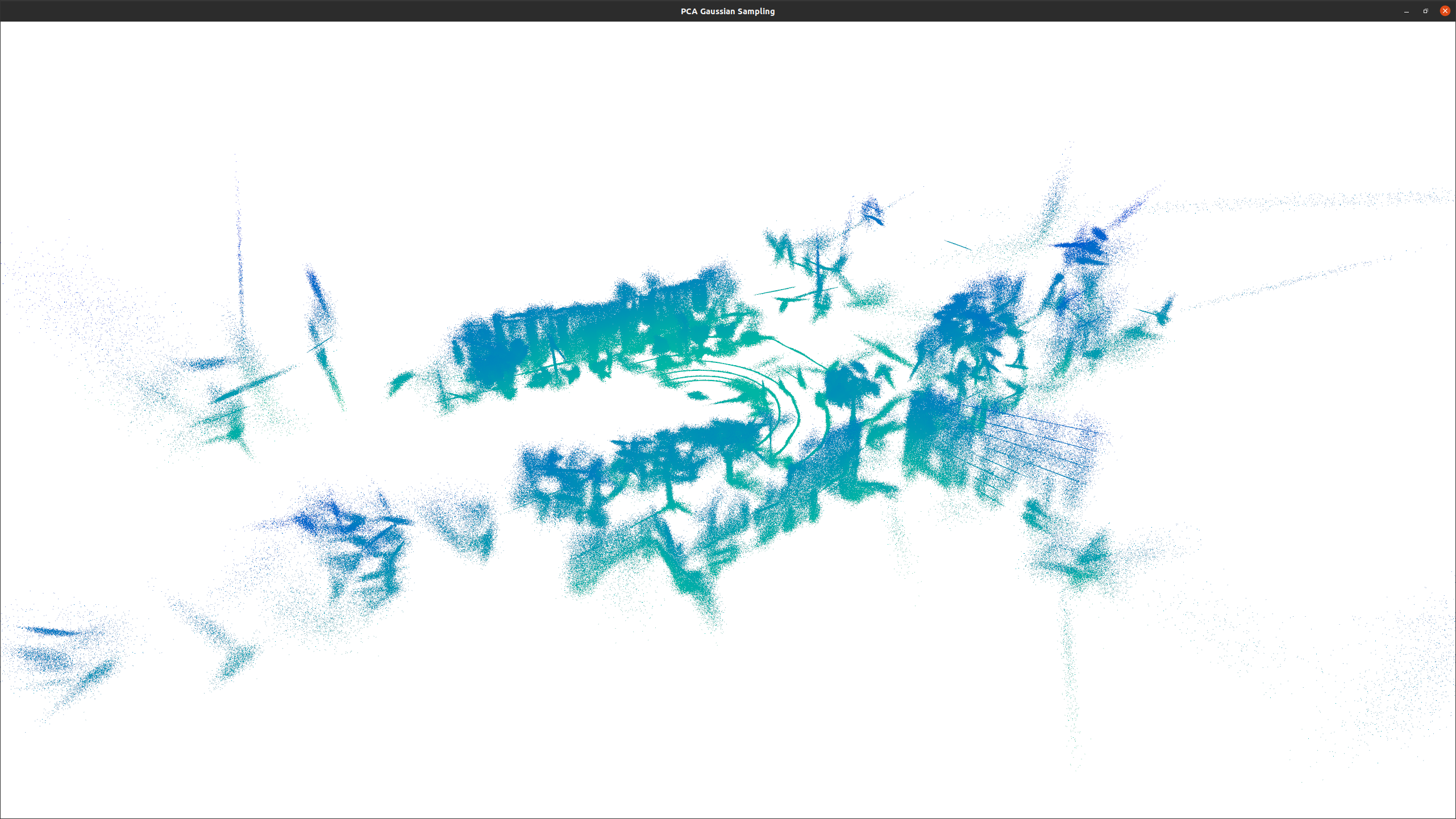} &
        \includegraphics[width=\imgw,height=\imgh,keepaspectratio,trim=750 300 550 400,clip]{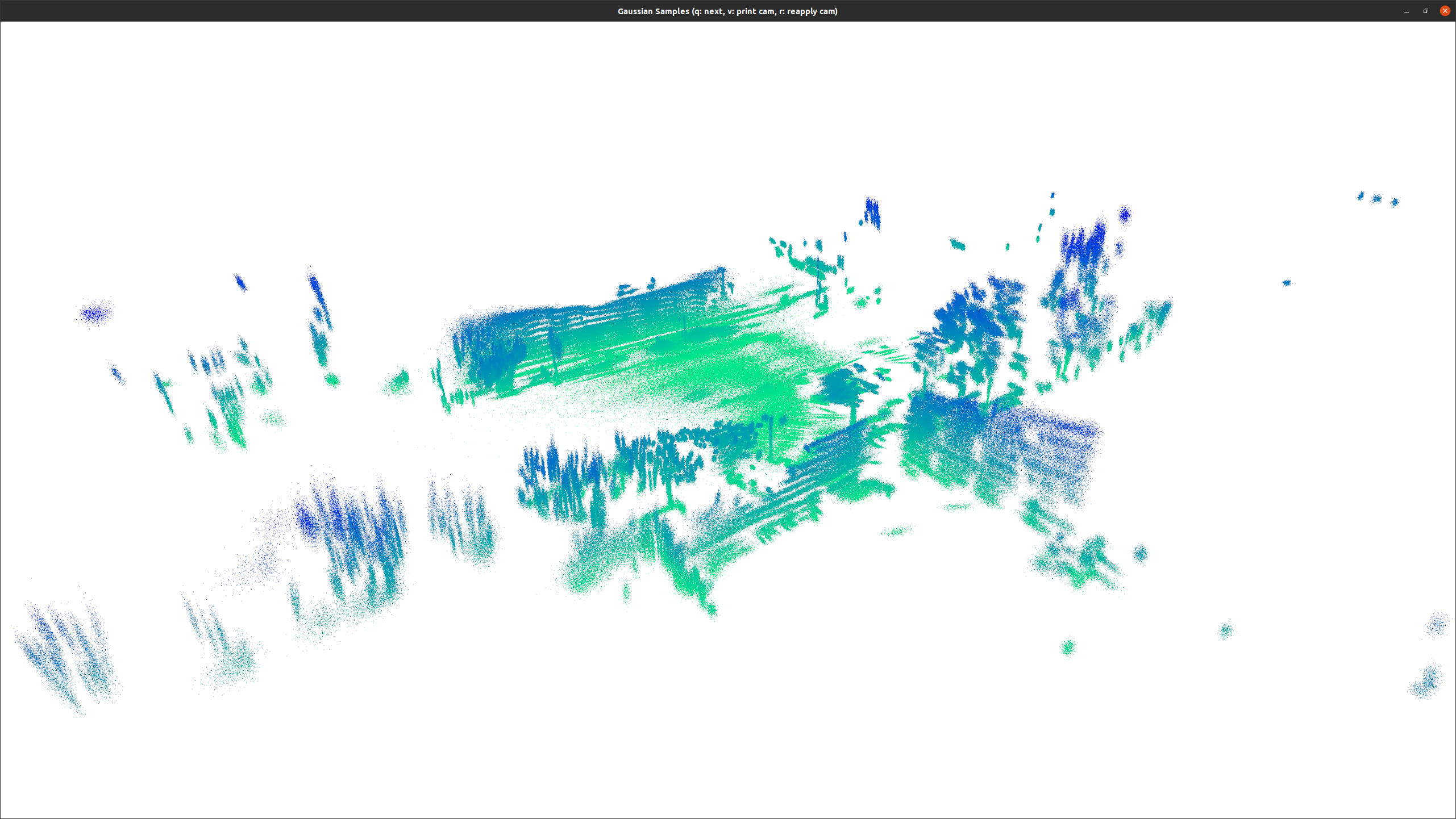} \\[\rowgap]

        % ---------- Row 5 ----------
        \rotatebox{90}{\textit{Roundabout}} &
        \includegraphics[width=\imgw,height=\imgh,keepaspectratio,trim=750 300 350 200,clip]{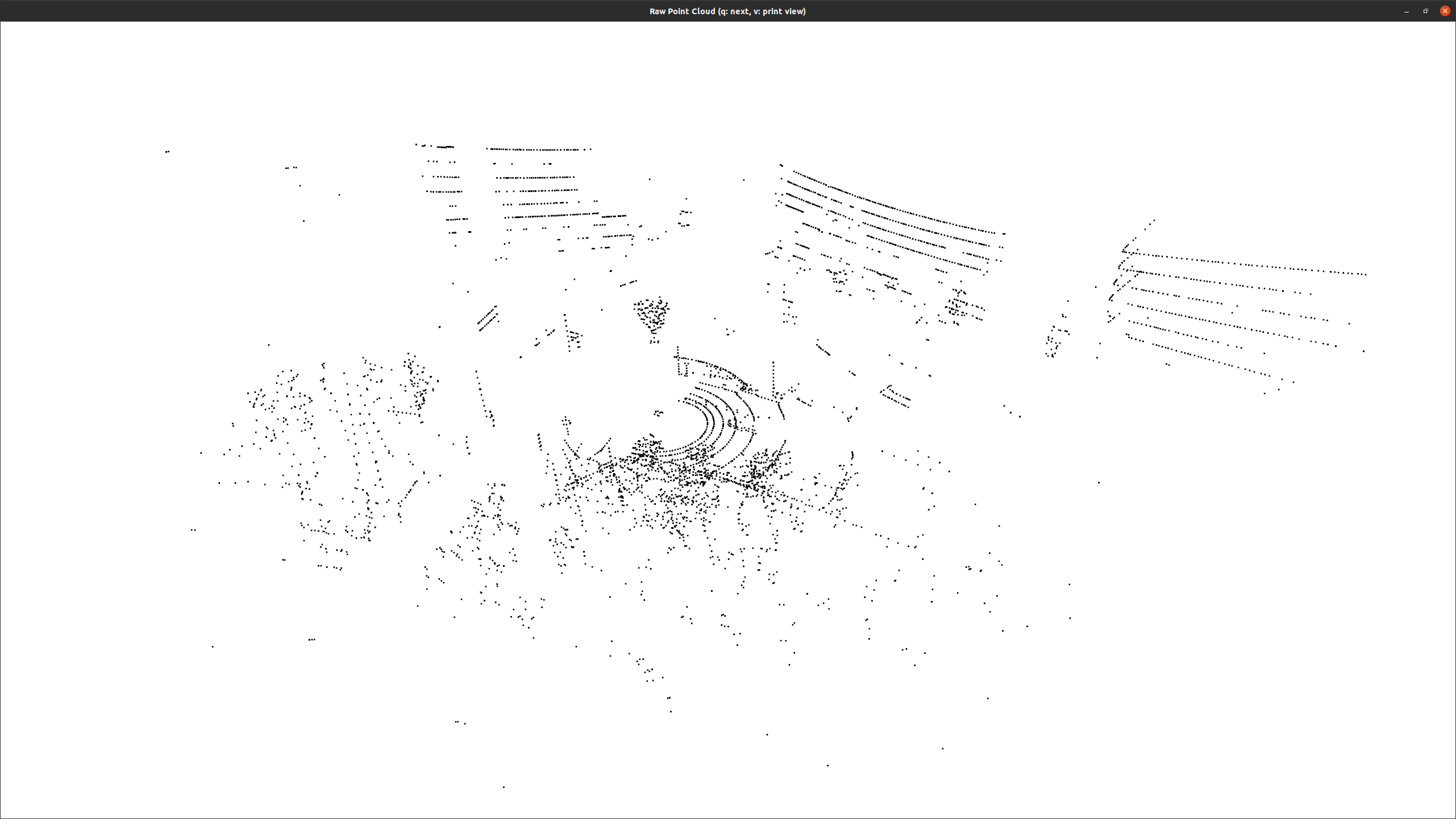} &
        \includegraphics[width=\imgw,height=\imgh,keepaspectratio,trim=750 300 350 200,clip]{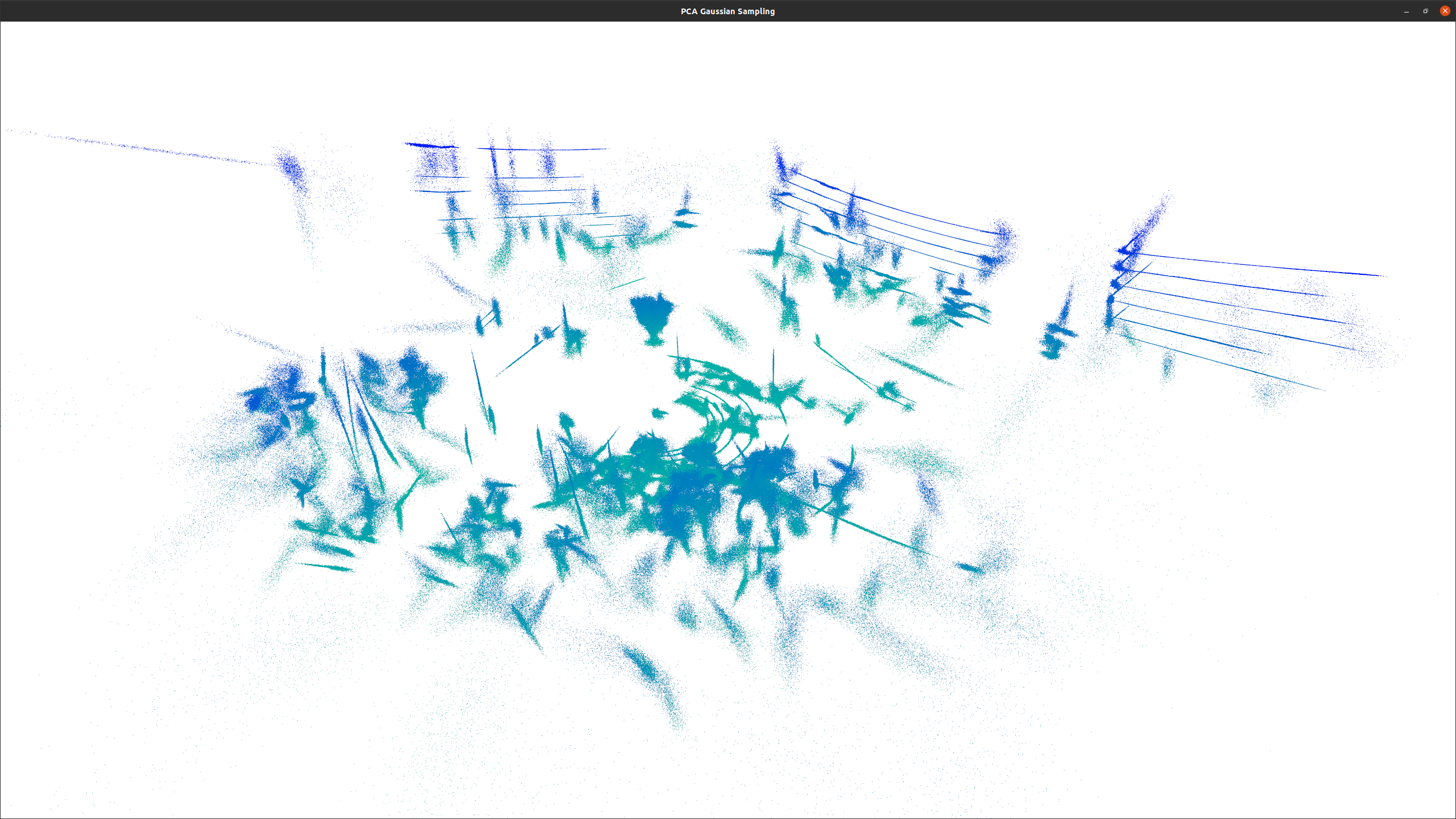} &
        \includegraphics[width=\imgw,height=\imgh,keepaspectratio,trim=750 300 350 200,clip]{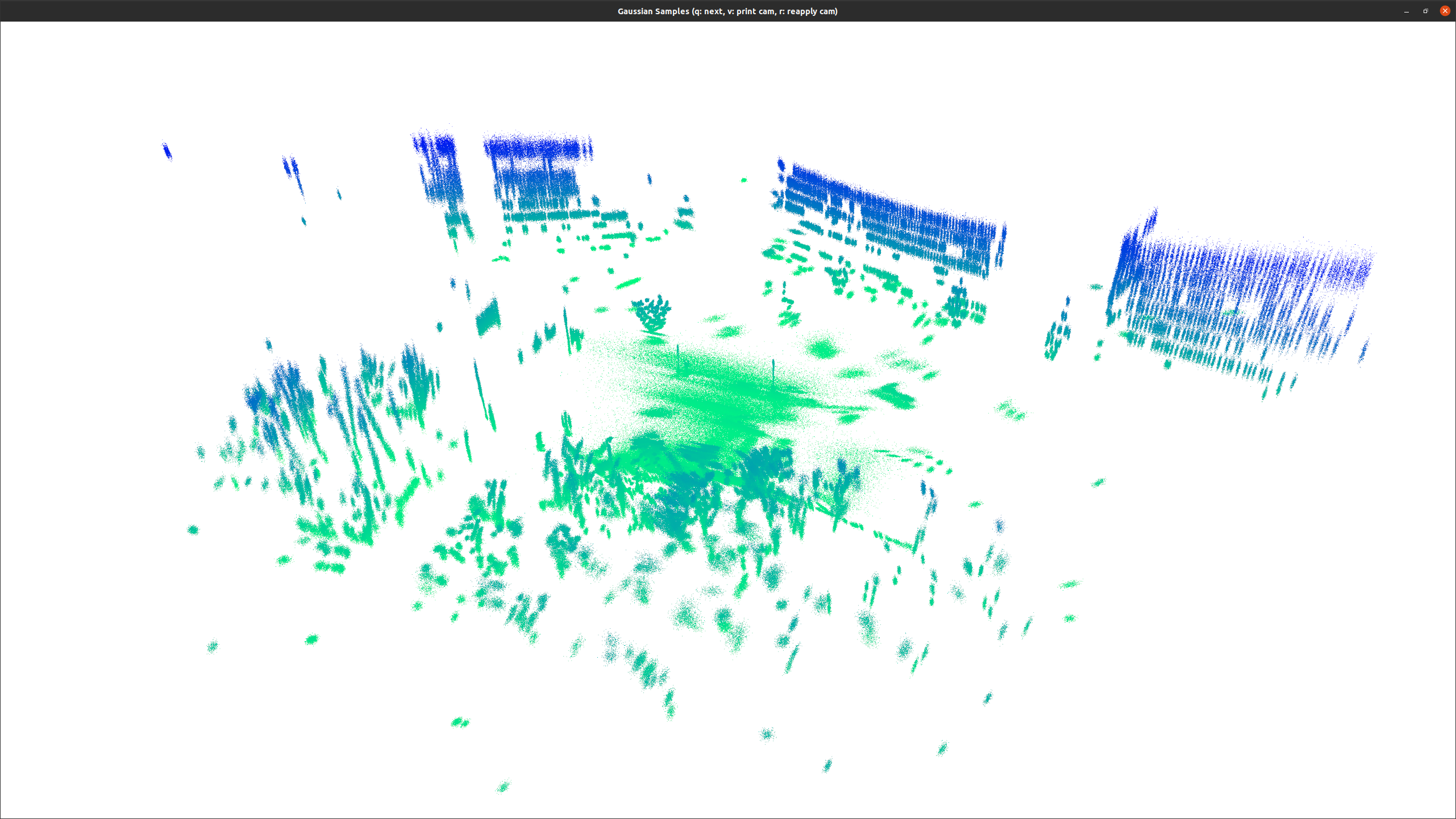} \\[\rowgap]

        % ---------- Row 6 ----------
        \rotatebox{90}{\textit{KAIST}} &
        \includegraphics[width=\imgw,height=\imgh,keepaspectratio,trim=650 300 650 300,clip]{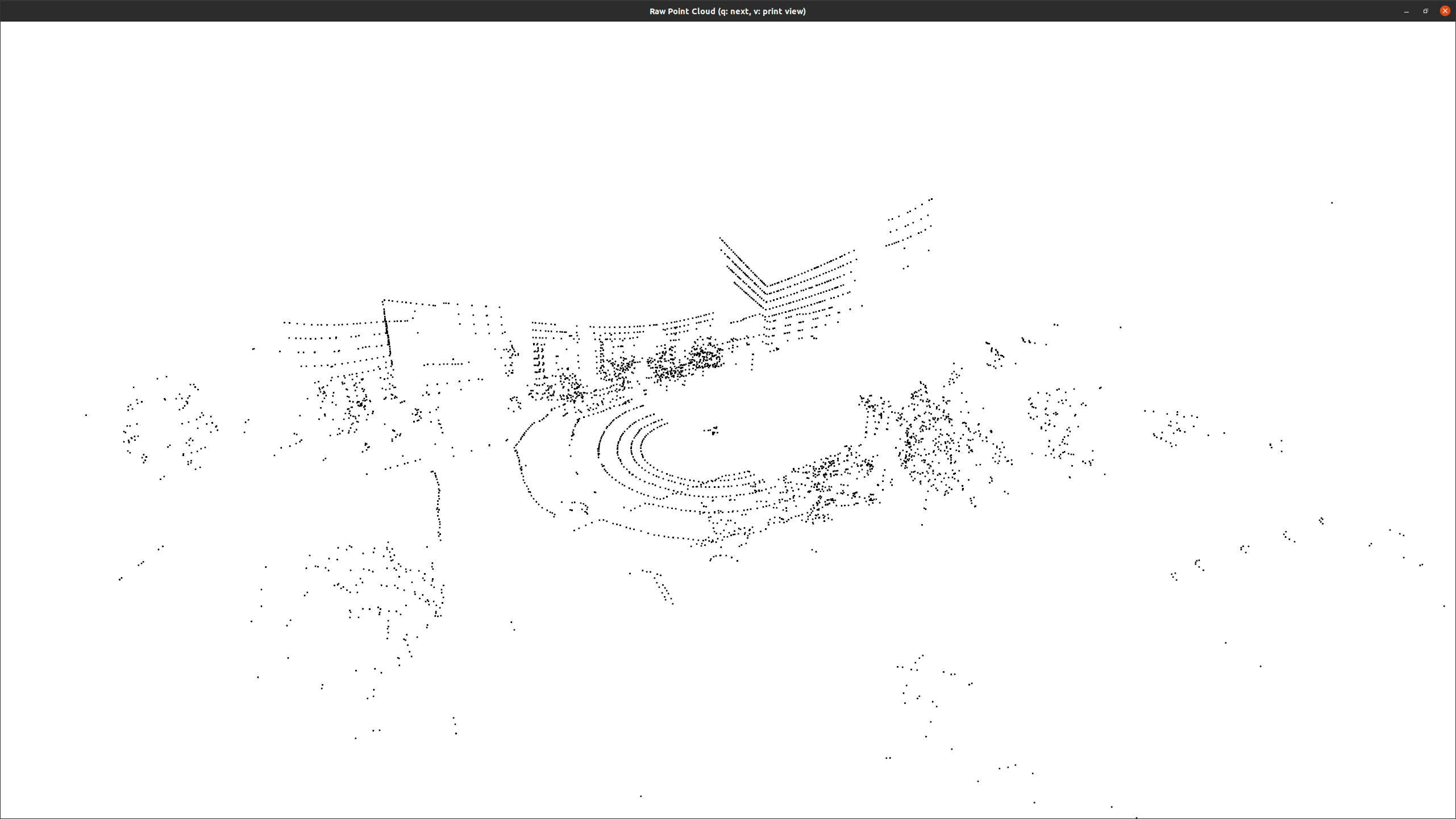} &
        \includegraphics[width=\imgw,height=\imgh,keepaspectratio,trim=650 300 650 300,clip]{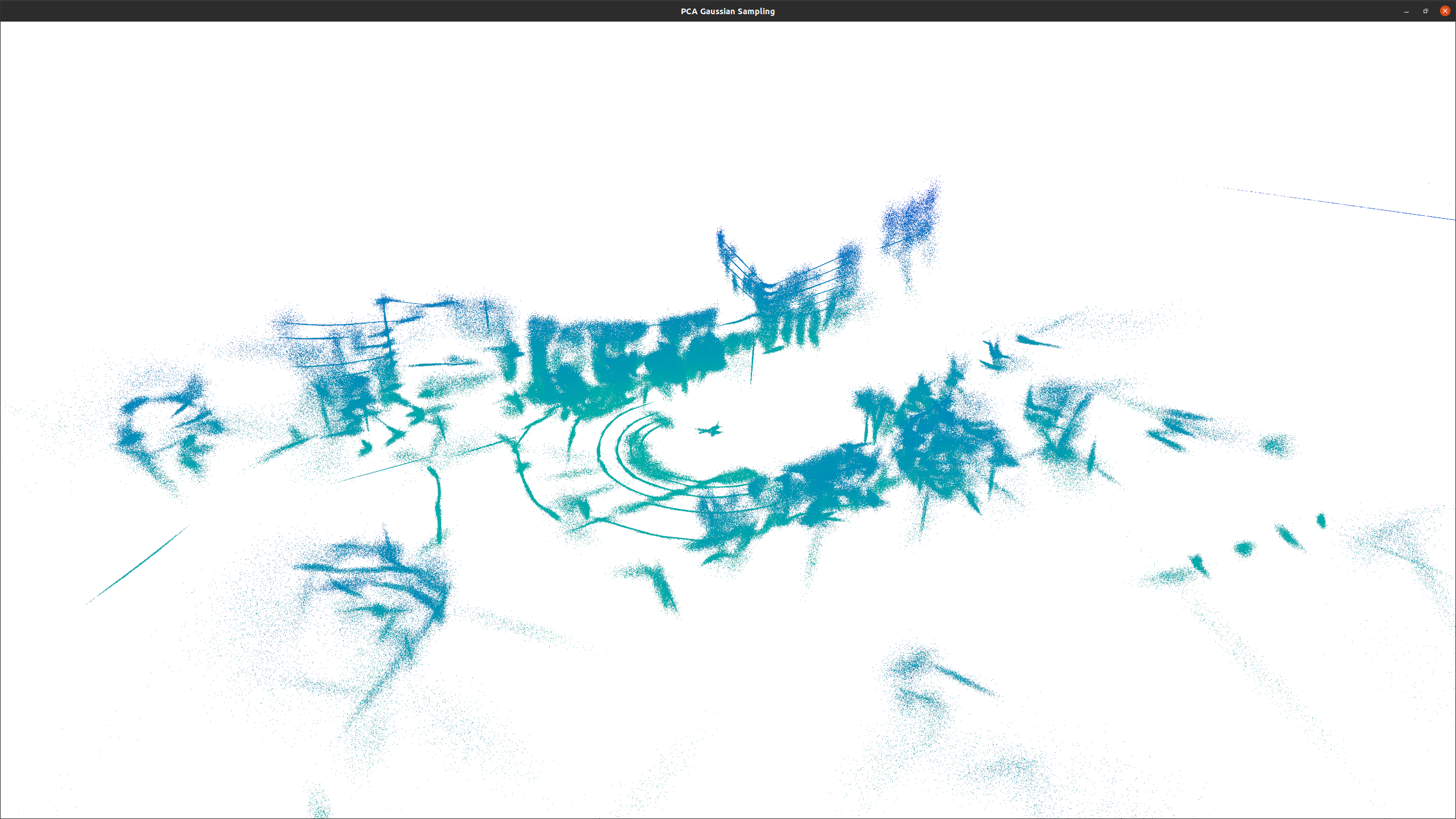} &
        \includegraphics[width=\imgw,height=\imgh,keepaspectratio,trim=650 300 650 300,clip]{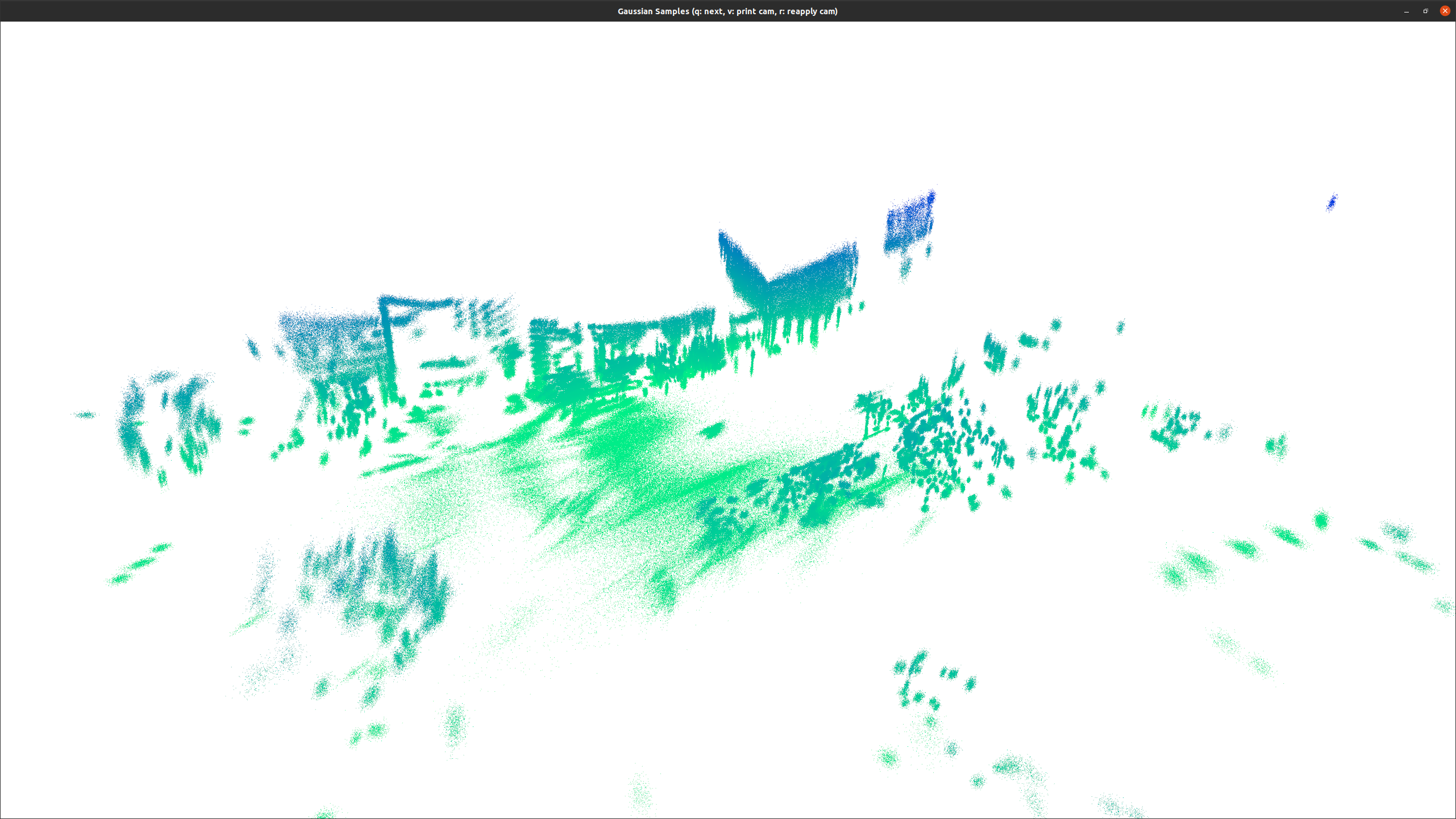} \\[6pt]

        % ---------- Column labels ----------
        &
        \multicolumn{1}{c}{Raw} &
        \multicolumn{1}{c}{PCA} &
        \multicolumn{1}{c}{POLI} \\

    \end{tabular}
    \caption{\textbf{Data augmentation using POLI.} Qualitative visualization of scan augmentation results using POLI on the HeLiPR~\citep{jung2024helipr} VLP-16 dataset. \textbf{Left:} raw point cloud. \textbf{Middle:} PCA-based augmentation results. \textbf{Right:} POLI-based augmentation results. PCA struggles to generalize under range-dependent variations in point density, whereas POLI demonstrates robust generalization across such variations.}

    \label{fig:augs}
\end{figure*}

\end{document}